\documentclass{article}

\usepackage{microtype}
\usepackage{graphicx}
\usepackage{subcaption}
\usepackage{booktabs} 
\usepackage{xcolor}
\usepackage{hyperref}

\usepackage[preprint]{icml2026}

\usepackage{amsmath}
\usepackage{amssymb}
\usepackage{mathtools}
\usepackage{amsthm}
\usepackage{bm}
\usepackage{dsfont}
\usepackage{multirow}
\usepackage[capitalize,noabbrev]{cleveref}

\usepackage{tcolorbox}
\hypersetup{
  colorlinks=true,
  linkcolor=red,   
  citecolor=blue,  
}

\theoremstyle{plain}

\theoremstyle{definition}

\theoremstyle{remark}

\newcommand{\green}[1]{\textcolor{green}{#1}} 
\newcommand{\red}[1]{\textcolor{red}{#1}}

\definecolor{myyellow}{RGB}{255,192,0}

\usepackage[textsize=tiny]{todonotes}

\icmltitlerunning{Component-Based Out-of-Distribution Detection}

\begin{document}

\twocolumn[
  \icmltitle{Component-Based Out-of-Distribution Detection}




  \begin{icmlauthorlist}
    \icmlauthor{Wenrui Liu}{ict,ucas}
    \icmlauthor{Hong Chang}{ict,ucas}
    \icmlauthor{Ruibing Hou}{ict}
    \icmlauthor{Shiguang Shan}{ict,ucas}
    \icmlauthor{Xilin Chen}{ict,ucas}
  \end{icmlauthorlist}

  \icmlaffiliation{ict}{The State Key Laboratory of AI Safety, Institute of Computing Technology, Chinese Academy of Sciences, Beijing, 100190, China}
  \icmlaffiliation{ucas}{The University of Chinese Academy of Sciences, Beijing 100049, China}

  \icmlcorrespondingauthor{Hong Chang}{changhong@ict.ac.cn}

  \icmlkeywords{Machine Learning, ICML}

  \vskip 0.3in
]



\printAffiliationsAndNotice{}  

\begin{abstract}
  Out-of-Distribution (OOD) detection requires sensitivity to subtle shifts without overreacting to natural In-Distribution (ID) diversity. However, from the viewpoint of \emph{detection granularity}, global representation inevitably suppress local OOD cues, while patch-based methods are unstable due to entangled spurious-correlation and noise. And neither them is effective in detecting \emph{compositional} OODs composed of valid ID components. Inspired by recognition-by-components theory, we present a training-free \textbf{Component-Based OOD Detection (CoOD)} framework that addresses the existing limitations by decomposing inputs into functional \emph{components}. To instantiate CoOD, we derive \textbf{Component Shift Score (CSS)} to detect local appearance shifts, and \textbf{Compositional Consistency Score (CCS)} to identify cross-component compositional inconsistencies. Empirically, CoOD achieves consistent improvements on both coarse- and fine-grained OOD detection.
\end{abstract}

\section{Introduction}
Real-world computer vision systems must not only produce accurate predictions on in-distribution (ID) inputs, but also abstain on out-of-distribution (OOD) inputs to avoid risky downstream decisions. Previous work on OOD detection has explored multiple complementary directions, such as designing scoring functions to reduce model overconfidence \citep{energy,mcm,gap} and learning more discriminative representations to improve detection \citep{eval,infoood,idlike}. Orthogonal to these directions, \emph{detection granularity} introduces another axis, determined by the spatial level at which evidence is aggregated. From this viewpoint, OOD detection approaches can be broadly categorized into global or local ones.

\begin{figure}[t]
  \begin{center}
    \centerline{\includegraphics[width=\columnwidth]{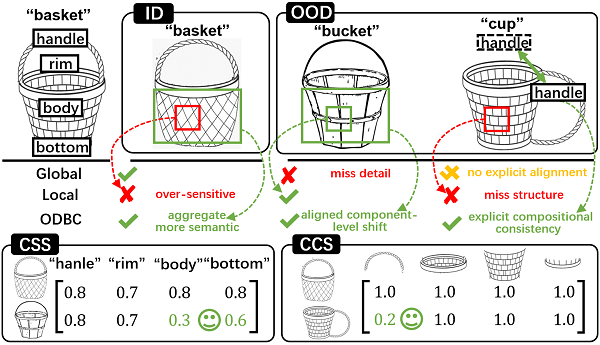}}
    \caption{Comparison between CoOD and global/local methods. 
    \textbf{Top:} From left to right are: training ID \textit{``basket''}, test ID \textit{``basket''}, fine-grained OOD \textit{``bucket''}, and compositional OOD \textit{``cup''}. 
    \textbf{Middle:} Causes of \green{success} or \red{failure}.
    \textbf{Bottom:} CSS exposes fine-grained shift in components \textit{``bucket's body''}, while the CCS detects a compositional misalignment in \textit{``cup's handle''}.}
    \vskip -0.3in
    \label{fig:motivation1}
  \end{center}
\end{figure}

\textit{Global approaches} \citep{energy,mcm,gap} aggregate features across the entire image into a holistic representation and generally perform well under large semantic shifts. However, by pooling or global features aggregation, these methods project inputs onto dominant class-semantic directions. Consequently, they smooth out, and thereby weaken, the localized deviations \citep{expsen} that are crucial for \textbf{\textit{detecting fine-grained OODs}}. As shown in \cref{fig:motivation1}, the global method overlooks the texture difference in the \emph{``body''} of \emph{``bucket''}, thus fails to detect it as OOD.

By contrast, \textit{local (patch-based) approaches} \citep{gl,locoop} capture fine-grained OOD by preserving spatially dense information \citep{fineoodclosesetood}. However, they are inherently prone to the \textbf{\textit{sensitivity-robustness dilemma}}: fine-grained evidence increases sensitivity to subtle OODs, but simultaneously exposes the detector to patch noise, cross-patch contamination \cite{pesudocorr}, and model bias \citep{ocs}, thereby undermining robustness.
Furthermore, simple post-hoc strategies that select salient embeddings \citep{mood} or local patches \citep{gl,locoop,topkgl} still suffer from semantic confusion and low signal-to-noise ratio, as illustrated in \cref{fig:motivation1}. 

Another important challenge concerns \textbf{\textit{detecting compositional OODs}} that are constructed from components highly similar to ID parts. Local schemes that evaluate tokens independently cannot, by design, determine whether the global context is inconsistent, since they lack an explicit model of inter-token compositional relations\footnote{This failure pattern is pervasive in practice, as semantic shifts are frequently accompanied by structural misalignments.}. 
As shown in \cref{fig:motivation1}, the OOD \emph{``cup''} exhibits similar local presence with ID \emph{``basket''}, therefore can hardly be detected. 
Surprisingly, global methods also fall short for this case: uneven feature-learning \citep{fastslow} and limited receptive fields \citep{vitrfissue} cause the model to rely on a small subset of salient evidence, while inherent modeling biases \citep{capsule,clipblind} further restrict its ability to encode compositional structure.

An ideal OOD detector should resolve the sensitivity-robustness dilemma -- tolerating reasonable ID variation and mild noise, while rejecting semantic or structural OODs at \emph{any} scale \citep{dmad}. For this purpose, we introduce \textbf{Component-Based OOD Detection (CoOD)}, a component-centric training-free paradigm, motivated by Recognition-by-Components (RBC) theory \citep{rbc}. CoOD explicitly decomposes instances into \textbf{functional components} that serve as \emph{the finest yet meaningful granularity} to identify local OODs while averaging away patch-level noise, \textit{i.e.} finer-grained aggregation adds little signal but more noise, whereas coarser pooling suppresses local OOD cues. On this basis, the relationship on components further supplies global semantic and geometric context that complements the component-level evidence.

Crucially, CoOD makes the best of the two complementary evidence streams. \textbf{Component Shift Score (CSS)} aggregates tokens assigned to the same component to construct robust component representations. By averaging intra-component variations, CSS suppresses patch-level noise and spurious contextual leakage, thereby preserving component-specific semantics and amplifying sensitivity to fine-grained OODs. \textbf{Compositional Consistency Score (CCS)} measures semantic and geometric consistency across components by matching component structure to a compact ID coreset and estimating a spatial alignment. Thus, large alignment residuals or low feature agreements indicate invalid component combinations. 
We further use a plug-and-play \emph{non-component suppression} step to reduce background and cross-component interference. As \cref{fig:motivation1} shows, CSS captures the \textit{``body''} component shift (\textit{w.r.t.} texture) between \textit{``basket''} and \textit{``bucket''}, while CCS flags a \textit{``handle''} structural inconsistency that identifies \textit{``cup''} as OOD. This combination yields interpretable, component-level evidence and a practical pathway to improve both coarse- and fine-grained OOD detection without retraining.

Our contributions are threefold:
\begin{itemize}
\setlength{\itemsep}{0pt}
\setlength{\parsep}{0pt}
\setlength{\parskip}{0pt}
\item We propose \textbf{CoOD}, a training-free, component-centric and interpretable OOD detector that reconciles fine-grained sensitivity with robustness to ID variations. 
\item We instantiate CoOD via a component-aware aggregation framework, deriving \textbf{C}omponent \textbf{S}hift \textbf{S}core for fine-grained OODs and complementary \textbf{C}ompositional \textbf{C}onsistency \textbf{S}core for compositional OODs.
\item We theoretically and empirically verify that component-level OOD evidence is highly effective, reducing FPR by $\sim$55\% on the fine-grained CUB benchmark.
\end{itemize}

\section{Related Works}
\textbf{Global OOD Detection.}  
A major line of OOD research focuses on globally aggregated representations, most commonly classifier outputs or pooled embeddings. Representative detectors include maximum softmax probability \citep{msp}, energy score \citep{energy,deltaenergy}, density measure \citep{knn}, recent logit refinements \citep{gap} and ensemble learning \citep{cook}. These methods perform well for large semantic shifts because global representation amplifies the dominant class signals.

With the advent of vision-language models and zero-shot setting \citep{clip}, CLIP-based detectors have stronger ability in measuring similarity to textual representation \citep{mcm}, or expanded pseudo-OOD sets \citep{eoe,neg,csp}. However, such detectors still fall short due to the holistic measure, which tends to suppress localized appearance discrepancies. To address this limitation, CoOD preserves meaningful localized signals via component-guided aggregation, retaining sensitivity to fine-grained OODs.

\textbf{Local OOD Detection.} 
To recover details suppressed by global aggregation, some recent works incorporate spatial locality into OOD detection, \textit{e.g.} computing dense (patch-level) similarity \citep{mode,gl,topkgl} or implicitly enhancing locality via shallow visual feature comparisons \citep{mood,dual}. 
Others use training-time regularization, \textit{e.g.}, treating background patches as negative signals \citep{background} for outlier exposure \citep{oe}, prompt tuning \citep{locoop,idlike}, or localized contrastive learning \citep{mode,topkgl}. 

Although these strategies increase local sensitivity, they either require additional training or exhibit limited stability, because patch representations remain entangled with spurious correlations \cite{pesudocorr}, model priors \citep{ocs}, and incidental noise. Crucially, they cannot capture structural relationships among local patches. In contrast, CoOD is designed to suppress irrelevant noise and obtain robust component semantics, addressing both patch-level instability and the lack of compositional awareness.

\textbf{Component-based Methods.}
From recognition-by-Components theory \cite{rbc}, robust vision lies in decomposing objects into components and reasoning over their compositions. This insight has inspired modern deep models to explicitly capture intermediate visual components. For example, \citet{ppnet,dpnet} extract discriminative components to support interpretable evidence-based classification. \citet{nppnet} further develops stable part discovery by non-parametric clustering. Beyond individual parts, explicit compositional modeling can also improve robustness under occlusion and semantic variation \citep{CompositionUnderOcclusion,rock}.

Although modern networks implicitly encode component cues, they are often entangled with global context and noise, limiting robustness under semantic variations and sensitivity to detect fine-grained and compositional OODs. 
Simply combining global and local scores is insufficient, as post-hoc score mixing is merely a compromise solution, rather than disentangling nuisance factors that conflate localized evidence with global biases and local noise. To our knowledge, CoOD is the first OOD detector that decomposes inputs into components and provides principled evaluation criteria for both fine-grained and compositional OODs.

\section{Methodology}
This section starts by the problem setup and our theoretical intuition. We then describe how to construct CoOD by component-level representations, CSS and CCS.

\begin{figure*}[t]
    \centerline{\includegraphics[width=\linewidth]{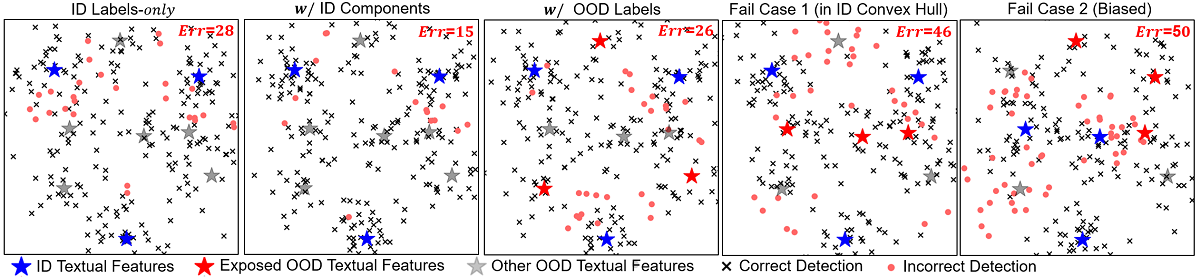}}
    \caption{Comparison between CoOD and pseudo-OOD methods on toy dataset, varing only in text selection. From left to right are: baseline (MCM), CoOD (components are drawn around the corresponding ID classes), ideal pseudo-OOD scenario (EOE), and two failure cases (EOE). $\red{Err}$ ($\red{\bullet}$) indicates the number of incorrect detection. Since CoOD has correct detection in the most of component perspectives, the aggregated score has lower variance and Err.}\vskip -0.2in
    \label{fig:compare2ece}
\end{figure*}

\subsection{Preliminaries}
Let $\mathcal{X}$ and $\mathcal{Y}\!=\!\{1,2,\ldots,C\}$ be the input space and semantic label space, with joint distribution $P_{\mathcal{X}\!\times\!\mathcal{Y}}$. During the test phase, an OOD detector observes only $\bm x\!\in\!\mathcal{X}$ and decides whether $\bm x$ comes from the ID marginal distribution $\bm x\!\sim\!P_{\mathcal{X}}$ or from OODs.

Given a visual encoder $f\!:\!\mathcal X\!\to\!\mathcal Z$, a test sample $\bm x$ is encoded as $\bm z\!=\!f(\bm x)$. We denote position representations collectively as $\bm e$ and the total number of spatial positions as $N$. 
For vision-language models with a text encoder $g$, the semantic space is defined by the encoded text $\bm t_y\!\in\!\{g(\text{``text}_y\text{''})\}_{y=1}^C$, where $\text{``text}_y\text{"}$ typically follows a prompt such as ``a photo of a $\langle class_y\rangle$''. Logit activation between a visual embedding $\bm z$ and text embedding $\bm t_y$ is given by cosine similarity $\langle\bm z, \bm t_y\rangle=\frac{\bm z^\top\bm t_y}{\Vert\bm z\Vert \Vert\bm t_y\Vert}$ and the posterior is:
\begin{equation}
\label{ptheta}
P_\theta(y\mid\bm x)=e^{\langle\bm z, \bm t_y\rangle}/{{\sum}_{y'\in\mathcal{Y}}e^{\langle\bm z, \bm t_{y'}\rangle}}.
\end{equation}

In this paper, each class $y$ is associated with a component set $\mathcal{P}_y$. For each $p\in \mathcal{P}_y$, the semantics are encoded as $\bm t_{yp}\in\!\{g(\text{``text}_{yp}\text{''})\}_{p=1}^{|\mathcal{P}_y|}$. With  estimated foreground mask $\bm m_{y}\!\in\![0,1]^{H\!\times\!W}$ and component masks $\bm m_{yp}\!\in\![0,1]^{H\!\times\!W}$, we compute two complementary evidences: component-shift score $\mathrm{CSS}$ and compositional-consistency score $\mathrm{CCS}$. Finally, the \emph{OOD score} is the weighted average:  
\begin{equation}
\label{eq:alpha}
\mathrm{CoOD}(\bm x)= \mathrm{CSS}(\bm x)+\alpha\mathrm{CCS}(\bm x).
\end{equation}

\subsection{Theoretical Motivation}

\begin{figure*}[t]
    \centerline{\includegraphics[width=.9\linewidth]{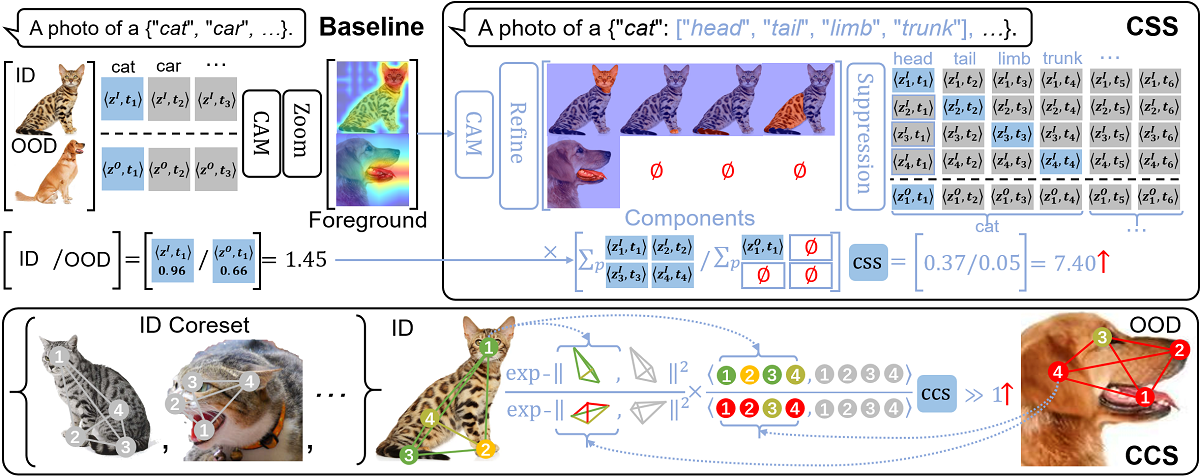}}
    \caption{Illustration of CoOD framework, which firstly represents class components, and then computes complementary \textbf{CSS} and \textbf{CCS}.}\vskip -0.2in
    \label{fig:cssccs}
\end{figure*}

As illustrated by \textit{“Foreground”} in \cref{fig:cssccs}, discriminative models often concentrate gradient mass on a small set of foreground components. Such models may change their confidence sharply in presence of ID variations, and miss fine-grained OOD shifts that occur beyond the focused regions. 
As will be analyzed theoretically in this section, CoOD can address these issues from three aspects: (1) explicitly includes diverse component representations to mitigate the loss of detection-relevant local information (\cref{theory num}), (2) reduces confusion from irrelevant context and noise (\cref{theory corr}), and (3) supplies compositional evidence of inter-component relationships \cite{rbc}.

\subsubsection{Introducing Component Representations}
\label{theory num}
Prior work \citep{infoood} shows that auxiliary information can improve OOD detection. Following widely used setup \citep{neg,csp,dual}, we define an \emph{existence score} for each component $p$ as $s_{yp}\!=\!\langle\bm z_{yp},\bm t_{yp}\rangle$ and model the  component presence as a Bernoulli random variable with parameter $\psi\!=\!P(\!s_{yp}\!>\!\tau|\bm x\!)$, which represents the probability of being an ID component. 
The aggregated ID score $\mathcal{S}\!=\!\sum_p\!\mathds{1}\{s_{yp}\!>\!\tau\}$ can be modeled by a Binomial distribution 
$\mathcal{B}(|\mathcal{P}_y|,\psi)$, and approximated\footnote{For small $|\mathcal{P}_y|$, we provide more accurate computation in Appendix~\ref{pbfpr}.} by Normal distribution as $\mathcal{N}(|\mathcal{P}_y|\psi, |\mathcal{P}_y|\psi(1-\psi))$. The false positive rate at true positive rate $\lambda$ is:
\begin{align}
\label{eqCoODfpr}
\text{FPR}_{\lambda} = \Phi\Big(\frac{\mu^\mathrm{out} - \mu^\mathrm{in}}{\sigma^\mathrm{out}} + \frac{\sigma^\mathrm{in}}{\sigma^\mathrm{out}} \Phi^{-1}(\lambda) \Big),
\end{align}
where $\Phi$ is monotonic standard normal CDF and $(\mu,\!\sigma^2\!)$ parameterize the Normal distribution on detection scores of ID and OOD. Differentiating $\text{FPR}_{\lambda}$ \textit{w.r.t.} $|\mathcal{P}_y|$ gives:
\begin{equation}
\label{deltap}
\frac{\partial \ \text{FPR}_{\lambda}}{\partial  \ |\mathcal{P}_y|}\ \propto \ \frac{\psi^\mathrm{out} - \psi^\mathrm{in}}{2\!\sqrt{|\mathcal{P}_y| \psi^\mathrm{out} (1\!-\!\psi^\mathrm{out})}}.
\end{equation}
Thus adding component-wise evidence ($|\mathcal{P}_y|\big\uparrow$) reduces FPR$_\lambda$ provided $\psi^{\mathrm{in}}$$>$$\psi^{\mathrm{out}}$. 
Some previous methods alternatively utilize additional information from pseudo-OOD \citep{eoe}, but their discriminability depends critically on the feature-space relation of ID and OOD, and may fail in practice, \textit{e.g.} pseudo OODs may fall inside the convex hull of IDs, producing opposing signals as shown in \cref{fig:compare2ece}. 

By contrast, we introduce components near ID textual representation to reinforce ID proximity and OOD discriminability. Real OODs tend to be distant from most ID components, whereas IDs remain close to at least some. By aggregating component evidence, CoOD suppresses irrelevant noise and prevents easy-to-learn \cite{simbias} or dominant features from attenuating other component signals, thereby enhancing the discrimination of fine-grained OOD.

\subsubsection{Reducing Nuisance Correlations}
\label{theory corr}
However, in practice, visual representations extracted from \emph{full} images often encode undesired correlations (\textit{e.g.}, backgrounds or spurious correlations), which could raise FPR.
To quantify this effect, we add a component ``$new$'' with discriminative existence probability ($\psi^{\mathrm{in}}\!>\!\psi^{\mathrm{out}}$). Denotes the argument of $\Phi(\cdot)$ in \cref{eqCoODfpr} as $A_{\mathrm{cor}}$ if the new component correlates to some existing ones and $A_{\mathrm{idp}}$ if not. The difference between the two values can be approximated by first-order Taylor expansion as:
\begin{align}
\label{taylorp}
&A_{\text{cor}}-A_{\text{idp}}\approx\frac{1}{\sigma^\mathrm{out}}{\Big (}
\frac{\mu^\mathrm{in} - \mu^\mathrm{out}}{(\sigma^\mathrm{out})^2}{\sum}_j\text{Cov}^\mathrm{out}_{new, j}\notag\\
&+\!\frac{\Phi^{\!-1\!}(\lambda)}{\sigma^\mathrm{in}}\!{\sum}_j\!\text{Cov}^\mathrm{in}_{\!new, j}\!-\!{\frac{\sigma^\mathrm{in}\Phi^{\!-1\!}(\lambda)}{(\sigma^\mathrm{out})^2}}\!{\sum}_j\!\text{Cov}^\mathrm{out}_{\!new, j}\!{\Big )}\!,\!
\end{align}
where $\text{Cov}^\mathrm{in}_{\!new, j}$ and $\text{Cov}^\mathrm{out}_{\!new, j}$ denote covariance  terms between the new component score and the $j^{th}\ (j\!=\!1,\!\ldots\!,|\mathcal{P}\!_y|)$ one under ID and OOD, respectively. If the new component score covaries non-negatively with existing ones, the \textit{first two terms} in \cref{taylorp} are positive. 
The third opposing term is typically much smaller in magnitude than the second. Because ID components tend to be consistently present (low $\sigma^{\mathrm{in}}$) and to co-occur (high $\text{Cov}^\mathrm{in}_{new, j}$) in ID images. By contrast, OODs generally exhibit greater diversity (high $\sigma^{\mathrm{out}}$) and weaker co-occurrence (low $\text{Cov}^\mathrm{out}_{new, j}$) in ID component presence. 
Consequently, the FPR is increased when non-negligible positive covariances exist. 

Accordingly, we reduce FPR by suppressing undesired component correlations, such as the spurious statistical dependencies introduced by cross‑region contamination. To this end, we design a text–image–feature tri‑level suppression mechanism in \cref{secCoODcidandrep}, which enforces semantic alignment between the final embedding and each component (see Fig.~\ref{fig:sup}). Empirically, this strategy mitigates cross‑region interference and yields more robust component‑based OOD detection.

\subsection{Component Identification and Representation}
\label{secCoODcidandrep}
In this section, CoOD obtains semantically meaningful components to facilitate subsequent detection.

\subsubsection{Constructing a Component Vocabulary}
\label{vlmtext}
A vocabulary of visually discriminative components, constructed automatically from coarse ID classes, can provide the semantic for component identification and representation. To minimize manual effort, we resort to LLMs to generate and filter component candidates\footnote{We also implement a purely visual construction method in \cref{ana} by deep clustering \citep{ace} over superpixels, serving as a variant of component vocabulary.}.

Concretely, we prompt an LLM to decompose each class into a \textbf{``taxonomic tree''} of \emph{visually structured component texts}, encouraging candidates to be both class-relevant and semantically complete. We then apply prompts to regularize the outputs (\textit{e.g.} remove CLIP-insensitive modifiers such as direction or position \citep{clipblind}, merge synonyms, and replace ambiguous or rare terms). This pipeline (see Appendix \ref{llm} for more details) aligns the extracted vocabulary with regions that a VLM is more likely to segment reliably, and thus helps ensure the discriminative component set $\mathcal{P}_y$ covers class $y$ while reducing noise from trivial components.

\subsubsection{Component Localization}
To reduce irrelevant global semantics and undesired nuisances, we suppress background in images and amplify true component signals without retraining. Specifically, CoOD computes a foreground mask $\bm m_{y}$ via CAM \citep{vitcam} and zoom-in input to foreground region as \cref{fig:cssccs} shown. Similarly, the candidate masks $\bm m^\prime_{yp}$ within $\bm m_{y}$ are obtained from CAM and compete with each other to reduce overlapping effects, producing a refined $\bm m_{yp}$:
\begin{equation}
\label{eq:maskcomp}
\bm m_{yp}=e^{\bm m^\prime_{yp}}/(e^{1-\bm m_{y}}+{\sum}_{p'} e^{\bm m^\prime_{yp'}}),
\end{equation}
where $1\!-\!\bm m_{y}$ introduces a background competitor to suppress noise. Further refinement \citep{crf} can also improve the mask quality. 
Then we obtain the suppressed component-level images:
\begin{equation}
\label{eqCoODpromptimage}
\bm x_{yp}=\bm x\odot\bm m_{yp}+\text{Blur}(\bm x) \odot(1-\bm m_{yp}),
\end{equation}
where $\text{Blur}(\cdot)$ denotes a Gaussian blurring operation.

\subsubsection{Component Representation}
Although \cref{eqCoODpromptimage} focuses on image-space components, the initial ViT class token lacks component-specific guidance and may contaminate component representations via attention. To mitigate this, we introduce a feature-level component guidance into class token and position embedding.  

\begin{figure}[t]
    \centerline{\includegraphics[width=1.\linewidth]{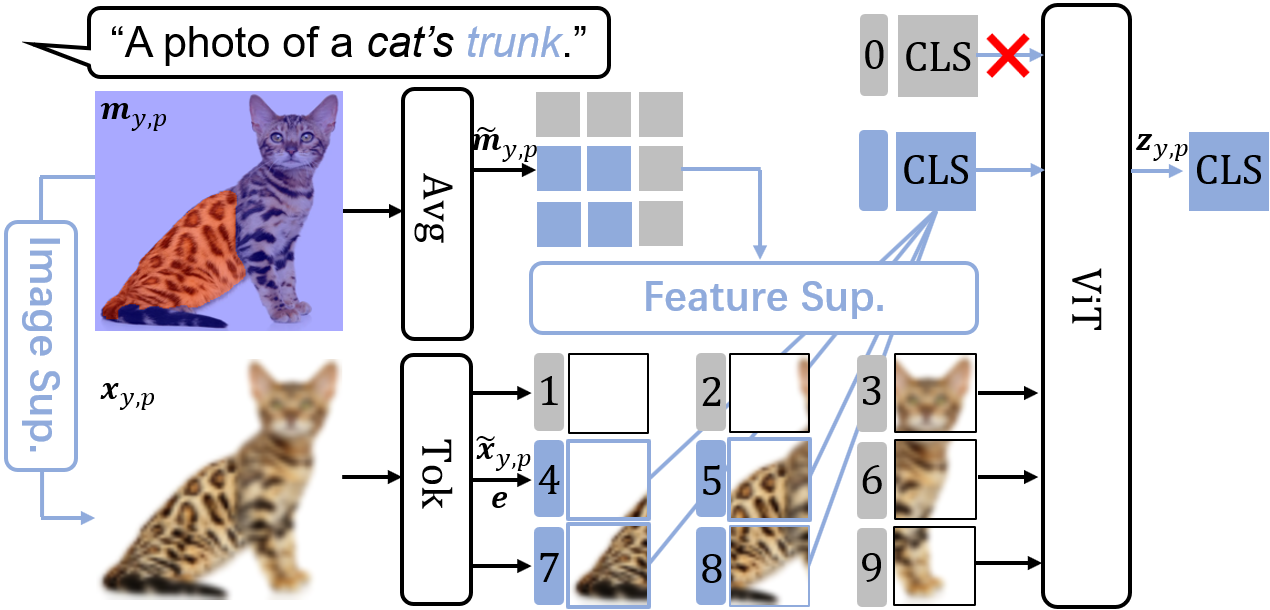}}
    \caption{Illustration of image-level and feature-level suppression.
    }\vskip -0.2in
    \label{fig:sup}
\end{figure}

As \cref{fig:sup} shown, for component $\bm x_{yp}$, the component mask is resized as $\bm m_{yp}\!\in\![0, 1]^{H\!\times\!W}\!\to\!\tilde{\bm m}_{yp}\!\in\![0, 1]^N$. 
Let $\tilde{\bm x}_{yp}\!=\!f_{\mathrm{Tok}}(\bm x_{yp})\!\in\!\mathbb{R}^{N\!\times\!D}$ be the matrix of $N$ patch tokens, and $\bm e\!\in\!\mathbb{R}^{N\times D}$ the position embedding. 
The original patch and position tokens are concatenated to the averaged ones (\textit{i.e.} $\frac{\tilde{\bm m}_{yp}^\top \tilde{\bm x}_{yp}}{\mathbf{1}^\top\tilde{\bm m}_{yp}}$ and $\frac{\tilde{\bm m}_{yp}^\top \bm e}{\mathbf{1}^\top\tilde{\bm m}_{yp}}$), respectively.
The component feature $\bm z_{yp}$ is then calculated as: 
 \begin{align}
 \label{eq:sup}
 \bm z_{yp}\!=\!f_{\mathrm{ViT}}\!{\bigg (}\!{\Big [}\!\frac{\tilde{\bm m}_{yp}^\top \tilde{\bm x}_{yp}}{\mathbf{1}^\top\tilde{\bm m}_{yp}}\!;\tilde{\bm x}_{yp}{\Big ]}\!,\!
 {\Big [}\!\frac{\tilde{\bm m}_{yp}^\top \bm e}{\mathbf{1}^\top\tilde{\bm m}_{yp}}\!;\bm e{\Big ]}\!{\bigg )}\!,
 \end{align}
which biases toward the target component and reduces contamination from irrelevant global semantics.

\subsection{Component Shift Score}
The OOD detection objective can be decomposed as aggregating component-level evidences for class $y$:
\begin{equation}
\label{eqCoODmodeling}
P(\mathrm{ID}\!\mid\!\bm {x})\!=\!{\sum}_{yp}\!P(\mathrm{ID}\!\mid\!\bm{x},yp)\;P(p\!\mid\!\bm{x},y)\;P(y\!\mid\!\bm{x}).
\end{equation}

From \citet{rbc}, once the observed components $\{p\}$ are estimated, the component-level signals explain the majority of class-relevant variation, while residual pixel-level signals provide relatively limited supplementary information.
This implies a generation graph $\bm x\!\to\!\mathcal{P}_y\!\to\! y,\{\text{ID},\text{OOD}\}$. 
Without extra prior, assume each component $p\in \mathcal{P}_y$ contributes equally to the ID evidence:
\begin{equation}
\label{eqCoOD2modelingmethod}
P(\mathrm{ID}|\bm x,\!y,\!p)\!\approx\! P(\mathrm{ID}| y,\!p)\!=\!
\left\{
\begin{array}{ll}
    \textstyle 1/|\mathcal{P}_y|, & p\!\in\! \mathcal{P}_y,\\
    \textstyle 0,  & p\!\notin\! \mathcal{P}_y.
\end{array}
\right.
\end{equation}

Substituting \cref{eqCoOD2modelingmethod} into \cref{eqCoODmodeling} and approximating the posteriors with a neural network $P_\theta(\cdot)$ as \cref{{ptheta}}, the Component Shift Score is:
\begin{equation}
\label{generalcss}
\mathrm{CSS}_y(\bm x)=P_\theta(y| \bm x) \frac{1}{|\mathcal{P}_y|}{\sum}_{p\in \mathcal{P}_y} P_\theta (p|\bm x_{yp}),
\end{equation}
where the network $P_\theta$ is shared for both predictions: component prediction applies suppression ($y$ is integrated into $\bm x_{yp}$) as in \cref{secCoODcidandrep} while class prediction does not. According to \citet{mcm}, CSS can be instantiated as:
\begin{equation}
\label{eq:css}
\mathrm{CSS}(\bm x)\!=\!\max_y\!\frac{e^{\langle \bm z, \bm t_y\rangle}}{{\sum}_{y'}e^{\langle \bm z, \bm t_{y'}\!\rangle}}\!
\!\sum_{p\in \mathcal{P}_y}\!\frac{e^{\langle \bm z_{yp}, \bm t_{yp}\rangle}}{|\mathcal{P}_y|{\sum_{p'}}e^{\langle \bm z_{yp}, \bm t_{yp'}\!\rangle}}\!.
\end{equation}

\subsection{Compositional Consistency Score}
\label{secccs}
To detect compositional or structural OODs constructed from ID components, we design component compositional consistency score ($\mathrm{CCS}$) which measures input variation as a global optimal component matching to ID samples. Given $N$ patch tokens $\{\!\bm z_n\!\}_{n\!=\!1}^{N}$, the component features $\{\!\bm z_{yp}\!\}_{p\!=\!1}^{|\mathcal{P}_y|}$ computed in \cref{eq:sup}, the corresponding masks $\{\tilde{\bm m}_{yp}\}_{p=1}^{|\mathcal{P}_y|}$, and let $(\tilde{\mathbf{m}}_{yp})_n$ denote the $n^{th}$ position of the mask, we select $\mathrm{Top}k$ patches within each component whose features are closest to the component representation $\bm z_{yp}$, yielding a set of patch indices:
\begin{equation}
\label{ccssampling}
\mathcal{I}_y(\bm x)={\Big\{}\underset{n\in\{i|(\tilde{\bm m}_{yp})_i>\tau\}}{\arg\text{Top}k}-\Vert \bm z_n-\bm z_{yp}\Vert_2, \forall p {\Big\}}.
\end{equation}
To constrain matching cost while covering ID diversity, we select a core set $\mathcal{D}^*$ from training set $\mathcal{D}$ via Farthest-Point Sampling \citep{fps}. 
For a test $\bm x$ and a $\bm x^\mathrm{r}$ from $\mathcal{D}^*$, 
we use Hungarian matching \citep{km} by one-to-one assignment over candidate pairs $\mathcal{I}_y(\bm x)\times \mathcal{I}_y(\bm x^{\mathrm r})$:
\begin{equation}
\mathcal{K}^*_y(\bm x,\bm x^{\mathrm r})
=\underset{\mathcal{K}\subseteq \mathcal{I}_y(\bm x)\times \mathcal{I}_y(\bm x^{\mathrm r})}{\arg\max}
\sum_{(i,j)\in\mathcal{K}} \langle\bm z_i,\bm z_j^{\mathrm r}\rangle.
\end{equation}
Then we estimate the corresponding spatial transform $\mathcal{M}$ (\textit{e.g.} Affine) on the paired locations, and select the optimal $\bm x^{\mathrm{r}*}$ with minimal Euclidean residual:
\begin{equation}
\label{ccsmatching}
\bm x^{\mathrm{r}*}=\underset{\bm x^\mathrm{r}\in \mathcal{D}^*_y}{\arg\text{min}}\frac{\sum_{(i,j)\in\mathcal{K}^*_y(\bm x, \bm x^\mathrm{r})}\Vert \mathcal{M}(\bm e_i)-\bm e_j\Vert_2}{|\mathcal{K}^*_y(\bm x, \bm x^\mathrm{r})|}.
\end{equation}
We then define CCS as the distance-weighted feature similarity over the optimal match:
\begin{equation}
\label{eq:ccs}
\mathrm{CCS}(\bm x)\!=\!\underset{y}{\text{max}}\frac{\sum_{(i,j)\in\mathcal{K}^*_y(\bm x, \bm x^{\mathrm{r}*})}\!e^{\!-\!\Vert \mathcal{M}(\!\bm e_i\!)\!-\!\bm e_j\!\Vert^2}\!\langle\!\bm z_i\!,\!\bm z_j\!\rangle}{|\mathcal{K}^*_y(\bm x, \bm x^{\mathrm{r}*})|}\!.
\end{equation}
The exponential distance decay increases sensitivity to geometric misalignment, while the feature similarity captures semantic agreement. Together, CCS reveals compositional and structural deviations that CSS alone cannot detect.

\begin{table*}[t]
\caption{OOD detection on ImageNet-1K and CUB with various OOD detectors and models. \textbf{Boldface} indicates the best performance. $^*$ indicates methods that require OOD labels (We use the original ID-only version of Dual and $\Delta E$ for a fair comparison with CoOD).}
\label{tab:1kcub}
\centering
{\scriptsize
\begin{tabular}{cccccccccccccccccccc}
    \toprule
    \makebox[0.005\textwidth][c]{$\mathcal{D}$} & \makebox[0.005\textwidth][c]{Metr.} & \makebox[0.04\textwidth][c]{Tr.} & \makebox[0.025\textwidth][c]{MLS} & \makebox[0.025\textwidth][c]{KL} & \makebox[0.025\textwidth][c]{EBO} & \makebox[0.025\textwidth][c]{ViM} & \makebox[0.025\textwidth][c]{ODIN} & \makebox[0.025\textwidth][c]{MCM} & \makebox[0.025\textwidth][c]{Gap} & \makebox[0.025\textwidth][c]{GL/R} & \makebox[0.025\textwidth][c]{EOE$^*$} & \makebox[0.025\textwidth][c]{NEG$^*$} & \makebox[0.025\textwidth][c]{Dual} & \makebox[0.025\textwidth][c]{CSP$^*$} & \makebox[0.025\textwidth][c]{COOK} & \makebox[0.025\textwidth][c]{$\Delta E$} & \makebox[0.025\textwidth][c]{CoOD-F} & \makebox[0.025\textwidth][c]{CoOD} & \makebox[0.025\textwidth][c]{CoOD$^*$}\\\cmidrule(lr){1-3}\cmidrule(lr){4-8}\cmidrule(lr){9-17}\cmidrule(lr){18-20}
    \makebox[0.005\textwidth][c]{\multirow{10}{*}{\rotatebox{90}{ImageNet-1k}}} & \makebox[0.005\textwidth][c]{\multirow{5}{*}{\rotatebox{90}{AUC}}} & CLIP & \makebox[0.005\textwidth][c]{84.6} & \makebox[0.005\textwidth][c]{74.9}& \makebox[0.005\textwidth][c]{81.5} & \makebox[0.005\textwidth][c]{61.9} & \makebox[0.005\textwidth][c]{90.4} & \makebox[0.005\textwidth][c]{90.2} & \makebox[0.005\textwidth][c]{89.3} & \makebox[0.005\textwidth][c]{91.0} & \makebox[0.005\textwidth][c]{93.3} & \makebox[0.005\textwidth][c]{94.2} & \makebox[0.005\textwidth][c]{90.9} & \makebox[0.005\textwidth][c]{95.7} & \makebox[0.005\textwidth][c]{90.8} & \makebox[0.005\textwidth][c]{90.3} & \makebox[0.005\textwidth][c]{95.4} & \makebox[0.005\textwidth][c]{96.2} & \makebox[0.005\textwidth][c]{\textbf{97.4}}\\
    & & CoOp & \makebox[0.005\textwidth][c]{84.8} & \makebox[0.005\textwidth][c]{72.9} & \makebox[0.005\textwidth][c]{81.6} & \makebox[0.005\textwidth][c]{63.2} & \makebox[0.005\textwidth][c]{90.5} & \makebox[0.005\textwidth][c]{91.9} & \makebox[0.005\textwidth][c]{88.1} & \makebox[0.005\textwidth][c]{92.6} & \makebox[0.005\textwidth][c]{92.3} & \makebox[0.005\textwidth][c]{91.0} & \makebox[0.005\textwidth][c]{85.4} & \makebox[0.005\textwidth][c]{92.6} & \makebox[0.005\textwidth][c]{89.8} & \makebox[0.005\textwidth][c]{86.3} & \makebox[0.005\textwidth][c]{96.2} & \makebox[0.005\textwidth][c]{96.7}& \makebox[0.005\textwidth][c]{\textbf{97.6}}\\
    & & \makebox[0.05\textwidth][c]{LoCoOp} & \makebox[0.005\textwidth][c]{79.8} & \makebox[0.005\textwidth][c]{75.6} & \makebox[0.005\textwidth][c]{73.2} & \makebox[0.005\textwidth][c]{60.4} & \makebox[0.005\textwidth][c]{90.5} & \makebox[0.005\textwidth][c]{92.7} & \makebox[0.005\textwidth][c]{89.0} & \makebox[0.005\textwidth][c]{93.6} & \makebox[0.005\textwidth][c]{93.2} & \makebox[0.005\textwidth][c]{86.4} & \makebox[0.005\textwidth][c]{86.5} & \makebox[0.005\textwidth][c]{87.9} & \makebox[0.005\textwidth][c]{90.0} & \makebox[0.005\textwidth][c]{88.1} & \makebox[0.005\textwidth][c]{96.3} & \makebox[0.005\textwidth][c]{96.9}& \makebox[0.005\textwidth][c]{\textbf{97.8}}\\
    & & IDLike &\makebox[0.005\textwidth][c]{83.0} & \makebox[0.005\textwidth][c]{78.6} & \makebox[0.005\textwidth][c]{81.1} & \makebox[0.005\textwidth][c]{70.3} & \makebox[0.005\textwidth][c]{90.3} & \makebox[0.005\textwidth][c]{91.2} & \makebox[0.005\textwidth][c]{90.5} & \makebox[0.005\textwidth][c]{92.7} & \makebox[0.005\textwidth][c]{89.6} & \makebox[0.005\textwidth][c]{86.2} & \makebox[0.005\textwidth][c]{89.5} & \makebox[0.005\textwidth][c]{88.8} & \makebox[0.005\textwidth][c]{91.7} & \makebox[0.005\textwidth][c]{90.4} & \makebox[0.005\textwidth][c]{96.3} & \makebox[0.005\textwidth][c]{96.8}& \makebox[0.005\textwidth][c]{\textbf{96.9}} \\
    & & LoPro  &\makebox[0.005\textwidth][c]{84.5}&\makebox[0.005\textwidth][c]{74.6}&\makebox[0.005\textwidth][c]{81.7}&\makebox[0.005\textwidth][c]{64.5}&\makebox[0.005\textwidth][c]{90.4}&\makebox[0.005\textwidth][c]{92.6}&\makebox[0.005\textwidth][c]{89.4}&\makebox[0.005\textwidth][c]{93.6}&\makebox[0.005\textwidth][c]{93.4}&\makebox[0.005\textwidth][c]{89.8}&\makebox[0.005\textwidth][c]{90.0} & \makebox[0.005\textwidth][c]{93.2} & \makebox[0.005\textwidth][c]{90.9} & \makebox[0.005\textwidth][c]{90.4} & \makebox[0.005\textwidth][c]{95.4}&\makebox[0.005\textwidth][c]{96.2} &\makebox[0.005\textwidth][c]{\textbf{97.4}}\\\cmidrule(lr){2-3}\cmidrule(lr){4-8}\cmidrule(lr){9-17}\cmidrule(lr){18-20}
    & \makebox[0.005\textwidth][c]{\multirow{5}{*}{\rotatebox{90}
    {FPR$_{95}$}}} & CLIP & \makebox[0.005\textwidth][c]{46.7} & \makebox[0.005\textwidth][c]{66.2} & \makebox[0.005\textwidth][c]{49.2} & \makebox[0.005\textwidth][c]{69.4} & \makebox[0.005\textwidth][c]{41.4} & \makebox[0.005\textwidth][c]{37.6} & \makebox[0.005\textwidth][c]{40.1} & \makebox[0.005\textwidth][c]{35.7} & \makebox[0.005\textwidth][c]{26.2} & \makebox[0.005\textwidth][c]{25.4} & \makebox[0.005\textwidth][c]{39.5} & \makebox[0.005\textwidth][c]{22.3} & \makebox[0.005\textwidth][c]{35.9} & \makebox[0.005\textwidth][c]{35.6} & \makebox[0.005\textwidth][c]{17.9}& \makebox[0.005\textwidth][c]{18.6} & \makebox[0.005\textwidth][c]{\textbf{11.8}}\\
    & & CoOp & \makebox[0.005\textwidth][c]{53.8} & \makebox[0.005\textwidth][c]{67.8} & \makebox[0.005\textwidth][c]{57.1} & \makebox[0.005\textwidth][c]{70.5} & \makebox[0.005\textwidth][c]{41.4} & \makebox[0.005\textwidth][c]{36.1} & \makebox[0.005\textwidth][c]{46.0} & \makebox[0.005\textwidth][c]{30.0} & \makebox[0.005\textwidth][c]{34.6} & \makebox[0.005\textwidth][c]{40.2} & \makebox[0.005\textwidth][c]{53.4} & \makebox[0.005\textwidth][c]{34.3} & \makebox[0.005\textwidth][c]{41.0} & \makebox[0.005\textwidth][c]{50.3} & \makebox[0.005\textwidth][c]{14.2} & \makebox[0.005\textwidth][c]{16.0}& \makebox[0.005\textwidth][c]{\textbf{10.7}}\\
    & & \makebox[0.05\textwidth][c]{LoCoOp} & \makebox[0.005\textwidth][c]{63.6} & \makebox[0.005\textwidth][c]{64.5} & \makebox[0.005\textwidth][c]{69.8} & \makebox[0.005\textwidth][c]{77.9} & \makebox[0.005\textwidth][c]{41.4} & \makebox[0.005\textwidth][c]{34.0} & \makebox[0.005\textwidth][c]{44.5} & \makebox[0.005\textwidth][c]{28.7} & \makebox[0.005\textwidth][c]{31.0} & \makebox[0.005\textwidth][c]{50.9} & \makebox[0.005\textwidth][c]{52.0} & \makebox[0.005\textwidth][c]{47.5} & \makebox[0.005\textwidth][c]{40.5} & \makebox[0.005\textwidth][c]{47.7} & \makebox[0.005\textwidth][c]{14.0} & \makebox[0.005\textwidth][c]{15.8}& \makebox[0.005\textwidth][c]{\textbf{10.1}}\\
    & & IDLike &\makebox[0.005\textwidth][c]{52.6} & \makebox[0.005\textwidth][c]{60.8} & \makebox[0.005\textwidth][c]{54.3} & \makebox[0.005\textwidth][c]{64.0} & \makebox[0.005\textwidth][c]{41.2} & \makebox[0.005\textwidth][c]{37.7} & \makebox[0.005\textwidth][c]{40.9} & \makebox[0.005\textwidth][c]{38.0} & \makebox[0.005\textwidth][c]{38.2} & \makebox[0.005\textwidth][c]{42.4} & \makebox[0.005\textwidth][c]{43.5} & \makebox[0.005\textwidth][c]{36.6} & \makebox[0.005\textwidth][c]{35.9} & \makebox[0.005\textwidth][c]{38.9} & \makebox[0.005\textwidth][c]{14.0} & \makebox[0.005\textwidth][c]{15.5}& \makebox[0.005\textwidth][c]{\textbf{12.1}} \\
    & & LoPro  &\makebox[0.005\textwidth][c]{47.0}&\makebox[0.005\textwidth][c]{70.7}&\makebox[0.005\textwidth][c]{49.2}&\makebox[0.005\textwidth][c]{67.1}&\makebox[0.005\textwidth][c]{41.4}&\makebox[0.005\textwidth][c]{37.3}&\makebox[0.005\textwidth][c]{39.7}&\makebox[0.005\textwidth][c]{28.3}&\makebox[0.005\textwidth][c]{25.7}&\makebox[0.005\textwidth][c]{32.2}&\makebox[0.005\textwidth][c]{38.6} & \makebox[0.005\textwidth][c]{26.7} & \makebox[0.005\textwidth][c]{35.6} & \makebox[0.005\textwidth][c]{35.6} & \makebox[0.005\textwidth][c]{17.7}& \makebox[0.005\textwidth][c]{18.7}&\makebox[0.005\textwidth][c]{\textbf{11.8}}\\
    \midrule
    \makebox[0.005\textwidth][c]{\multirow{10}{*}{\rotatebox{90}{CUB}}} & \makebox[0.005\textwidth][c]{\multirow{5}{*}{\rotatebox{90}{AUC}}} & CLIP & \makebox[0.005\textwidth][c]{77.5} & \makebox[0.005\textwidth][c]{65.2} & \makebox[0.005\textwidth][c]{76.9} & \makebox[0.005\textwidth][c]{63.4} & \makebox[0.005\textwidth][c]{74.2} & \makebox[0.005\textwidth][c]{75.4} & \makebox[0.005\textwidth][c]{67.5} & \makebox[0.005\textwidth][c]{73.5} & \makebox[0.005\textwidth][c]{77.8} & \makebox[0.005\textwidth][c]{76.3} & \makebox[0.005\textwidth][c]{76.4} & \makebox[0.005\textwidth][c]{76.5} & \makebox[0.005\textwidth][c]{75.6} & \makebox[0.005\textwidth][c]{76.8} & \makebox[0.005\textwidth][c]{92.1} & \makebox[0.005\textwidth][c]{95.2} & \makebox[0.005\textwidth][c]{\textbf{96.4}}\\
    & & CoOp & \makebox[0.005\textwidth][c]{81.9} & \makebox[0.005\textwidth][c]{63.1} & \makebox[0.005\textwidth][c]{81.5} & \makebox[0.005\textwidth][c]{65.8} & \makebox[0.005\textwidth][c]{78.1} & \makebox[0.005\textwidth][c]{80.4} & \makebox[0.005\textwidth][c]{78.8} & \makebox[0.005\textwidth][c]{78.2} & \makebox[0.005\textwidth][c]{82.1} & \makebox[0.005\textwidth][c]{78.1} & \makebox[0.005\textwidth][c]{81.3} & \makebox[0.005\textwidth][c]{78.3} & \makebox[0.005\textwidth][c]{79.4} & \makebox[0.005\textwidth][c]{76.8} & \makebox[0.005\textwidth][c]{96.4} & \makebox[0.005\textwidth][c]{98.1} & \makebox[0.005\textwidth][c]{\textbf{98.5}}\\
    & & \makebox[0.05\textwidth][c]{LoCoOp} & \makebox[0.005\textwidth][c]{78.3} & \makebox[0.005\textwidth][c]{60.0} & \makebox[0.005\textwidth][c]{76.1} & \makebox[0.005\textwidth][c]{64.1} & \makebox[0.005\textwidth][c]{79.2} & \makebox[0.005\textwidth][c]{80.7} & \makebox[0.005\textwidth][c]{80.5} & \makebox[0.005\textwidth][c]{78.3} & \makebox[0.005\textwidth][c]{82.5} & \makebox[0.005\textwidth][c]{74.2} & \makebox[0.005\textwidth][c]{82.0} & \makebox[0.005\textwidth][c]{79.3} & \makebox[0.005\textwidth][c]{79.3} & \makebox[0.005\textwidth][c]{76.8} & \makebox[0.005\textwidth][c]{97.2} & \makebox[0.005\textwidth][c]{97.9} & \makebox[0.005\textwidth][c]{\textbf{98.5}}\\
    & & IDLike &\makebox[0.005\textwidth][c]{80.9} & \makebox[0.005\textwidth][c]{65.3} & \makebox[0.005\textwidth][c]{80.5} & \makebox[0.005\textwidth][c]{65.7} & \makebox[0.005\textwidth][c]{75.2} & \makebox[0.005\textwidth][c]{76.2} & \makebox[0.005\textwidth][c]{70.2} & \makebox[0.005\textwidth][c]{75.5} & \makebox[0.005\textwidth][c]{73.9} & \makebox[0.005\textwidth][c]{79.0} & \makebox[0.005\textwidth][c]{79.2} &\makebox[0.005\textwidth][c]{80.8} & \makebox[0.005\textwidth][c]{75.3} & \makebox[0.005\textwidth][c]{76.6} & \makebox[0.005\textwidth][c]{94.5} & \makebox[0.005\textwidth][c]{96.2} & \makebox[0.005\textwidth][c]{\textbf{97.0}} \\
    & & LoPro  &\makebox[0.005\textwidth][c]{77.0}&\makebox[0.005\textwidth][c]{65.3}&\makebox[0.005\textwidth][c]{76.3}&\makebox[0.005\textwidth][c]{63.4}&\makebox[0.005\textwidth][c]{74.5}&\makebox[0.005\textwidth][c]{75.2}&\makebox[0.005\textwidth][c]{71.9}&\makebox[0.005\textwidth][c]{75.9}&\makebox[0.005\textwidth][c]{77.5}&\makebox[0.005\textwidth][c]{75.5}&\makebox[0.005\textwidth][c]{76.1} & \makebox[0.005\textwidth][c]{76.1} & \makebox[0.005\textwidth][c]{75.2} & \makebox[0.005\textwidth][c]{74.3} & \makebox[0.005\textwidth][c]{91.9}& \makebox[0.005\textwidth][c]{95.2} &\makebox[0.005\textwidth][c]{\textbf{96.4}}\\\cmidrule(lr){2-3}\cmidrule(lr){4-8}\cmidrule(lr){9-17}\cmidrule(lr){18-20}
    & \makebox[0.005\textwidth][c]{\multirow{5}{*}{\rotatebox{90}{FPR$_{95}$}}} & CLIP & \makebox[0.005\textwidth][c]{71.2} & \makebox[0.005\textwidth][c]{81.6} & \makebox[0.005\textwidth][c]{71.3} & \makebox[0.005\textwidth][c]{83.8} & \makebox[0.005\textwidth][c]{76.4} & \makebox[0.005\textwidth][c]{73.0} & \makebox[0.005\textwidth][c]{74.0} & \makebox[0.005\textwidth][c]{76.3} & \makebox[0.005\textwidth][c]{73.0} & \makebox[0.005\textwidth][c]{74.8} & \makebox[0.005\textwidth][c]{70.5} & \makebox[0.005\textwidth][c]{83.5} & \makebox[0.005\textwidth][c]{72.7} & \makebox[0.005\textwidth][c]{76.2} & \makebox[0.005\textwidth][c]{31.7} & \makebox[0.005\textwidth][c]{22.7} & \makebox[0.005\textwidth][c]{\textbf{18.4}}\\
    & & CoOp & \makebox[0.005\textwidth][c]{65.8} & \makebox[0.005\textwidth][c]{86.7} & \makebox[0.005\textwidth][c]{66.7} & \makebox[0.005\textwidth][c]{78.5} & \makebox[0.005\textwidth][c]{73.0} & \makebox[0.005\textwidth][c]{67.8} & \makebox[0.005\textwidth][c]{66.6} & \makebox[0.005\textwidth][c]{73.4} & \makebox[0.005\textwidth][c]{67.3} & \makebox[0.005\textwidth][c]{72.2} & \makebox[0.005\textwidth][c]{66.7} & \makebox[0.005\textwidth][c]{68.4} & \makebox[0.005\textwidth][c]{69.1} & \makebox[0.005\textwidth][c]{74.0} & \makebox[0.005\textwidth][c]{19.0} & \makebox[0.005\textwidth][c]{10.7} & \makebox[0.005\textwidth][c]{\textbf{8.0}}\\
    & & \makebox[0.05\textwidth][c]{LoCoOp} & \makebox[0.005\textwidth][c]{68.0} & \makebox[0.005\textwidth][c]{88.1} & \makebox[0.005\textwidth][c]{73.1} & \makebox[0.005\textwidth][c]{82.1} & \makebox[0.005\textwidth][c]{67.3} & \makebox[0.005\textwidth][c]{64.6} & \makebox[0.005\textwidth][c]{63.1} & \makebox[0.005\textwidth][c]{67.2} & \makebox[0.005\textwidth][c]{62.1} & \makebox[0.005\textwidth][c]{76.4} & \makebox[0.005\textwidth][c]{62.1} & \makebox[0.005\textwidth][c]{67.6} & \makebox[0.005\textwidth][c]{66.3} & \makebox[0.005\textwidth][c]{74.0} & \makebox[0.005\textwidth][c]{14.1} & \makebox[0.005\textwidth][c]{10.4} & \makebox[0.005\textwidth][c]{\textbf{6.8}}\\
    & & IDLike &\makebox[0.005\textwidth][c]{66.5} & \makebox[0.005\textwidth][c]{81.4} & \makebox[0.005\textwidth][c]{68.6} & \makebox[0.005\textwidth][c]{80.6} & \makebox[0.005\textwidth][c]{76.6} & \makebox[0.005\textwidth][c]{73.9} & \makebox[0.005\textwidth][c]{70.5} & \makebox[0.005\textwidth][c]{74.9} & \makebox[0.005\textwidth][c]{77.2} & \makebox[0.005\textwidth][c]{76.4} & \makebox[0.005\textwidth][c]{69.4} & \makebox[0.005\textwidth][c]{67.8} & \makebox[0.005\textwidth][c]{74.2} & \makebox[0.005\textwidth][c]{73.7} & \makebox[0.005\textwidth][c]{29.4} & \makebox[0.005\textwidth][c]{22.7} & \makebox[0.005\textwidth][c]{\textbf{20.1}} \\
    & & LoPro  &\makebox[0.005\textwidth][c]{71.3}&\makebox[0.005\textwidth][c]{82.0}&\makebox[0.005\textwidth][c]{73.0}&\makebox[0.005\textwidth][c]{82.7}&\makebox[0.005\textwidth][c]{76.2}&\makebox[0.005\textwidth][c]{73.9}&\makebox[0.005\textwidth][c]{76.7}&\makebox[0.005\textwidth][c]{72.0}&\makebox[0.005\textwidth][c]{70.5}&\makebox[0.005\textwidth][c]{75.7}&\makebox[0.005\textwidth][c]{72.3} & \makebox[0.005\textwidth][c]{85.0} & \makebox[0.005\textwidth][c]{74.2} & \makebox[0.005\textwidth][c]{77.1} & \makebox[0.005\textwidth][c]{32.6}& \makebox[0.005\textwidth][c]{22.7}&\makebox[0.005\textwidth][c]{\textbf{18.4}}\\
    \bottomrule
  \end{tabular}}
\end{table*}

\begin{table*}[t]
\caption{Compositional OOD detection on ImageNet-splits and generated counterfactual dataset with various OOD detectors.}
\label{tab: counterfactual}
\centering
{\scriptsize
\begin{tabular}{ccccccccccccccccccc}
    \toprule
    \makebox[0.035\textwidth][c]{Metr.} & \makebox[0.055\textwidth][c]{$\mathcal{D}$} & \makebox[0.025\textwidth][c]{MLS} & \makebox[0.025\textwidth][c]{KL} & \makebox[0.025\textwidth][c]{EBO} & \makebox[0.025\textwidth][c]{ViM} & \makebox[0.025\textwidth][c]{ODIN} & \makebox[0.025\textwidth][c]{MCM} & \makebox[0.025\textwidth][c]{Gap} & \makebox[0.025\textwidth][c]{GL} & \makebox[0.025\textwidth][c]{EOE$^*$} & \makebox[0.025\textwidth][c]{NEG$^*$} & \makebox[0.025\textwidth][c]{Dual} & \makebox[0.025\textwidth][c]{CSP$^*$} & \makebox[0.025\textwidth][c]{COOK} & \makebox[0.025\textwidth][c]{$\Delta E$} & \makebox[0.025\textwidth][c]{CoOD-F} & \makebox[0.025\textwidth][c]{CoOD} & \makebox[0.025\textwidth][c]{CoOD$^*$}\\\cmidrule(lr){1-2}\cmidrule(lr){3-7}\cmidrule(lr){8-16}\cmidrule(lr){17-19}
    \makebox[0.005\textwidth][c]{\multirow{2}{*}{AUC}} & \makebox[0.005\textwidth][c]{ImageNet-Split} & \makebox[0.005\textwidth][c]{68.3} & \makebox[0.005\textwidth][c]{51.4} & \makebox[0.005\textwidth][c]{65.5} & \makebox[0.005\textwidth][c]{55.6} & \makebox[0.005\textwidth][c]{68.0} & \makebox[0.005\textwidth][c]{68.8} & \makebox[0.005\textwidth][c]{69.9} & \makebox[0.005\textwidth][c]{69.7} & \makebox[0.005\textwidth][c]{68.6} & \makebox[0.005\textwidth][c]{68.4} & \makebox[0.005\textwidth][c]{70.0} & \makebox[0.005\textwidth][c]{68.4} & \makebox[0.005\textwidth][c]{56.3} & \makebox[0.005\textwidth][c]{68.4} & \makebox[0.005\textwidth][c]{64.6} & \makebox[0.005\textwidth][c]{75.4} & \makebox[0.005\textwidth][c]{\textbf{76.0}}\\    
    & \makebox[0.005\textwidth][c]{Counterfact} & \makebox[0.005\textwidth][c]{59.6} & \makebox[0.005\textwidth][c]{39.7} & \makebox[0.005\textwidth][c]{59.1} & \makebox[0.005\textwidth][c]{76.3} & \makebox[0.005\textwidth][c]{52.5} & \makebox[0.005\textwidth][c]{54.5} & \makebox[0.005\textwidth][c]{57.9} & \makebox[0.005\textwidth][c]{54.6} & \makebox[0.005\textwidth][c]{51.3} & \makebox[0.005\textwidth][c]{55.6} & \makebox[0.005\textwidth][c]{57.5} & \makebox[0.005\textwidth][c]{55.7} & \makebox[0.005\textwidth][c]{58.3} & \makebox[0.005\textwidth][c]{50.3} & \makebox[0.005\textwidth][c]{81.1}& \makebox[0.005\textwidth][c]{\textbf{94.2}}& \makebox[0.005\textwidth][c]{93.8}\\
    \midrule
    \makebox[0.005\textwidth][c]{\multirow{2}{*}{FPR}} & \makebox[0.005\textwidth][c]{ImageNet-Split} & \makebox[0.005\textwidth][c]{78.7} & \makebox[0.005\textwidth][c]{94.0} & \makebox[0.005\textwidth][c]{82.1} & \makebox[0.005\textwidth][c]{89.3} & \makebox[0.005\textwidth][c]{86.2} & \makebox[0.005\textwidth][c]{78.0} & \makebox[0.005\textwidth][c]{76.4} & \makebox[0.005\textwidth][c]{77.5} & \makebox[0.005\textwidth][c]{74.9} & \makebox[0.005\textwidth][c]{75.5} & \makebox[0.005\textwidth][c]{78.7} & \makebox[0.005\textwidth][c]{76.3} & \makebox[0.005\textwidth][c]{88.1} & \makebox[0.005\textwidth][c]{75.8} & \makebox[0.005\textwidth][c]{83.4}& \makebox[0.005\textwidth][c]{66.7} & \makebox[0.005\textwidth][c]{\textbf{69.1}}\\
    & \makebox[0.005\textwidth][c]{Counterfact} & \makebox[0.005\textwidth][c]{92.0} & \makebox[0.005\textwidth][c]{90.2} & \makebox[0.005\textwidth][c]{93.8} & \makebox[0.005\textwidth][c]{55.4} & \makebox[0.005\textwidth][c]{94.6} & \makebox[0.005\textwidth][c]{95.5} & \makebox[0.005\textwidth][c]{92.9} & \makebox[0.005\textwidth][c]{94.5} & \makebox[0.005\textwidth][c]{99.1} & \makebox[0.005\textwidth][c]{92.9} & \makebox[0.005\textwidth][c]{93.5} & \makebox[0.005\textwidth][c]{92.0} & \makebox[0.005\textwidth][c]{81.3} & \makebox[0.005\textwidth][c]{83.9} &\makebox[0.005\textwidth][c]{58.0}& \makebox[0.005\textwidth][c]{\textbf{46.4}} & \makebox[0.005\textwidth][c]{47.4} \\
    \bottomrule
  \end{tabular}}
\end{table*}

\section{Experiments}

\subsection{Experiment Setup}
\textbf{Datasets.}
We evaluate CoOD under four settings: 
\textbf{(i) ImageNet-1K} \citep{in} in the CLIP protocol, where SUN \citep{sun}, Places \citep{pla}, iNaturalist \citep{inat}, and DTD \citep{dtd} are used as OOD datasets; 
\textbf{(ii) Fine-grained CUB} \citep{cub}, where half of the subclasses within each superclass are used as ID and the remaining subclasses, together with classes without a superclass, are treated as OOD; 
\textbf{(iii) Covariate-IDs on ObjectNet}, which uses the 104-class intersection of ImageNet and ObjectNet \citep{on} to test the sensitivity of CCS to geometric variations (\textit{e.g.}, viewpoint, rotation, and scale), as CCS relies on spatial structure. In this setting, ObjectNet serves as covariate-ID (\textit{a.k.a.} close-set OOD \citep{deltaenergy}), while the real OOD (\textit{a.k.a.} open-set OOD) sets are the same as in \textbf{(i)}; 
\textbf{(iv) Compositional OOD on counterfactual data and ImageNet splits}, which isolate compositional changes via two complementary constructions: \textbf{(iv.a)} generating counterfactual samples using a diffusion model (FLUX.2-9B) with prompt \textit{``A surreal industrial design prototype of a $\langle class\rangle$, where $\langle component\rangle$ mounted $\langle impossible\ position\rangle$.''}; and \textbf{(iv.b)} manually regrouping ImageNet classes that share similar components but exhibit compositional inconsistencies. Note that we use the same preprocessing and balance sampling for fairness.

Additional variants include ImageNet-OpenOOD \cite{gap} for \textbf{setup-i}, which identifies near-OOD NINCO, ImageNet-O, and ImageNetOOD from ImageNet1K; ImageNet/ObjectNet-54, which uses 54 classes with clearly separable components to validate component-specific effectiveness for \textbf{setup-i/iii}; and ImageNet-R / -A / -S / -V2 for \textbf{setup-iii}, which use the overlapping 314 classes as covariate-ID and the remaining 686 ImageNet classes as OOD, following \citet{deltaenergy}. More details are provided in Appendix~\ref{moreexp}, \ref{compositional ood dataset} and \ref{llm}.

\textbf{Compared Methods.}
We focus primarily on CLIP-based methods: zero-shot inference (CLIP) \citep{mcm}, prompt learning (CoOp) \citep{coop}, local regularization (LoCoOp) \citep{locoop}, ID-like prompt (IDLike) \citep{idlike}, and local prompt (LoPro) \citep{topkgl}. The detectors are detached from training tricks, including output space methods: MaxLogit (MLS) \citep{maxlogit}, energy (EBO) \citep{energy}, Kullback–Leibler divergence (KL) \citep{maxlogit}, dominant-component score (ViM) \citep{vim} and input perturbation (ODIN) \citep{odin}; and CLIP-based methods: maximum concept matching (MCM) \citep{mcm}, local enhancement (GL) \citep{gl} / (R) \citep{topkgl}, energy change ($\Delta E$) \citep{deltaenergy} ensemble learning (COOK) \citep{cook}, visual similarity (Dual) \citep{dual}, OOD label grouping/exposure/semantic-pool (NEG/EOE/CSP) \citep{neg,eoe,csp}, and logit aggregation (Gap) \citep{gap}.

\vspace{-2pt}Moreover, we design a variant CoOD$^*$ that uses same pseudo-OOD texts as \citet{eoe} and a faster version of CoOD-F: $\smash[t]{\underset{y}{\max}\!\frac{e^{\langle \bm z, \bm t_{y}\rangle}}{{\sum}_{y'}e^{\langle \bm z, \bm t_{y'}\rangle}}\!\underset{p\in \mathcal{P}_y}{\sum}\frac{e^{\langle \bm z, \bm t_{p}\rangle}}{|\mathcal{P}_y|{\sum}_{p'}e^{\langle \bm z, \bm t_{p'}\rangle}}}$, that uses the global feature $z$ and same components.

\textbf{Implementation Details.}
We generate the components text by ChatGPT-4o with prompts shown in Appendix \ref{llm}, encode features by CLIP ViT-B/16, and obtain visual components by GradCAM \citep{vitcam} to keep CoOD post-hoc without extra supervision. 
CoOD uses 1\%$|\mathcal{D}|$ coreset, $k$=4 keypoints to enable affine estimation, binarization $\tau$=0.5, and selects the weight $\alpha$ on a validation set or pseudo OODs generated by rotation and blurring (empirical values are discussed in \cref{ana}). 
Most methods that require fine-tuning follow the original papers \citep{locoop} without additional regularization (50 epoch, 0.002 learning rate, 32 batch size, 1 warm-up epoch, 20-shot), while IDlike and LoPro follow their best setting. We report macro‑average of area under the receiver operating characteristic curve (AUC) and false positive rate at 95\% recall (FPR).

\begin{table*}[t]
\caption{Ablation study of the weight factor $\alpha$ in \cref{eq:alpha}. The results of CoOD / CoOD$^*$ are reported for each setting.}
\label{tab:alpha}
\centering
{\scriptsize
\begin{tabular}{ccccccccccc}
    \toprule
    \makebox[0.07\textwidth][c]{\multirow{2}{*}{$\alpha$}}&\multicolumn{5}{c}{Coarse-Grained OOD}&\multicolumn{5}{c}{Fine-Grained OOD}\\\cmidrule(lr){2-6}\cmidrule(lr){7-11}
     & \makebox[0.065\textwidth][c]{0.1} & \makebox[0.065\textwidth][c]{0.25} & \makebox[0.065\textwidth][c]{0.5$^*$} & \makebox[0.065\textwidth][c]{0.75} & \makebox[0.065\textwidth][c]{1} & \makebox[0.065\textwidth][c]{0.1} & \makebox[0.065\textwidth][c]{0.2$^*$} & \makebox[0.065\textwidth][c]{0.3} & \makebox[0.065\textwidth][c]{0.4} & \makebox[0.065\textwidth][c]{0.5} \\
     \midrule
    \makebox[0.005\textwidth][c]{AUC} & \makebox[0.005\textwidth][c]{97.1 / 98.4} & \makebox[0.005\textwidth][c]{97.7 / 98.8} & \makebox[0.005\textwidth][c]{\textbf{98.0 / 98.9}} & \makebox[0.005\textwidth][c]{\textbf{98.0} / 98.7} & \makebox[0.005\textwidth][c]{97.8 / 98.5} & \makebox[0.005\textwidth][c]{96.4 / 97.0} & \makebox[0.005\textwidth][c]{\textbf{96.5 / 97.1}} & \makebox[0.005\textwidth][c]{96.4 / 97.0} & \makebox[0.005\textwidth][c]{96.3 / 96.9} & \makebox[0.005\textwidth][c]{96.1 / 96.7}\\
    \makebox[0.005\textwidth][c]{FPR} & \makebox[0.005\textwidth][c]{12.9 / 7.3} & \makebox[0.005\textwidth][c]{10.0 / \textbf{5.8}} & \makebox[0.005\textwidth][c]{\textbf{9.9 / 5.8}} & \makebox[0.005\textwidth][c]{\textbf{9.9} / 6.5} & \makebox[0.005\textwidth][c]{10.7 / 7.5} & \makebox[0.005\textwidth][c]{16.8 / 12.7} & \makebox[0.005\textwidth][c]{\textbf{16.3 / 12.2}} & \makebox[0.005\textwidth][c]{16.4 / 12.8} & \makebox[0.005\textwidth][c]{18.4 / 14.5} & \makebox[0.005\textwidth][c]{19.0 / 16.1}\\
    \bottomrule
  \end{tabular}}
\end{table*}

\begin{table*}[t]
\caption{OOD detection on ImageNet and CUB benchmarks with various OOD detectors and CLIP ViT-L/14.}
\label{vitl14}
\centering
{\scriptsize
\begin{tabular}{ccccccccccccccccccc}
    \toprule
     \makebox[0.04\textwidth][c]{Metr.} & \makebox[0.05\textwidth][c]{$\mathcal{D}$} & \makebox[0.025\textwidth][c]{MLS} & \makebox[0.025\textwidth][c]{KL} & \makebox[0.025\textwidth][c]{EBO} & \makebox[0.025\textwidth][c]{ViM} & \makebox[0.025\textwidth][c]{ODIN} & \makebox[0.025\textwidth][c]{MCM} & \makebox[0.025\textwidth][c]{Gap} & \makebox[0.025\textwidth][c]{GL/R} & \makebox[0.025\textwidth][c]{EOE$^*$} & \makebox[0.025\textwidth][c]{NEG$^*$} & \makebox[0.025\textwidth][c]{Dual} & \makebox[0.025\textwidth][c]{CSP$^*$} & \makebox[0.025\textwidth][c]{COOK} & \makebox[0.025\textwidth][c]{$\Delta E$} & \makebox[0.03\textwidth][c]{CoOD-F} & \makebox[0.03\textwidth][c]{CoOD} & \makebox[0.03\textwidth][c]{CoOD$^*$}\\\cmidrule(lr){1-2}\cmidrule(lr){3-7}\cmidrule(lr){8-16}\cmidrule(lr){17-19}
    \makebox[0.005\textwidth][c]{\multirow{2}{*}{AUC}} & \makebox[0.005\textwidth][c]{ImageNet} & \makebox[0.005\textwidth][c]{83.5} & \makebox[0.005\textwidth][c]{64.8}& \makebox[0.005\textwidth][c]{80.0} & \makebox[0.005\textwidth][c]{68.6} & \makebox[0.005\textwidth][c]{91.1} & \makebox[0.005\textwidth][c]{90.4} & \makebox[0.005\textwidth][c]{90.6} & \makebox[0.005\textwidth][c]{91.1} & \makebox[0.005\textwidth][c]{93.3} & \makebox[0.005\textwidth][c]{94.3} & \makebox[0.005\textwidth][c]{90.7} & \makebox[0.005\textwidth][c]{95.7} & \makebox[0.005\textwidth][c]{91.5} & \makebox[0.005\textwidth][c]{90.4} & \makebox[0.005\textwidth][c]{96.6} & \makebox[0.005\textwidth][c]{97.2} & \makebox[0.005\textwidth][c]{\textbf{97.8}}\\
    & CUB& \makebox[0.005\textwidth][c]{80.5} & \makebox[0.005\textwidth][c]{64.6} & \makebox[0.005\textwidth][c]{80.1} & \makebox[0.005\textwidth][c]{69.8} & \makebox[0.005\textwidth][c]{77.5} & \makebox[0.005\textwidth][c]{77.6} & \makebox[0.005\textwidth][c]{70.8} & \makebox[0.005\textwidth][c]{76.7} & \makebox[0.005\textwidth][c]{79.1} & \makebox[0.005\textwidth][c]{80.1} & \makebox[0.005\textwidth][c]{78.8} & \makebox[0.005\textwidth][c]{81.1} & \makebox[0.005\textwidth][c]{77.6} & \makebox[0.005\textwidth][c]{76.2} & \makebox[0.005\textwidth][c]{94.9}& \makebox[0.005\textwidth][c]{96.4} & \makebox[0.005\textwidth][c]{\textbf{97.1}}\\
    \midrule
    \makebox[0.005\textwidth][c]{\multirow{2}{*}{FPR}} & \makebox[0.005\textwidth][c]{ImageNet} & \makebox[0.005\textwidth][c]{48.6} & \makebox[0.005\textwidth][c]{80.5} & \makebox[0.005\textwidth][c]{51.2} & \makebox[0.005\textwidth][c]{63.7} & \makebox[0.005\textwidth][c]{37.4} & \makebox[0.005\textwidth][c]{37.8} & \makebox[0.005\textwidth][c]{37.2} & \makebox[0.005\textwidth][c]{39.3} & \makebox[0.005\textwidth][c]{27.7} & \makebox[0.005\textwidth][c]{25.3} & \makebox[0.005\textwidth][c]{37.2} & \makebox[0.005\textwidth][c]{23.1} & \makebox[0.005\textwidth][c]{34.4} & \makebox[0.005\textwidth][c]{38.3} & \makebox[0.005\textwidth][c]{14.3} & \makebox[0.005\textwidth][c]{13.0} & \makebox[0.005\textwidth][c]{\textbf{8.9}}\\
    & CUB & \makebox[0.005\textwidth][c]{68.9} & \makebox[0.005\textwidth][c]{85.3} & \makebox[0.005\textwidth][c]{69.8} & \makebox[0.005\textwidth][c]{81.3} & \makebox[0.005\textwidth][c]{76.2} & \makebox[0.005\textwidth][c]{74.9} & \makebox[0.005\textwidth][c]{70.7} & \makebox[0.005\textwidth][c]{76.8} & \makebox[0.005\textwidth][c]{72.4} & \makebox[0.005\textwidth][c]{71.0} & \makebox[0.005\textwidth][c]{73.6} & \makebox[0.005\textwidth][c]{66.8} & \makebox[0.005\textwidth][c]{75.1} & \makebox[0.005\textwidth][c]{77.1} & \makebox[0.005\textwidth][c]{27.2} & \makebox[0.005\textwidth][c]{18.7} & \makebox[0.005\textwidth][c]{\textbf{14.8}}\\
    \bottomrule
  \end{tabular}}
\end{table*}

\begin{table}[t]
\caption{Ablation study of proposed modules on ImageNet54. - indicates unavailable modules when CSS or Suppression is ablated.}
\label{tab:ab}
\centering
{\scriptsize
\begin{tabular}{cccccccc}
    \toprule
    \makebox[0.04\textwidth][c]{CSS} & \makebox[0.03\textwidth][c]{\multirow{4}{*}{\rotatebox{90}{Baseline\ \ }}}&&&\checkmark&\checkmark&\checkmark&\checkmark\\
    \makebox[0.015\textwidth][c]{CCS} & &\checkmark&\checkmark&&&&\checkmark\\\cmidrule(lr){1-1}\cmidrule(lr){3-4}\cmidrule(lr){5-7}\cmidrule(lr){8-8}
    \makebox[0.015\textwidth][c]{Sup} & &-&-&&\checkmark&\checkmark&\checkmark\\
    \makebox[0.015\textwidth][c]{Zoom} & &&\checkmark&-&&\checkmark&\checkmark\\\cmidrule(lr){1-1}\cmidrule(lr){2-2}\cmidrule(lr){3-4}\cmidrule(lr){5-7}\cmidrule(lr){8-8}
    \makebox[0.015\textwidth][c]{AUC} & \makebox[0.02\textwidth][c]{93.5} &\makebox[0.02\textwidth][c]{95.2}&\makebox[0.02\textwidth][c]{96.7}&\makebox[0.02\textwidth][c]{94.8}&\makebox[0.02\textwidth][c]{95.9}&\makebox[0.02\textwidth][c]{96.6}&\makebox[0.02\textwidth][c]{97.8}\\
    \makebox[0.015\textwidth][c]{FPR}&\makebox[0.035\textwidth][c]{24.0}&\makebox[0.035\textwidth][c]{16.2}&\makebox[0.035\textwidth][c]{14.2}&\makebox[0.035\textwidth][c]{19.8}&\makebox[0.035\textwidth][c]{16.1}&\makebox[0.035\textwidth][c]{14.3}&\makebox[0.035\textwidth][c]{9.7}\\
    \bottomrule
  \end{tabular}}
\end{table}

\subsection{Main Results}
\textbf{Component-level evidence provides consistent improvements across detectors, models and datasets.} 
\Cref{tab:1kcub,vitl14,tab:54104,openood} report the results of CoOD and its variants on coarse-grained ImageNet and near-OOD benchmarks. 
Across a wide range of detectors, models, and evaluation settings, CoOD consistently outperforms both traditional OOD detectors and recent CLIP-based methods, suggesting that component-level evidence improves detection reliability and complements training-based approaches. 
The faster variant without explicit visual component extraction (CoOD-F) also achieves competitive performance and surpasses methods that do not use OOD labels, though below the full version CoOD.
Moreover, when combined with auxiliary OOD labels, CoOD$^*$ is further improved, indicating compatibility with existing OOD-exposure strategies.

\textbf{CoOD is particularly effective for fine-grained and compositional OOD detection.}
As shown in \cref{tab:1kcub,tab: counterfactual}, CoOD yields a substantial FPR reduction on fine-grained and compositional OOD benchmarks relative to strong baselines. 
In particular, fine-grained appearance variations (\textit{e.g.}, bird plumage or leg in CUB) are captured by CSS, whereas compositional changes (\textit{e.g.}, placing a handle inside a cup in \cref{fig:counterfactual}) are captured by CCS. Taken together, these results indicate that CoOD can address distribution shifts across different granularities and compositional structures.

\textbf{CoOD remains effective under challenging conditions.} 
On ImageNet-54/-104 and the more challenging ObjectNet variants, which introduce substantial covariate shifts (\textit{e.g.} cluttered backgrounds, occlusions, extreme viewpoints, and scale variations, \cref{tab:54104}), most methods experience noticeable performance degradation. In this setting, which stresses the impact of extreme geometric shift on CCS, CoOD consistently maintains competitive performance, demonstrating robustness in complex and realistic settings.

\subsection{Analysis}
\label{ana}

\textbf{Non-textual CoOD.}  
When reliable language models are unavailable, we implement CoOD with a \textbf{visual-vocabulary variant} of $\bm t_{yp}$ by deep clustering \citep{ace} over super-pixel. We then feed them into the same pipeline, reaching only slightly worse results compared to vanilla CoOD, \textit{e.g.} -0.5/-0.1 AUC +1.8/+0.9 FPR on ImageNet-54, and -2.4/-3.0 AUC +9.5/+15.1 FPR on CUB.
While language can aid semantic alignment, the principal gains of CoOD stem from its component-centric design rather than reliance on textual signals alone.

\textbf{Effectiveness of Proposed Modules.}
\Cref{tab:ab} summarizes ablations for our key modules. \textbf{CCS} supplies complementary compositional cues, but without the \textbf{Zoom} preprocessing, the keypoint sampling and matching is noisy due to low foreground resolution and background interference. For \textbf{CSS}, extracting components without \textbf{Sup}pression noticeably degrades performance, consistent with the increased inter-component nuisance correlation in \cref{taylorp}, the global context leakage, and background noise. 
After suppression, \textbf{Zoom} further improves the component quality. Combining \textbf{CSS} with \textbf{CCS} yields the best results, confirming their complementary benefits.

\textbf{Weight $\alpha$.}
Ablation experiments in \cref{tab:alpha} indicate that CoOD is not sensitive to the weight $\alpha$ in \eqref{eq:alpha}, with different $\alpha$ yielding comparable results. Take a closer look, a relatively small $\alpha\!=\!0.2$ is slightly better for fine-grained detection, as these tasks typically involve minimal geometric changes within the same superclass. In contrast, a larger $\alpha\!=\!0.5$ benefits coarse-grained detection, since it identifies global OODs through compositional inconsistencies.

\begin{table}[t]
\caption{Ablation study of component number on CUB.}
\label{tab:cub123456}
\centering
{\scriptsize
\begin{tabular}{cccccccc}
    \toprule
    \makebox[0.055\textwidth][c]{$|\mathcal{P}_y|$} & \makebox[0.032\textwidth][c]{0} & \makebox[0.032\textwidth][c]{1} & \makebox[0.032\textwidth][c]{2} & \makebox[0.032\textwidth][c]{3} & \makebox[0.032\textwidth][c]{4$^*$} & \makebox[0.032\textwidth][c]{5} & \makebox[0.032\textwidth][c]{12}\\\cmidrule(lr){1-1}\cmidrule(lr){2-8}
    \makebox[0.005\textwidth][c]{AUC} & \makebox[0.005\textwidth][c]{75.4} & \makebox[0.005\textwidth][c]{95.8} & \makebox[0.005\textwidth][c]{96.4} & \makebox[0.005\textwidth][c]{96.5} & \makebox[0.005\textwidth][c]{\textbf{96.6}} & \makebox[0.005\textwidth][c]{96.2} & \makebox[0.005\textwidth][c]{95.9}\\
    \makebox[0.005\textwidth][c]{FPR} & \makebox[0.005\textwidth][c]{73.0} & \makebox[0.005\textwidth][c]{18.7} & \makebox[0.005\textwidth][c]{17.2} & \makebox[0.005\textwidth][c]{16.5} & \makebox[0.005\textwidth][c]{\textbf{15.8}} & \makebox[0.005\textwidth][c]{19.0} & \makebox[0.005\textwidth][c]{21.3}\\
    \bottomrule
  \end{tabular}}
\end{table}

\textbf{Vocabulary Quality.}
Appendix~\ref{llm} shows that the error rate of the vocabulary is negligible.
Because our prompts allow conservative fallbacks to higher-level concepts or analogy-based decomposition.
We further evaluate the vocabulary robustness via adversarial prompt:\textit{``Fewer than 5 components is strictly prohibited; each class should include at least 1 \textbf{rare or spurious component}''}.
Despite injecting errors and spurious correlations, CoOD achieves nearly the same performance (IN-54, -0.1 AUC/+0.6 FPR), indicating that CoOD is robust to vocabulary quality.

\textbf{Component Numbers.}
\Cref{tab:cub123456} shows the effect of component cardinality using the standard CUB part annotations. For a small number of components, LLM-guided decomposition reliably extracts the most salient parts (\textit{e.g.}, ``bird head'' at $0\!\to\!1$), and yields a large performance gain. This is more efficient than methods relying on thousands of OOD labels.
As the component number grows and fully covers the visual foreground, CoOD performance saturates due to the limited representation capacity of visual encoders for extremely fine-grained or low-resolution components. To close the gap, we appropriately summarize the closely related leaf parts into higher-level taxonomy nodes (\textit{e.g.} ``left / right leg''$\to$``limb'') as in \cref{vlmtext}, in order to improve the stability of component representation at low-resolution.

\textbf{Non-Rigid Cases}
Although non‑rigid classes cannot be directly modeled with rigid transformations, CoOD covers their diversy via coreset sampling, and most residual variations remain anatomically constrained (\textit{e.g.} CUB birds). For amorphous classes, LLMs usually produce a small number of components (reducing CoOD to a global detector) or uses semantically related components rather than forcing invalid rigid decompositions. Consequently, the impact on CCS is limited, and CoOD still outperforms baselines on ImageNet1K despite more amorphous classes.

\begin{table}[t]
\caption{Ablation study of coreset size. We report the increase in AUC and decrease in FPR when 10$\times$ larger coreset is used.}
\label{tab:dsize}
\centering
{\scriptsize
\begin{tabular}{ccc}
    \toprule
    \makebox[0.1\textwidth][c]{Metric} & \makebox[0.15\textwidth][c]{CoOD} & \makebox[0.15\textwidth][c]{CoOD$^*$}\\\cmidrule(lr){1-1}\cmidrule(lr){2-2}\cmidrule(lr){3-3}
    AUC & +0.2 / +0.2 & +0.2 / +0.3 \\
    FPR & -0.6 / -0.4 & -0.3 / -0.6\\
    \bottomrule
  \end{tabular}}
\end{table}

\textbf{Other Backbone.}
We conduct additional experiments on CLIP ViT-L/14, and the results demonstrate that CoOD is fully compatible with larger backbones. While all methods benefit from the stronger encoder, CoOD consistently maintains superior performance.

\textbf{Coreset size.}
We study the effect of enlarging the coreset size $|\mathcal{D}^*|$ by $10\times$ on ImageNet-54. 
A larger coreset provides more diverse ID composition for computing CCS, which improves the coverage of rare ID samples, leading to higher detection performance as shown in \cref{tab:dsize}. On the other hand, the marginal gains verify the effectiveness and efficiency of the core set strategy

\textbf{Efficiency.}
We parallelize the coreset matching in \cref{ccsmatching} and conduct all experiments on an EPYC9654–RTX4090 server. For a single image, MCM, CoOD-F, and CoOD($^*$) achieve 89, 85, and 1.5 FPS, respectively.
The lower efficiency of full CoOD mainly stems from its \textbf{fairness-oriented design}. In particular, we adopt GradCAM and traditional keypoint matching, rather than a segmentation or registration network.
Nevertheless, this additional computational overhead is justified by the gains in challenging fine-grained and compositional OOD detection.

\begin{figure}[t]
    \centerline{\includegraphics[width=1.\linewidth]{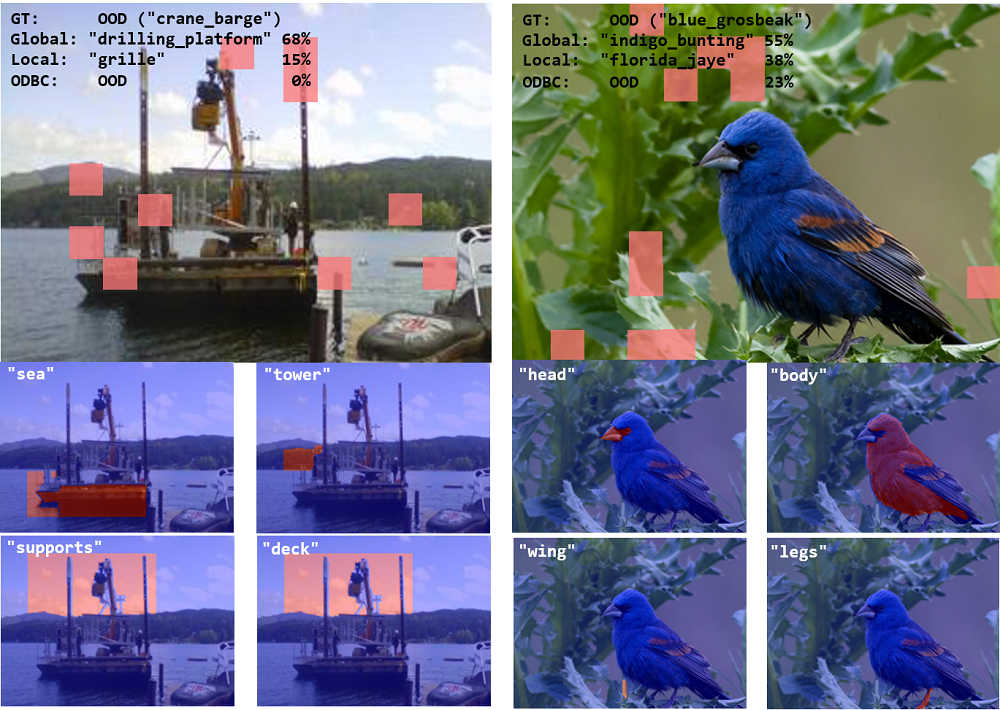}}
    \caption{
    Illustration of challenging OODs. The percentages represent ranking percentiles, facilitating comparison across different detectors, with lower values indicating more probable OODs. The highlight regions show local responses \citep{topkgl}.
    }\vskip -0.2in
    \label{fig:vis}
\end{figure}

\textbf{Visualization.} 
\Cref{fig:vis} shows two challenging OOD samples that traditional methods cannot detect. For example, global detector mis-classifies the two samples (whose ground truth classes are ``crane barge'' and ``blue grosbeak'' respectively) to ID classes due to high visual similarity. Local method cannot output correct results either, as it produces noisy responses and misses long-range relations. 
For CoOD, the first sample lacks the components of ``drilling-platform'' thus receives a high component shift score, and the spatial misalignments with ID compositions are also flagged by CCS. 
In the second case, CoOD localizes the components corresponding to ``indigo bunting'' (the same as ``blue grosbeak'') and suppresses the orange-feather signal on the ``wing'' which is absent from the ID component, thereby correctly detects this fine-grained OOD.

\section{Conclusion}
We present OOD Detection-by-Components (CoOD), a training-free framework that (1) identifies and aggregates patches into component representations to suppress noise and expose fine-grained OODs (via Component Shift Score, CSS), and (2) evaluates compositional OOD among components to reveal invalid compositions (via Compositional Consistency Score, CCS). CoOD provides interpretable component-level evidence and consistently improves both coarse and fine-grained OOD detection.
Future work will extend spatial components to more general feature-level components, such as color and viewpoint. When combined with large-scale pretrained models, this approach may better characterize and detect arbitrary distribution shifts arising in practical deployments, providing a realistic pathway toward more adaptable OOD detection in real-world settings.

\section*{Impact Statement}
Fine-grained distributional shifts are common in real-world deployments, and practitioners frequently require interpretable, component-level evidence to understand or validate OOD decisions.
This paper introduces a component-centric, training-free paradigm for OOD detection that reconciles the trade-off between sensitivity to fine-grained OODs and robustness to natural ID variability.
Our method detects subtle appearance and compositional OODs without retraining, is broadly applicable to practical vision systems, and improves the safety and reliability of downstream decision-making.
More broadly, this work highlights the potential of representation learning grounded solely in training distributions and strategies to model existing data more faithfully, fostering progress in building trustworthy AI.

\bibliography{example_paper}
\bibliographystyle{icml2026}

\newpage
\appendix
\onecolumn

\section{The Counterfactual Dataset and ImageNet Compositional OOD Splits}
\label{compositional ood dataset} 
The compositional OOD splits of ImageNet are manually constructed by grouping classes that share similar visual components and differ mainly in composition. Specifically, we identify the following subgroups as more likely to exhibit shared components and compositional variations: \texttt{['barber chair', 'folding chair', 'rocking chair']}, \texttt{['dining table', 'desk']}, \texttt{['bicycle-built-for-two', 'tricycle', 'unicycle', 'mountain bike']}, \texttt{['warplane', 'airliner']}, \texttt{['rifle', 'assault rifle']}, \texttt{['ladle', 'wooden spoon']}, \texttt{['mixing bowl', 'soup bowl']}, \texttt{['teapot', 'pitcher']}, \texttt{['water bottle', 'beer glass', 'measuring cup', 'red wine']}, \texttt{['cup', 'coffee mug']}. Within each subgroup, we systematically rotate the ID assignment such that each class is designated as ID in turn, with all other subclasses regarded as OOD during that iteration.

\begin{figure}[h]
  \begin{center}
    \centerline{\includegraphics[width=0.6\columnwidth]{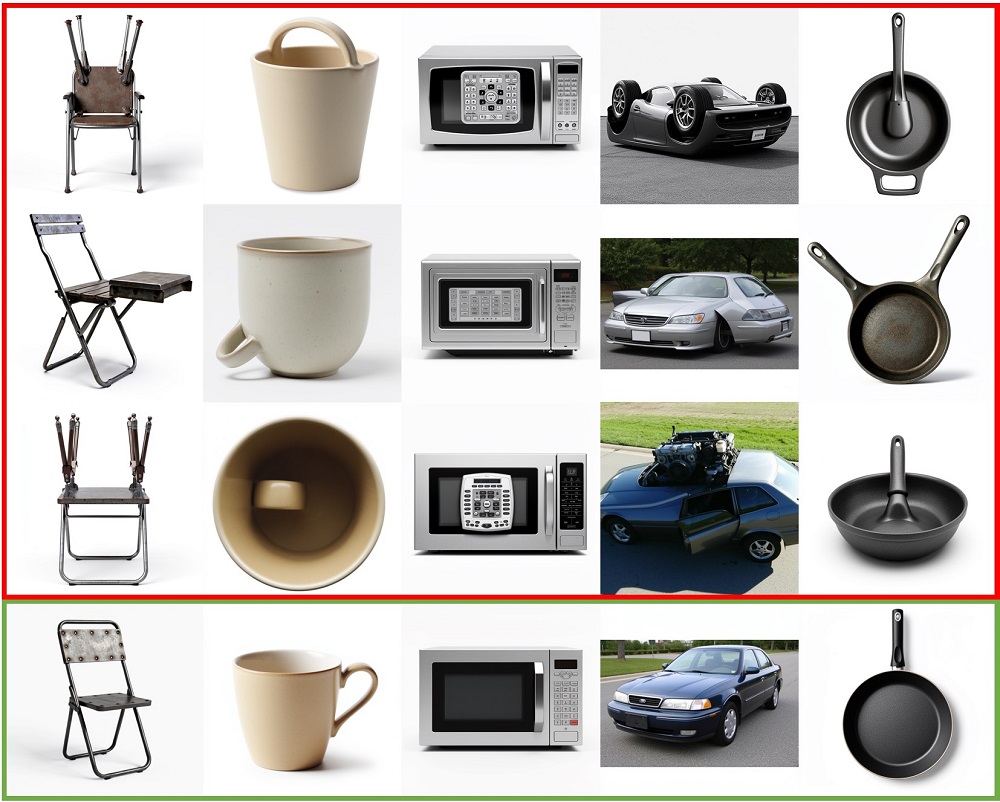}}
    \caption{Illustration of generated counterfactual dataset for testing compositional OODs.
    }\vskip -0.3in
    \label{fig:counterfactual}
  \end{center}
\end{figure}

While this ImageNet split provides a benchmark for compositional OOD evaluation, it does not fully control for component consistency and often entangles compositional variation with other nuisance factors. To isolate compositional changes, we further generate one-component-changed counterfactual samples using a diffusion model (FLUX.2-9B) conditioned on prompts of the form \textit{``A surreal industrial design prototype of a $\langle class\rangle$, where $\langle component\rangle$ mounted $\langle impossible\ position\rangle$.''}. This procedure enables the controlled generation of diverse OOD samples that mainly differ in component composition. We then manually inspect the generated images to remove samples with substantial component appearance difference or loss of global semantic consistency, while using the corresponding prompt \textit{``A surreal industrial design prototype of a $\langle class\rangle$, where $\langle component\rangle$ mounted $\langle normal\ position\rangle$.''} and same seed to generate ID data. \cref{fig:counterfactual} provides representative examples of the generated counterfactual dataset.

\section{More Experiment Details}
\label{moreexp}
\textbf{Main Experiment.} We provide more experimental results on ImageNet-54, ObjectNet-54, ImageNet-104, and ObjectNet-104 in Table~\ref{tab:54104} for \textbf{setup-i} (standard ImageNet ID \textit{vs.} 4 CLIP OOD dataset) and \textbf{setup-iii} (challenging covariate-ID from ObjectNet \textit{vs.} 4 CLIP OOD dataset). The results again support the conclusions in the main paper.

\begin{table*}[h]
\caption{OOD detection on ImageNet-54, ObjectNet-54, ImageNet-104, and ObjectNet-104 with various OOD detectors and models.}
\label{tab:54104}
\centering
{\scriptsize
\begin{tabular}{cccccccccccccccccccc}
    \toprule
    \makebox[0.005\textwidth][c]{$\mathcal{D}$} & \makebox[0.005\textwidth][c]{Metr.} & \makebox[0.04\textwidth][c]{Tr.} & \makebox[0.025\textwidth][c]{MLS} & \makebox[0.025\textwidth][c]{KL} & \makebox[0.025\textwidth][c]{EBO} & \makebox[0.025\textwidth][c]{ViM} & \makebox[0.025\textwidth][c]{ODIN} & \makebox[0.025\textwidth][c]{MCM} & \makebox[0.025\textwidth][c]{Gap} & \makebox[0.025\textwidth][c]{GL/R} & \makebox[0.025\textwidth][c]{EOE$^*$} & \makebox[0.025\textwidth][c]{NEG$^*$} & \makebox[0.025\textwidth][c]{Dual} & \makebox[0.025\textwidth][c]{CSP$^*$} & \makebox[0.025\textwidth][c]{COOK} & \makebox[0.025\textwidth][c]{$\Delta E$} & \makebox[0.025\textwidth][c]{CoOD-F} & \makebox[0.025\textwidth][c]{CoOD} & \makebox[0.025\textwidth][c]{CoOD$^*$}\\\cmidrule(lr){1-3}\cmidrule(lr){4-8}\cmidrule(lr){9-17}\cmidrule(lr){18-20}
    \makebox[0.005\textwidth][c]{\multirow{10}{*}{\rotatebox{90}{ImageNet-54}}} & \makebox[0.005\textwidth][c]{\multirow{5}{*}{\rotatebox{90}{AUC}}} & CLIP & \makebox[0.005\textwidth][c]{81.5} & \makebox[0.005\textwidth][c]{93.1} & \makebox[0.005\textwidth][c]{81.5} & \makebox[0.005\textwidth][c]{84.5} & \makebox[0.005\textwidth][c]{92.8} & \makebox[0.005\textwidth][c]{93.4} & \makebox[0.005\textwidth][c]{90.8} & \makebox[0.005\textwidth][c]{93.6} & \makebox[0.005\textwidth][c]{90.3} & \makebox[0.005\textwidth][c]{96.9} & \makebox[0.005\textwidth][c]{93.7}& \makebox[0.005\textwidth][c]{97.7} & \makebox[0.005\textwidth][c]{93.4} & \makebox[0.005\textwidth][c]{93.9} & \makebox[0.005\textwidth][c]{96.9} & \makebox[0.005\textwidth][c]{98.4}& \makebox[0.005\textwidth][c]{\textbf{98.6}}\\
    & & CoOp & \makebox[0.005\textwidth][c]{90.2} & \makebox[0.005\textwidth][c]{91.6} & \makebox[0.005\textwidth][c]{90.2} & \makebox[0.005\textwidth][c]{87.2} & \makebox[0.005\textwidth][c]{93.3} & \makebox[0.005\textwidth][c]{93.9} & \makebox[0.005\textwidth][c]{92.1} & \makebox[0.005\textwidth][c]{94.0} & \makebox[0.005\textwidth][c]{93.2} & \makebox[0.005\textwidth][c]{95.3} & \makebox[0.005\textwidth][c]{94.1} & \makebox[0.005\textwidth][c]{96.5} & \makebox[0.005\textwidth][c]{93.9} & \makebox[0.005\textwidth][c]{94.1} & \makebox[0.005\textwidth][c]{95.1} & \makebox[0.005\textwidth][c]{98.7} & \makebox[0.005\textwidth][c]{\textbf{99.2}}\\
    & & \makebox[0.05\textwidth][c]{LoCoOp} & \makebox[0.005\textwidth][c]{83.7} & \makebox[0.005\textwidth][c]{91.2 } & \makebox[0.005\textwidth][c]{83.7} & \makebox[0.005\textwidth][c]{85.1} & \makebox[0.005\textwidth][c]{93.3} & \makebox[0.005\textwidth][c]{95.5} & \makebox[0.005\textwidth][c]{94.2} & \makebox[0.005\textwidth][c]{95.6} & \makebox[0.005\textwidth][c]{92.6} & \makebox[0.005\textwidth][c]{91.4} & \makebox[0.005\textwidth][c]{95.7} & \makebox[0.005\textwidth][c]{93.5} & \makebox[0.005\textwidth][c]{95.5} & \makebox[0.005\textwidth][c]{95.6} & \makebox[0.005\textwidth][c]{96.0} & \makebox[0.005\textwidth][c]{98.6} & \makebox[0.005\textwidth][c]{\textbf{99.1}}\\
    & & IDLike &\makebox[0.005\textwidth][c]{87.4} & \makebox[0.005\textwidth][c]{93.1} & \makebox[0.005\textwidth][c]{87.4} & \makebox[0.005\textwidth][c]{86.3} & \makebox[0.005\textwidth][c]{93.2} & \makebox[0.005\textwidth][c]{92.7} & \makebox[0.005\textwidth][c]{93.3} & \makebox[0.005\textwidth][c]{93.0} & \makebox[0.005\textwidth][c]{91.9} & \makebox[0.005\textwidth][c]{93.6} & \makebox[0.005\textwidth][c]{95.2} & \makebox[0.005\textwidth][c]{95.3} & \makebox[0.005\textwidth][c]{92.8} & \makebox[0.005\textwidth][c]{95.1} & \makebox[0.005\textwidth][c]{94.5} & \makebox[0.005\textwidth][c]{98.6} & \makebox[0.005\textwidth][c]{\textbf{98.7}}\\
    & & LoPro &\makebox[0.005\textwidth][c]{81.2}&\makebox[0.005\textwidth][c]{91.7}&\makebox[0.005\textwidth][c]{81.2}&\makebox[0.005\textwidth][c]{84.4}&\makebox[0.005\textwidth][c]{92.9}&\makebox[0.005\textwidth][c]{93.2}&\makebox[0.005\textwidth][c]{90.7}&\makebox[0.005\textwidth][c]{93.4}&\makebox[0.005\textwidth][c]{89.8}&\makebox[0.005\textwidth][c]{93.7}&\makebox[0.005\textwidth][c]{93.5} & \makebox[0.005\textwidth][c]{96.8} & \makebox[0.005\textwidth][c]{93.2} & \makebox[0.005\textwidth][c]{93.6} & \makebox[0.005\textwidth][c]{96.2} & \makebox[0.005\textwidth][c]{97.8}&\makebox[0.005\textwidth][c]{\textbf{98.6}}\\\cmidrule(lr){2-3}\cmidrule(lr){4-8}\cmidrule(lr){9-17}\cmidrule(lr){18-20}
    & \makebox[0.005\textwidth][c]{\multirow{5}{*}{\rotatebox{90}{FPR$_{95}$}}} & CLIP & \makebox[0.005\textwidth][c]{46.1} & \makebox[0.005\textwidth][c]{25.1} & \makebox[0.005\textwidth][c]{46.1} & \makebox[0.005\textwidth][c]{31.3} & \makebox[0.005\textwidth][c]{23.6} & \makebox[0.005\textwidth][c]{23.9} & \makebox[0.005\textwidth][c]{29.6} & \makebox[0.005\textwidth][c]{23.6} & \makebox[0.005\textwidth][c]{30.2} & \makebox[0.005\textwidth][c]{14.2} & \makebox[0.005\textwidth][c]{23.0} & \makebox[0.005\textwidth][c]{10.1} & \makebox[0.005\textwidth][c]{23.8} & \makebox[0.005\textwidth][c]{22.9} & \makebox[0.005\textwidth][c]{13.9} & \makebox[0.005\textwidth][c]{9.7} & \makebox[0.005\textwidth][c]{\textbf{6.5}}\\
    & & CoOp & \makebox[0.005\textwidth][c]{31.1} & \makebox[0.005\textwidth][c]{30.9} & \makebox[0.005\textwidth][c]{31.1} & \makebox[0.005\textwidth][c]{27.1} & \makebox[0.005\textwidth][c]{24.6} & \makebox[0.005\textwidth][c]{23.8} & \makebox[0.005\textwidth][c]{27.3} & \makebox[0.005\textwidth][c]{23.4} & \makebox[0.005\textwidth][c]{24.3} & \makebox[0.005\textwidth][c]{18.4} & \makebox[0.005\textwidth][c]{22.8} & \makebox[0.005\textwidth][c]{14.8} & \makebox[0.005\textwidth][c]{23.7} & \makebox[0.005\textwidth][c]{23.3} & \makebox[0.005\textwidth][c]{22.2} & \makebox[0.005\textwidth][c]{6.4} & \makebox[0.005\textwidth][c]{\textbf{3.4}}\\
    & & \makebox[0.05\textwidth][c]{LoCoOp} & \makebox[0.005\textwidth][c]{43.9} & \makebox[0.005\textwidth][c]{31.9 } & \makebox[0.005\textwidth][c]{43.9} & \makebox[0.005\textwidth][c]{29.2} & \makebox[0.005\textwidth][c]{24.3} & \makebox[0.005\textwidth][c]{18.4} & \makebox[0.005\textwidth][c]{21.9} & \makebox[0.005\textwidth][c]{17.8} & \makebox[0.005\textwidth][c]{23.6} & \makebox[0.005\textwidth][c]{28.5} & \makebox[0.005\textwidth][c]{17.4} & \makebox[0.005\textwidth][c]{22.8} & \makebox[0.005\textwidth][c]{18.3} & \makebox[0.005\textwidth][c]{18.1} & \makebox[0.005\textwidth][c]{17.2} & \makebox[0.005\textwidth][c]{6.0} & \makebox[0.005\textwidth][c]{\textbf{4.0}}\\
    & & IDLike &\makebox[0.005\textwidth][c]{39.7} & \makebox[0.005\textwidth][c]{23.9} & \makebox[0.005\textwidth][c]{39.7} & \makebox[0.005\textwidth][c]{28.8} & \makebox[0.005\textwidth][c]{23.6} & \makebox[0.005\textwidth][c]{27.3} & \makebox[0.005\textwidth][c]{23.1} & \makebox[0.005\textwidth][c]{26.9} & \makebox[0.005\textwidth][c]{29.2} & \makebox[0.005\textwidth][c]{25.0} & \makebox[0.005\textwidth][c]{18.1} & \makebox[0.005\textwidth][c]{19.5} & \makebox[0.005\textwidth][c]{27.2} & \makebox[0.005\textwidth][c]{18.6} & \makebox[0.005\textwidth][c]{22.6} & \makebox[0.005\textwidth][c]{6.8} & \makebox[0.005\textwidth][c]{\textbf{5.6}}\\
    & & LoPro &\makebox[0.005\textwidth][c]{46.1}&\makebox[0.005\textwidth][c]{28.1}&\makebox[0.005\textwidth][c]{46.2}&\makebox[0.005\textwidth][c]{31.7}&\makebox[0.005\textwidth][c]{23.7}&\makebox[0.005\textwidth][c]{24.3}&\makebox[0.005\textwidth][c]{29.3}&\makebox[0.005\textwidth][c]{23.9}&\makebox[0.005\textwidth][c]{30.6}&\makebox[0.005\textwidth][c]{21.9}&\makebox[0.005\textwidth][c]{23.6} & \makebox[0.005\textwidth][c]{14.0} & \makebox[0.005\textwidth][c]{24.2} & \makebox[0.005\textwidth][c]{23.1} & \makebox[0.005\textwidth][c]{15.7}&\makebox[0.005\textwidth][c]{9.7} &\makebox[0.005\textwidth][c]{\textbf{6.8}}\\
    \midrule
    \makebox[0.005\textwidth][c]{\multirow{10}{*}{\rotatebox{90}{ObjectNet-54}}} & \makebox[0.005\textwidth][c]{\multirow{5}{*}{\rotatebox{90}{AUC}}} & CLIP & \makebox[0.005\textwidth][c]{86.8} & \makebox[0.005\textwidth][c]{90.9} & \makebox[0.005\textwidth][c]{86.8} & \makebox[0.005\textwidth][c]{84.9} & \makebox[0.005\textwidth][c]{91.7} & \makebox[0.005\textwidth][c]{88.4} & \makebox[0.005\textwidth][c]{83.5} & \makebox[0.005\textwidth][c]{88.7} & \makebox[0.005\textwidth][c]{90.6} & \makebox[0.005\textwidth][c]{96.8} & \makebox[0.005\textwidth][c]{88.6} & \makebox[0.005\textwidth][c]{98.0} & \makebox[0.005\textwidth][c]{88.4} & \makebox[0.005\textwidth][c]{89.5} & \makebox[0.005\textwidth][c]{91.0} & \makebox[0.005\textwidth][c]{94.3} & \makebox[0.005\textwidth][c]{\textbf{98.5}}\\
    & & CoOp & \makebox[0.005\textwidth][c]{87.8} & \makebox[0.005\textwidth][c]{92.2} & \makebox[0.005\textwidth][c]{87.8} & \makebox[0.005\textwidth][c]{85.5} & \makebox[0.005\textwidth][c]{89.8} & \makebox[0.005\textwidth][c]{87.8} & \makebox[0.005\textwidth][c]{82.6} & \makebox[0.005\textwidth][c]{88.1} & \makebox[0.005\textwidth][c]{90.4} & \makebox[0.005\textwidth][c]{92.8} & \makebox[0.005\textwidth][c]{88.2} &\makebox[0.005\textwidth][c]{94.8} & \makebox[0.005\textwidth][c]{87.8} & \makebox[0.005\textwidth][c]{89.1} & \makebox[0.005\textwidth][c]{91.9} & \makebox[0.005\textwidth][c]{95.3} & \makebox[0.005\textwidth][c]{\textbf{98.5}}\\
    & & \makebox[0.05\textwidth][c]{LoCoOp} & \makebox[0.005\textwidth][c]{88.0} & \makebox[0.005\textwidth][c]{88.3} & \makebox[0.005\textwidth][c]{87.9} & \makebox[0.005\textwidth][c]{85.7} & \makebox[0.005\textwidth][c]{90.4} & \makebox[0.005\textwidth][c]{90.7} & \makebox[0.005\textwidth][c]{87.5} & \makebox[0.005\textwidth][c]{91.0} & \makebox[0.005\textwidth][c]{92.5} & \makebox[0.005\textwidth][c]{93.5} & \makebox[0.005\textwidth][c]{90.9} & \makebox[0.005\textwidth][c]{95.4} & \makebox[0.005\textwidth][c]{90.7} & \makebox[0.005\textwidth][c]{91.7} & \makebox[0.005\textwidth][c]{92.4} & \makebox[0.005\textwidth][c]{95.4} & \makebox[0.005\textwidth][c]{\textbf{98.3}}\\
    & & IDLike &\makebox[0.005\textwidth][c]{94.4} & \makebox[0.005\textwidth][c]{93.5} & \makebox[0.005\textwidth][c]{94.4} & \makebox[0.005\textwidth][c]{90.8} & \makebox[0.005\textwidth][c]{90.3} & \makebox[0.005\textwidth][c]{96.0} & \makebox[0.005\textwidth][c]{87.5} & \makebox[0.005\textwidth][c]{96.2} & \makebox[0.005\textwidth][c]{95.8} & \makebox[0.005\textwidth][c]{97.2} & \makebox[0.005\textwidth][c]{92.2} & \makebox[0.005\textwidth][c]{98.0} & \makebox[0.005\textwidth][c]{96.1} & \makebox[0.005\textwidth][c]{93.0} & \makebox[0.005\textwidth][c]{93.9} & \makebox[0.005\textwidth][c]{97.8} & \makebox[0.005\textwidth][c]{\textbf{98.6}} \\
    & & LoPro &\makebox[0.005\textwidth][c]{86.5}&\makebox[0.005\textwidth][c]{90.8}&\makebox[0.005\textwidth][c]{86.5}&\makebox[0.005\textwidth][c]{84.8}&\makebox[0.005\textwidth][c]{91.8}&\makebox[0.005\textwidth][c]{88.7}&\makebox[0.005\textwidth][c]{83.5}&\makebox[0.005\textwidth][c]{88.9}&\makebox[0.005\textwidth][c]{90.5}&\makebox[0.005\textwidth][c]{95.8}&\makebox[0.005\textwidth][c]{88.9}&\makebox[0.005\textwidth][c]{98.0}&\makebox[0.005\textwidth][c]{88.7}&\makebox[0.005\textwidth][c]{89.9}& \makebox[0.005\textwidth][c]{91.0}&\makebox[0.005\textwidth][c]{94.5} &\makebox[0.005\textwidth][c]{\textbf{98.5}}\\\cmidrule(lr){2-3}\cmidrule(lr){4-8}\cmidrule(lr){9-17}\cmidrule(lr){18-20}
    & \makebox[0.005\textwidth][c]{\multirow{5}{*}{\rotatebox{90}{FPR$_{95}$}}} & CLIP & \makebox[0.005\textwidth][c]{39.1} & \makebox[0.005\textwidth][c]{31.7} & \makebox[0.005\textwidth][c]{39.1} & \makebox[0.005\textwidth][c]{40.3} & \makebox[0.005\textwidth][c]{27.5} & \makebox[0.005\textwidth][c]{38.4} & \makebox[0.005\textwidth][c]{45.5} & \makebox[0.005\textwidth][c]{38.1} & \makebox[0.005\textwidth][c]{34.2} & \makebox[0.005\textwidth][c]{16.8} & \makebox[0.005\textwidth][c]{37.5} & \makebox[0.005\textwidth][c]{9.4} & \makebox[0.005\textwidth][c]{38.4} & \makebox[0.005\textwidth][c]{35.8} & \makebox[0.005\textwidth][c]{22.7}& \makebox[0.005\textwidth][c]{23.2} & \makebox[0.005\textwidth][c]{\textbf{8.0}}\\
    & & CoOp & \makebox[0.005\textwidth][c]{41.0} & \makebox[0.005\textwidth][c]{31.8} & \makebox[0.005\textwidth][c]{41.0} & \makebox[0.005\textwidth][c]{40.2} & \makebox[0.005\textwidth][c]{33.4} & \makebox[0.005\textwidth][c]{41.3} & \makebox[0.005\textwidth][c]{47.5} & \makebox[0.005\textwidth][c]{40.9} & \makebox[0.005\textwidth][c]{36.3} & \makebox[0.005\textwidth][c]{29.9} & \makebox[0.005\textwidth][c]{40.2} & \makebox[0.005\textwidth][c]{24.2} & \makebox[0.005\textwidth][c]{41.2} & \makebox[0.005\textwidth][c]{37.8} & \makebox[0.005\textwidth][c]{21.4} & \makebox[0.005\textwidth][c]{22.0} & \makebox[0.005\textwidth][c]{\textbf{7.9}}\\
    & & \makebox[0.05\textwidth][c]{LoCoOp} & \makebox[0.005\textwidth][c]{38.8} & \makebox[0.005\textwidth][c]{39.8} & \makebox[0.005\textwidth][c]{38.8} & \makebox[0.005\textwidth][c]{37.4} & \makebox[0.005\textwidth][c]{31.8} & \makebox[0.005\textwidth][c]{33.3} & \makebox[0.005\textwidth][c]{39.0} & \makebox[0.005\textwidth][c]{32.6} & \makebox[0.005\textwidth][c]{27.3} & \makebox[0.005\textwidth][c]{25.0} & \makebox[0.005\textwidth][c]{32.4} & \makebox[0.005\textwidth][c]{18.4} & \makebox[0.005\textwidth][c]{33.3} & \makebox[0.005\textwidth][c]{30.0} & \makebox[0.005\textwidth][c]{26.1} & \makebox[0.005\textwidth][c]{20.6} & \makebox[0.005\textwidth][c]{\textbf{8.0}}\\
    & & IDLike &\makebox[0.005\textwidth][c]{19.1} & \makebox[0.005\textwidth][c]{24.6} & \makebox[0.005\textwidth][c]{19.0} & \makebox[0.005\textwidth][c]{25.6} & \makebox[0.005\textwidth][c]{30.8} & \makebox[0.005\textwidth][c]{16.8} & \makebox[0.005\textwidth][c]{38.8} & \makebox[0.005\textwidth][c]{16.6} & \makebox[0.005\textwidth][c]{17.1} & \makebox[0.005\textwidth][c]{10.8} & \makebox[0.005\textwidth][c]{30.0} & \makebox[0.005\textwidth][c]{8.4} & \makebox[0.005\textwidth][c]{16.8} & \makebox[0.005\textwidth][c]{27.9} & \makebox[0.005\textwidth][c]{20.2} & \makebox[0.005\textwidth][c]{10.0} & \makebox[0.005\textwidth][c]{\textbf{7.0}} \\
    & & LoPro &\makebox[0.005\textwidth][c]{39.1}&\makebox[0.005\textwidth][c]{31.7}&\makebox[0.005\textwidth][c]{39.1}&\makebox[0.005\textwidth][c]{40.3}&\makebox[0.005\textwidth][c]{27.5}&\makebox[0.005\textwidth][c]{38.3}&\makebox[0.005\textwidth][c]{45.2}&\makebox[0.005\textwidth][c]{37.8}&\makebox[0.005\textwidth][c]{33.7}&\makebox[0.005\textwidth][c]{50.9}&\makebox[0.005\textwidth][c]{37.5}&\makebox[0.005\textwidth][c]{8.9}&\makebox[0.005\textwidth][c]{38.2}&\makebox[0.005\textwidth][c]{35.0} & \makebox[0.005\textwidth][c]{31.0} &\makebox[0.005\textwidth][c]{22.9} &\makebox[0.005\textwidth][c]{\textbf{8.0}}\\
    \midrule
    \makebox[0.005\textwidth][c]{\multirow{10}{*}{\rotatebox{90}{ImageNet-104}}} & \makebox[0.005\textwidth][c]{\multirow{5}{*}{\rotatebox{90}{AUC}}} & CLIP & \makebox[0.005\textwidth][c]{81.3} & \makebox[0.005\textwidth][c]{93.2} & \makebox[0.005\textwidth][c]{81.3} & \makebox[0.005\textwidth][c]{89.6} & \makebox[0.005\textwidth][c]{93.6} & \makebox[0.005\textwidth][c]{93.6} & \makebox[0.005\textwidth][c]{90.6} & \makebox[0.005\textwidth][c]{93.8} & \makebox[0.005\textwidth][c]{92.0} & \makebox[0.005\textwidth][c]{96.5} & \makebox[0.005\textwidth][c]{93.6} & \makebox[0.005\textwidth][c]{97.4} & \makebox[0.005\textwidth][c]{93.6} & \makebox[0.005\textwidth][c]{94.1} & \makebox[0.005\textwidth][c]{96.0} & \makebox[0.005\textwidth][c]{97.3}& \makebox[0.005\textwidth][c]{\textbf{98.3}}\\
    & & CoOp & \makebox[0.005\textwidth][c]{90.2} & \makebox[0.005\textwidth][c]{92.6} & \makebox[0.005\textwidth][c]{90.2} & \makebox[0.005\textwidth][c]{92.2} & \makebox[0.005\textwidth][c]{93.9} & \makebox[0.005\textwidth][c]{93.7} & \makebox[0.005\textwidth][c]{91.2} & \makebox[0.005\textwidth][c]{93.8} & \makebox[0.005\textwidth][c]{93.8} & \makebox[0.005\textwidth][c]{94.5} & \makebox[0.005\textwidth][c]{93.8} &\makebox[0.005\textwidth][c]{96.3} & \makebox[0.005\textwidth][c]{93.7} & \makebox[0.005\textwidth][c]{94.1} &  \makebox[0.005\textwidth][c]{96.6} & \makebox[0.005\textwidth][c]{97.7}& \makebox[0.005\textwidth][c]{\textbf{98.6}}\\
    & & \makebox[0.05\textwidth][c]{LoCoOp} & \makebox[0.005\textwidth][c]{88.1} & \makebox[0.005\textwidth][c]{91.6} & \makebox[0.005\textwidth][c]{88.1} & \makebox[0.005\textwidth][c]{91.3} & \makebox[0.005\textwidth][c]{93.7} & \makebox[0.005\textwidth][c]{93.3} & \makebox[0.005\textwidth][c]{90.6} & \makebox[0.005\textwidth][c]{93.5} & \makebox[0.005\textwidth][c]{93.0} & \makebox[0.005\textwidth][c]{93.8} & \makebox[0.005\textwidth][c]{93.3} & \makebox[0.005\textwidth][c]{96.2} & \makebox[0.005\textwidth][c]{93.3} & \makebox[0.005\textwidth][c]{93.5} & \makebox[0.005\textwidth][c]{97.2} & \makebox[0.005\textwidth][c]{98.4}& \makebox[0.005\textwidth][c]{\textbf{98.9}}\\
    & & IDLike &\makebox[0.005\textwidth][c]{83.1} & \makebox[0.005\textwidth][c]{93.2} & \makebox[0.005\textwidth][c]{83.0} & \makebox[0.005\textwidth][c]{89.6} & \makebox[0.005\textwidth][c]{93.8} & \makebox[0.005\textwidth][c]{88.3} & \makebox[0.005\textwidth][c]{92.4} & \makebox[0.005\textwidth][c]{88.5} & \makebox[0.005\textwidth][c]{88.9} & \makebox[0.005\textwidth][c]{88.8} & \makebox[0.005\textwidth][c]{94.8} & \makebox[0.005\textwidth][c]{92.0} & \makebox[0.005\textwidth][c]{88.3} & \makebox[0.005\textwidth][c]{95.0} & \makebox[0.005\textwidth][c]{96.3} & \makebox[0.005\textwidth][c]{97.4}& \makebox[0.005\textwidth][c]{\textbf{98.2}}\\
    & & LoPro &\makebox[0.005\textwidth][c]{80.8}&\makebox[0.005\textwidth][c]{91.6}&\makebox[0.005\textwidth][c]{80.8}&\makebox[0.005\textwidth][c]{89.3}&\makebox[0.005\textwidth][c]{93.8}&\makebox[0.005\textwidth][c]{93.1}&\makebox[0.005\textwidth][c]{90.4}&\makebox[0.005\textwidth][c]{93.3}&\makebox[0.005\textwidth][c]{91.3}&\makebox[0.005\textwidth][c]{91.4}&\makebox[0.005\textwidth][c]{93.2}&\makebox[0.005\textwidth][c]{95.6}&\makebox[0.005\textwidth][c]{93.2}&\makebox[0.005\textwidth][c]{93.6}&\makebox[0.005\textwidth][c]{95.5} &\makebox[0.005\textwidth][c]{97.2} &\makebox[0.005\textwidth][c]{\textbf{98.4}}\\\cmidrule(lr){2-3}\cmidrule(lr){4-8}\cmidrule(lr){9-17}\cmidrule(lr){18-20}
    & \makebox[0.005\textwidth][c]{\multirow{5}{*}{\rotatebox{90}{FPR$_{95}$}}} & CLIP & \makebox[0.005\textwidth][c]{45.9} & \makebox[0.005\textwidth][c]{25.8} & \makebox[0.005\textwidth][c]{45.9} & \makebox[0.005\textwidth][c]{27.1} & \makebox[0.005\textwidth][c]{25.5} & \makebox[0.005\textwidth][c]{23.8} & \makebox[0.005\textwidth][c]{31.4} & \makebox[0.005\textwidth][c]{23.4} & \makebox[0.005\textwidth][c]{26.0} & \makebox[0.005\textwidth][c]{15.5} & \makebox[0.005\textwidth][c]{23.5} & \makebox[0.005\textwidth][c]{11.7} & \makebox[0.005\textwidth][c]{23.7} & \makebox[0.005\textwidth][c]{22.4} & \makebox[0.005\textwidth][c]{13.5} & \makebox[0.005\textwidth][c]{10.0}& \makebox[0.005\textwidth][c]{\textbf{7.4}}\\
    & & CoOp & \makebox[0.005\textwidth][c]{33.5} & \makebox[0.005\textwidth][c]{25.7} & \makebox[0.005\textwidth][c]{33.5} & \makebox[0.005\textwidth][c]{22.7} & \makebox[0.005\textwidth][c]{25.3} & \makebox[0.005\textwidth][c]{25.1} & \makebox[0.005\textwidth][c]{32.3} & \makebox[0.005\textwidth][c]{24.8} & \makebox[0.005\textwidth][c]{24.4} & \makebox[0.005\textwidth][c]{22.7} & \makebox[0.005\textwidth][c]{24.8} & \makebox[0.005\textwidth][c]{16.6} & \makebox[0.005\textwidth][c]{25.1} & \makebox[0.005\textwidth][c]{22.6} & \makebox[0.005\textwidth][c]{13.8} & \makebox[0.005\textwidth][c]{9.5}& \makebox[0.005\textwidth][c]{\textbf{6.3}}\\
    & & \makebox[0.05\textwidth][c]{LoCoOp} & \makebox[0.005\textwidth][c]{36.5} & \makebox[0.005\textwidth][c]{29.8} & \makebox[0.005\textwidth][c]{36.5} & \makebox[0.005\textwidth][c]{24.7} & \makebox[0.005\textwidth][c]{25.2} & \makebox[0.005\textwidth][c]{26.4} & \makebox[0.005\textwidth][c]{33.9} & \makebox[0.005\textwidth][c]{26.1} & \makebox[0.005\textwidth][c]{25.4} & \makebox[0.005\textwidth][c]{22.6} & \makebox[0.005\textwidth][c]{26.3} & \makebox[0.005\textwidth][c]{15.5} & \makebox[0.005\textwidth][c]{26.4} & \makebox[0.005\textwidth][c]{25.8} & \makebox[0.005\textwidth][c]{12.1} & \makebox[0.005\textwidth][c]{6.9}& \makebox[0.005\textwidth][c]{\textbf{4.7}}\\
    & & IDLike &\makebox[0.005\textwidth][c]{44.4} & \makebox[0.005\textwidth][c]{23.5} & \makebox[0.005\textwidth][c]{44.5} & \makebox[0.005\textwidth][c]{28.9} & \makebox[0.005\textwidth][c]{24.9} & \makebox[0.005\textwidth][c]{33.5} & \makebox[0.005\textwidth][c]{26.7} & \makebox[0.005\textwidth][c]{33.1} & \makebox[0.005\textwidth][c]{33.5} & \makebox[0.005\textwidth][c]{31.9} & \makebox[0.005\textwidth][c]{19.3} & \makebox[0.005\textwidth][c]{25.6} & \makebox[0.005\textwidth][c]{33.4} & \makebox[0.005\textwidth][c]{18.7} & \makebox[0.005\textwidth][c]{13.7} & \makebox[0.005\textwidth][c]{9.6}& \makebox[0.005\textwidth][c]{\textbf{7.5}}\\
    & & LoPro &\makebox[0.005\textwidth][c]{45.6}&\makebox[0.005\textwidth][c]{29.2}&\makebox[0.005\textwidth][c]{45.6}&\makebox[0.005\textwidth][c]{27.6}&\makebox[0.005\textwidth][c]{25.3}&\makebox[0.005\textwidth][c]{24.3}&\makebox[0.005\textwidth][c]{31.1}&\makebox[0.005\textwidth][c]{24.1}&\makebox[0.005\textwidth][c]{26.8}&\makebox[0.005\textwidth][c]{24.8}&\makebox[0.005\textwidth][c]{24.1}&\makebox[0.005\textwidth][c]{17.3}&\makebox[0.005\textwidth][c]{24.2}&\makebox[0.005\textwidth][c]{23.2}& \makebox[0.005\textwidth][c]{15.1} & \makebox[0.005\textwidth][c]{9.8}&\makebox[0.005\textwidth][c]{\textbf{7.5}}\\
    \midrule
    \makebox[0.005\textwidth][c]{\multirow{10}{*}{\rotatebox{90}{ObjectNet-104}}} & \makebox[0.005\textwidth][c]{\multirow{5}{*}{\rotatebox{90}{AUC}}} & CLIP & \makebox[0.005\textwidth][c]{92.8} & \makebox[0.005\textwidth][c]{91.0} & \makebox[0.005\textwidth][c]{92.8} & \makebox[0.005\textwidth][c]{89.6} & \makebox[0.005\textwidth][c]{95.1} & \makebox[0.005\textwidth][c]{89.3} & \makebox[0.005\textwidth][c]{84.2} & \makebox[0.005\textwidth][c]{89.6} & \makebox[0.005\textwidth][c]{93.3} & \makebox[0.005\textwidth][c]{98.4} & \makebox[0.005\textwidth][c]{89.4} & \makebox[0.005\textwidth][c]{99.0} & \makebox[0.005\textwidth][c]{89.2} & \makebox[0.005\textwidth][c]{90.4} & \makebox[0.005\textwidth][c]{92.0} & \makebox[0.005\textwidth][c]{93.6}& \makebox[0.005\textwidth][c]{\textbf{99.2}}\\
    & & CoOp & \makebox[0.005\textwidth][c]{95.9} & \makebox[0.005\textwidth][c]{92.8} & \makebox[0.005\textwidth][c]{95.9} & \makebox[0.005\textwidth][c]{92.7} & \makebox[0.005\textwidth][c]{93.7} & \makebox[0.005\textwidth][c]{88.7} & \makebox[0.005\textwidth][c]{82.8} & \makebox[0.005\textwidth][c]{89.0} & \makebox[0.005\textwidth][c]{94.4} & \makebox[0.005\textwidth][c]{97.6} & \makebox[0.005\textwidth][c]{88.9} & \makebox[0.005\textwidth][c]{98.2} & \makebox[0.005\textwidth][c]{88.7} & \makebox[0.005\textwidth][c]{90.2} &  \makebox[0.005\textwidth][c]{92.4} & \makebox[0.005\textwidth][c]{94.2}& \makebox[0.005\textwidth][c]{\textbf{98.7}}\\
    & & \makebox[0.05\textwidth][c]{LoCoOp} & \makebox[0.005\textwidth][c]{89.2} & \makebox[0.005\textwidth][c]{91.0} & \makebox[0.005\textwidth][c]{89.2} & \makebox[0.005\textwidth][c]{88.3} & \makebox[0.005\textwidth][c]{93.5} & \makebox[0.005\textwidth][c]{87.8} & \makebox[0.005\textwidth][c]{81.7} & \makebox[0.005\textwidth][c]{88.1} & \makebox[0.005\textwidth][c]{91.3} & \makebox[0.005\textwidth][c]{93.5} & \makebox[0.005\textwidth][c]{87.9} & \makebox[0.005\textwidth][c]{96.4} & \makebox[0.005\textwidth][c]{87.8} & \makebox[0.005\textwidth][c]{89.0} & \makebox[0.005\textwidth][c]{94.6} & \makebox[0.005\textwidth][c]{95.1}& \makebox[0.005\textwidth][c]{\textbf{98.7}}\\
    & & IDLike &\makebox[0.005\textwidth][c]{89.5} & \makebox[0.005\textwidth][c]{94.4} & \makebox[0.005\textwidth][c]{89.5} & \makebox[0.005\textwidth][c]{88.3} & \makebox[0.005\textwidth][c]{94.5} & \makebox[0.005\textwidth][c]{95.8} & \makebox[0.005\textwidth][c]{85.5} & \makebox[0.005\textwidth][c]{95.9} & \makebox[0.005\textwidth][c]{96.8} & \makebox[0.005\textwidth][c]{93.6} & \makebox[0.005\textwidth][c]{91.8} & \makebox[0.005\textwidth][c]{95.9} & \makebox[0.005\textwidth][c]{95.8} & \makebox[0.005\textwidth][c]{92.7} & \makebox[0.005\textwidth][c]{96.2} & \makebox[0.005\textwidth][c]{97.4}& \makebox[0.005\textwidth][c]{\textbf{98.9}}\\
    & & LoPro &\makebox[0.005\textwidth][c]{91.9}&\makebox[0.005\textwidth][c]{90.4}&\makebox[0.005\textwidth][c]{91.9}&\makebox[0.005\textwidth][c]{89.2}&\makebox[0.005\textwidth][c]{95.1}&\makebox[0.005\textwidth][c]{88.9}&\makebox[0.005\textwidth][c]{83.5}&\makebox[0.005\textwidth][c]{89.1}&\makebox[0.005\textwidth][c]{92.7}&\makebox[0.005\textwidth][c]{97.7}&\makebox[0.005\textwidth][c]{89.0}&\makebox[0.005\textwidth][c]{99.1}&\makebox[0.005\textwidth][c]{88.8}&\makebox[0.005\textwidth][c]{90.1}&\makebox[0.005\textwidth][c]{91.9}& \makebox[0.005\textwidth][c]{93.5}&\makebox[0.005\textwidth][c]{\textbf{99.3}}\\\cmidrule(lr){2-3}\cmidrule(lr){4-8}\cmidrule(lr){9-17}\cmidrule(lr){18-20}
    & \makebox[0.005\textwidth][c]{\multirow{5}{*}{\rotatebox{90}{FPR$_{95}$}}} & CLIP & \makebox[0.005\textwidth][c]{26.0} & \makebox[0.005\textwidth][c]{36.3} & \makebox[0.005\textwidth][c]{26.0} & \makebox[0.005\textwidth][c]{32.1} & \makebox[0.005\textwidth][c]{26.1} & \makebox[0.005\textwidth][c]{40.0} & \makebox[0.005\textwidth][c]{49.5} & \makebox[0.005\textwidth][c]{39.3} & \makebox[0.005\textwidth][c]{28.6} & \makebox[0.005\textwidth][c]{8.1} & \makebox[0.005\textwidth][c]{39.7} & \makebox[0.005\textwidth][c]{4.0} & \makebox[0.005\textwidth][c]{40.0} & \makebox[0.005\textwidth][c]{36.6} & \makebox[0.005\textwidth][c]{26.6} & \makebox[0.005\textwidth][c]{23.9}& \makebox[0.005\textwidth][c]{\textbf{3.5}}\\
    & & CoOp & \makebox[0.005\textwidth][c]{19.7} & \makebox[0.005\textwidth][c]{29.7} & \makebox[0.005\textwidth][c]{19.7} & \makebox[0.005\textwidth][c]{27.3} & \makebox[0.005\textwidth][c]{29.5} & \makebox[0.005\textwidth][c]{44.8} & \makebox[0.005\textwidth][c]{56.1} & \makebox[0.005\textwidth][c]{44.5} & \makebox[0.005\textwidth][c]{27.6} & \makebox[0.005\textwidth][c]{13.7} & \makebox[0.005\textwidth][c]{44.5} & \makebox[0.005\textwidth][c]{9.7} & \makebox[0.005\textwidth][c]{45.0} & \makebox[0.005\textwidth][c]{39.2} & \makebox[0.005\textwidth][c]{27.3} & \makebox[0.005\textwidth][c]{24.2}& \makebox[0.005\textwidth][c]{\textbf{7.2}}\\
    & & \makebox[0.05\textwidth][c]{LoCoOp} & \makebox[0.005\textwidth][c]{39.2} & \makebox[0.005\textwidth][c]{37.0} & \makebox[0.005\textwidth][c]{39.3} & \makebox[0.005\textwidth][c]{36.7} & \makebox[0.005\textwidth][c]{29.6} & \makebox[0.005\textwidth][c]{46.3} & \makebox[0.005\textwidth][c]{57.7} & \makebox[0.005\textwidth][c]{45.8} & \makebox[0.005\textwidth][c]{35.9} & \makebox[0.005\textwidth][c]{27.0} & \makebox[0.005\textwidth][c]{46.2} & \makebox[0.005\textwidth][c]{18.1} & \makebox[0.005\textwidth][c]{46.3} & \makebox[0.005\textwidth][c]{43.3} & \makebox[0.005\textwidth][c]{21.1} & \makebox[0.005\textwidth][c]{20.9}& \makebox[0.005\textwidth][c]{\textbf{7.3}}\\
    & & IDLike &\makebox[0.005\textwidth][c]{36.5} & \makebox[0.005\textwidth][c]{23.4} & \makebox[0.005\textwidth][c]{36.5} & \makebox[0.005\textwidth][c]{36.5} & \makebox[0.005\textwidth][c]{27.2} & \makebox[0.005\textwidth][c]{19.2} & \makebox[0.005\textwidth][c]{47.3} & \makebox[0.005\textwidth][c]{19.1} & \makebox[0.005\textwidth][c]{15.9} & \makebox[0.005\textwidth][c]{26.0} & \makebox[0.005\textwidth][c]{33.4} & \makebox[0.005\textwidth][c]{20.2} & \makebox[0.005\textwidth][c]{19.3} & \makebox[0.005\textwidth][c]{29.7} & \makebox[0.005\textwidth][c]{17.6} & \makebox[0.005\textwidth][c]{12.0}& \makebox[0.005\textwidth][c]{\textbf{5.8}}\\
    & & LoPro &\makebox[0.005\textwidth][c]{26.9}&\makebox[0.005\textwidth][c]{36.9}&\makebox[0.005\textwidth][c]{26.9}&\makebox[0.005\textwidth][c]{32.3}&\makebox[0.005\textwidth][c]{26.2}&\makebox[0.005\textwidth][c]{40.1}&\makebox[0.005\textwidth][c]{49.3}&\makebox[0.005\textwidth][c]{39.7}&\makebox[0.005\textwidth][c]{28.6}&\makebox[0.005\textwidth][c]{10.8}&\makebox[0.005\textwidth][c]{40.0}&\makebox[0.005\textwidth][c]{4.0}&\makebox[0.005\textwidth][c]{40.2}&\makebox[0.005\textwidth][c]{36.7} & \makebox[0.005\textwidth][c]{26.8}& \makebox[0.005\textwidth][c]{24.0}&\makebox[0.005\textwidth][c]{\textbf{3.1}}\\
    \bottomrule
  \end{tabular}}
\end{table*}

\textbf{Variant of Covariate-ID Benchmark.}
Similar to \textit{protocol-1} in \citet{deltaenergy}, which distinguishes closed-set OOD data from open-set OOD data on ImageNet-1K, we adopt the 314-class union of ImageNet-R and ImageNet-A as the covariate-shifted ID classes for our \textbf{setup-iii}. The corresponding 314 classes in ImageNet-Sketch and ImageNet-V2 are collected together with ImageNet-R/-A and treated as closed-set OOD data (\textit{i.e.}, covariate-shifted but semantically overlapping), while the remaining 686 original ImageNet classes are used as open-set OOD data (semantic OODs). Note that our Variant is more challenging, because all selected IDs are covariate-shifted and cover all corruption patterns, whereas OODs is purely semantic shifted. \Cref{tab: covvsseman} further confirms the effectiveness of our method under this setting. The strong performance of CoOD is likely due to CCS, which emphasizes compositional consistency and is less sensitive to appearance variations.

\begin{table*}[h]
\caption{Variant of \textbf{setup-iii} on ImageNet-A/-R/-S/-V2 (covariate IDs) \textit{vs.} original ImageNet (semantic OODs) with various detectors.}
\label{tab: covvsseman}
\centering
{\scriptsize
\begin{tabular}{cccccccccccccccccc}
    \toprule
    \makebox[0.05\textwidth][c]{Metric} & \makebox[0.03\textwidth][c]{MLS} & \makebox[0.03\textwidth][c]{KL} & \makebox[0.03\textwidth][c]{EBO} & \makebox[0.03\textwidth][c]{ViM} & \makebox[0.03\textwidth][c]{ODIN} & \makebox[0.03\textwidth][c]{MCM} & \makebox[0.03\textwidth][c]{Gap} & \makebox[0.03\textwidth][c]{GL} & \makebox[0.03\textwidth][c]{EOE$^*$} & \makebox[0.03\textwidth][c]{NEG$^*$} & \makebox[0.03\textwidth][c]{Dual} & \makebox[0.03\textwidth][c]{CSP$^*$} & \makebox[0.03\textwidth][c]{COOK} & \makebox[0.03\textwidth][c]{$\Delta E$} & \makebox[0.03\textwidth][c]{CoOD-F} & \makebox[0.03\textwidth][c]{CoOD} & \makebox[0.03\textwidth][c]{CoOD$^*$}\\\cmidrule(lr){1-1}\cmidrule(lr){2-6}\cmidrule(lr){7-15}\cmidrule(lr){16-18}
    \makebox[0.005\textwidth][c]{AUC} & \makebox[0.005\textwidth][c]{61.3} & \makebox[0.005\textwidth][c]{49.8} & \makebox[0.005\textwidth][c]{59.4} & \makebox[0.005\textwidth][c]{64.7} & \makebox[0.005\textwidth][c]{58.0} & \makebox[0.005\textwidth][c]{66.3} & \makebox[0.005\textwidth][c]{67.6} & \makebox[0.005\textwidth][c]{69.0} & \makebox[0.005\textwidth][c]{67.0} & \makebox[0.005\textwidth][c]{68.5} & \makebox[0.005\textwidth][c]{65.7} & \makebox[0.005\textwidth][c]{65.0} & \makebox[0.005\textwidth][c]{50.8} & \makebox[0.005\textwidth][c]{63.6} & \makebox[0.005\textwidth][c]{68.3} & \makebox[0.005\textwidth][c]{79.6} & \makebox[0.005\textwidth][c]{\textbf{80.1}} \\       
    \makebox[0.005\textwidth][c]{FPR} & \makebox[0.005\textwidth][c]{87.1} & \makebox[0.005\textwidth][c]{97.9} & \makebox[0.005\textwidth][c]{87.8} & \makebox[0.005\textwidth][c]{80.1} & \makebox[0.005\textwidth][c]{89.3} & \makebox[0.005\textwidth][c]{80.0} & \makebox[0.005\textwidth][c]{78.9} & \makebox[0.005\textwidth][c]{79.3} & \makebox[0.005\textwidth][c]{80.7} & \makebox[0.005\textwidth][c]{82.5} & \makebox[0.005\textwidth][c]{82.2} & \makebox[0.005\textwidth][c]{84.2} & \makebox[0.005\textwidth][c]{90.4} & \makebox[0.005\textwidth][c]{84.9} & \makebox[0.005\textwidth][c]{80.2} & \makebox[0.005\textwidth][c]{58.1} & \makebox[0.005\textwidth][c]{\textbf{55.8}} \\ 
    \bottomrule
  \end{tabular}}
\end{table*}

\textbf{OpenOOD Benchmark.}
Following OpenOOD protocol in \citet{gap}, we evaluate CoOD to detect near OOD from NINCO, far-OOD from ImageNet-O, and ImageNetOOD, under \textbf{setup-i}. The results still demonstrate our clear superiority.
\begin{table*}[h]
\caption{OOD detection on ImageNet1K-OpenOOD with various OOD detectors and models.}
\label{openood}
\centering
{\scriptsize
\begin{tabular}{lcccccccc}
    \toprule
    \makebox[0.07\textwidth][c]{\multirow{2}{*}{Method}} & \multicolumn{2}{c}{\makebox[0.11\textwidth][c]{NINCO}} & \multicolumn{2}{c}{\makebox[0.11\textwidth][c]{ImageNet-O}}& \multicolumn{2}{c}{\makebox[0.11\textwidth][c]{ImageNetOOD}}&\multicolumn{2}{c}{\makebox[0.11\textwidth][c]{AVG}}\\\cmidrule(lr){2-3}\cmidrule(lr){4-5}\cmidrule(lr){6-7}\cmidrule(lr){8-9}
    & \makebox[0.05\textwidth][c]{FPR$_{95}$} & \makebox[0.05\textwidth][c]{AUC} & \makebox[0.05\textwidth][c]{FPR$_{95}$} & \makebox[0.05\textwidth][c]{AUC} & \makebox[0.05\textwidth][c]{FPR$_{95}$} & \makebox[0.05\textwidth][c]{AUC} & \makebox[0.05\textwidth][c]{FPR$_{95}$} & \makebox[0.05\textwidth][c]{AUC}\\    
    \midrule
    EBO & 84.1 & 72.0 & 81.6 & 75.6 & 79.1 & 76.8 & 81.6 & 74.8\\
    \ \ +TAG & 83.1 & 71.1 & 79.7 & 77.1 & 78.3 & 78.0 & 80.4 & 75.4\\
    MCM & 79.6 & 73.6 & 75.9 & 79.5 & 81.0 & 78.3 & 78.8 & 77.1\\
    \ \ +TAG & 81.6 & 71.3 & 77.6 & 79.9 & 83.0 & 78.8 & 80.7 & 76.7\\
    MLS & 79.4 & 74.3 & 77.2 & 77.8 & 75.9 & 78.7 & 77.5 & 77.0\\
    \ \ +TAG & 77.7 & 73.9 & 74.5 & 79.5 & 75.2 & 80.1 & 75.8 & 77.8\\
    GL & 74.4 & 76.0 & 72.4 & 79.5 & 79.2 & 77.3 & 75.3 & 74.7\\
    Gap & 77.4 & 76.5 & 72.0 & 81.4 & 75.4 & 80.3 & 74.9 & 79.4 \\
    CoOD-F  & 75.6 & 74.4 & 54.2 & 82.2 & 58.6 & 81.8 & 62.8 & 79.4\\
    CoOD & 65.8 & 80.0 & 53.5 & \textbf{83.4} & \textbf{51.4} & \textbf{84.6} & \textbf{56.9} & \textbf{82.7}\\
    CoOD$^*$ & \textbf{61.4} & \textbf{81.8} & \textbf{52.6} & 82.9 & 57.0 & 82.9 & 57.0 & 82.5\\
    \midrule
    CoOp & 81.5 & 70.2 & 75.4 & 79.8 & 80.4 & 78.6 & 79.1 & 76.2\\
    \ \ +Gap & 80.7 & 74.0 & 72.7 & 81.1 & 76.1 & 80.2 & 76.5 & 78.4\\
    \ \ +CoOD-F & 81.5 & 70.2 & 62.3 & 80.7 & 65.9 & 79.6 & 69.9 & 76.8\\
    \ \ +CoOD & \textbf{52.2} & 80.6 & \textbf{52.6} & 81.5& 52.4 & 82.6 & \textbf{52.4} & 81.6\\
    \ \ +CoOD$^*$ & 69.3 & 81.1 & 53.9 & 83.0 & 52.3 & 83.1& 58.5& \textbf{82.4}\\
    ID-Like & 81.2 & 71.8 & 82.5 & 68.4 & 88.5 & 62.4 & 84.1 & 67.5\\
    \ \ +Gap & 77.5 & 75.4 & 78.1 & 75.3 & 82.6 & 72.9 & 79.4 & 74.5\\
    \ \ +CoOD-F &74.4&71.2 &57.7& 81.4 &59.4& 80.8 & 63.8& 77.8\\
    \ \ +CoOD &63.9& 79.4 &55.2& \textbf{82.6} &\textbf{52.2}& \textbf{83.9} &57.1& 82.0\\
    \ \ +CoOD$^*$ &60.0& \textbf{81.2} &58.0& 81.6 &56.0& 82.6 &58.0& 81.8\\
    \bottomrule
  \end{tabular}}
\end{table*}

\textbf{Computational Cost.}
The efficiency may be a limitation of original CoOD due to the \textbf{fairness-oriented design}, \textit{e.g.} GradCAM-based component representation rather than open-vocabulary segmentation, traditional keypoint matching rather than registration models \citep{dmad}. 
However, it is important to emphasize that the additional computational burden is justified given the challenge of achieving fine-grained and compositional OOD detection.

We use an EPYC9654-RTX4090 server and parallelize the coreset matching in \cref{ccsmatching}. For a single image, MCM baseline is 89 FPS (+TextEnc=1.37 FPS); full CoOD is 0.85$\sim$1.48 FPS (+TextEnc = 0.56$\sim$0.78 FPS, the value depends on the estimation quality and numbers of the components and keypoint); the faster variant is 85 FPS with comparable performance on ImageNet-1K, though below the full version.

\section{Semantic Text Selection and Component Text Generation}
\label{llm}
We generate component text by prompting ChatGPT-4o: 
\begin{tcolorbox}[
  colback=gray!15,
  colframe=gray!50,
  boxrule=0.5pt
]
From a 360-degree perspective, examine each of the following categories:
`[$\red{'class_1', 'class_2', \ldots}$]`.

For each item:

1. **Identify a compact, non-overlapping set of visually distinct parts or components** that are clearly visible from multiple angles and the visual size is large enough ("flange" is too small for "snorkel"). Focus on features that help **discriminate the object from visually similar categories** using CLIP-style segmentation or recognition.

2. **Build a part \textbf{taxonomy tree}** per object, from general to specific. Select final **leaf-level parts** that are:

   * Highly discriminative
   
   * Commonly visible
   
   * Not too fine-grained or rare
   
   * If necessary, back off to their parent or mid-level nodes for visual reliability.
   
   * If necessary, you can make an analogy for classes that hard to discompose into parts, e.g., chain mail can use parts of clothing, such as: "sleeves", "hem", etc.

3. **Clean the part names** by removing or summarizing:

   * Descriptors (e.g. “small,” “left,” “top”)
   * Redundant modifiers
   * Directional/locational terms
   * Duplicate terms

4. **Avoid excessive part lists**:

   * **4$\sim$6 parts per object** is preferred.
   * **More than 4** is allowed if all are visually salient.
   * **Fewer than 4** is conditionally acceptable **only if justified**, e.g., the object is visually minimal (e.g. “golf ball”).

5. Use **natural, social-media-friendly English terms** for parts (e.g. “handle” not “manubrium”).

   * If two or more part names refer to nearly the same or visually proximate structure, combine them using the format: **“limb-paw”**. Only merge parts that are visually indistinct or spatially inseparable.
   
Avoid generating overly technical or obscure terms. Keep focus on what is **recognizable by a non-expert viewer**.

Avoid phrases if a single word is clear and discriminative.

6. The results should be in the format of python dict, e.g. for class name of "cat", we have caption = \{"cat": ["head", "limb-paw", "tail", "trunk"],\} 
\end{tcolorbox}

Class names and the corresponding component names are as follows: 
\begin{tcolorbox}[
  colback=gray!15,
  colframe=gray!50,
  boxrule=0.5pt
]
\textbf{ImageNet}

\textbf{Class/Component names}: PART\_LABEL = \{
    "snorkel": ["mouthpiece", "tube", "glasses", "face"],
    "syringe": ["barrel", "plunger", "needle hub", "needle"],
    "binoculars": ["objective lens", "eyepiece", "focus knob", "bridge"],
    "cannon": ["barrel", "wheels", "carriage", "breech"],
    "crane\_machine": ["boom", "counterweight", "cab", "hoist", "mast"],
    "desktop\_computer": ["tower case", "power button", "ports panel", "vent"],
    "gas\_pump": ["nozzle", "hose", "display", "handle"],
    "muzzle": ["strap", "cover", "head", "snout"],
    "seat\_belt": ["webbing", "buckle", "seat", "human"],
    "patio": ["paving", "furniture", "canopy", "plant"],
    "home\_theater": ["screen", "speakers", "table", "couch"],
    "totem\_pole": ["base", "carved figures", "top figure"],
    "grille": ["body", "light", "license plate", "emblem"],
    "mountain\_tent": ["canopy", "entrance", "rainfly", "guylines"],
    "scoreboard": ["display panel", "frame", "supports", "digits"],
    "cocktail\_shaker": ["tin", "strainer", "cap", "body"],
    "kimono": ["sleeves", "collar", "obi"],
    "whiskey\_jug": ["body", "neck", "handle", "stopper"],
    "knee\_pad": ["pad", "straps", "leg"],
    "book\_jacket": ["cover", "pages", "title", "spine"],
    "crash\_helmet": ["shell", "visor", "chin strap", "padding"],
    "vestment": ["chasuble", "stole", "cuffs"],
    "cloak": ["hood", "body", "hem"],
    "scabbard": ["sheath", "belt loop", "mouth", "handle-grip", "blade-knife"],
    "beer\_glass": ["rim", "body", "base"],
    "swab": ["shaft", "tip-head", "handle", "strand"],
    "drilling\_platform": ["sea", "tower", "supports-legs", "deck"],
    "pencil\_box": ["lid", "body", "hinge", "zipper"],
    "punching\_bag": ["bag body", "chain strap", "mount"],
    "pencil\_sharpener": ["body", "handle", "blade", "shavings receptacle"],
    "shower\_cap": ["cap body", "head", "hair", "elastic band"],
    "trolleybus": ["pantograph", "body", "wheels", "windows"],
    "perfume": ["bottle", "spray nozzle", "cap"],
    "crate": ["slats", "frame"],
    "ballpoint": ["tip", "barrel", "cap"],
    "comic\_book": ["cover", "pages", "title", "spine"],
    "wooden\_spoon": ["bowl", "handle"],
    "ice\_lolly": ["stick", "frozen block", "head", "hand"],
    "carbonara": ["spaghetti", "sauce", "marinara", "bolognese", "plate"],
    "caldron": ["body", "handle", "lid", "legs"],
    "backpack": ["compartment", "pocket", "strap", "handle"],
    "banana": ["peel", "flesh", "stem", "tip"],
    "Band\_Aid": ["pad", "adhesive", "backing"],
    "shopping\_basket": ["body", "handle", "rim", "base", "mesh"],
    "bath\_towel": ["body", "hem", "hanger loop", "border"],
    "beer\_bottle": ["neck", "body", "label", "cap"],
    "park\_bench": ["seat", "backrest", "armrest", "legs"],
    "binder": ["cover", "spine", "rings", "label"],
    "bottlecap": ["top", "side wall", "liner", "bottle"],
    "French\_loaf": ["crust", "scored-slashes", "tip", "slice-face", "wrapper", "basket"],
    "broom": ["handle", "brush-bristles", "ferrule", "shaft", "tip-head", "strand"],
    "bucket": ["body", "handle", "rim", "base"],
    "cleaver": ["blade", "table", "board", "handle"],
    "can\_opener": ["handle", "cutting-wheel", "gear", "arm", "can"],
    "candle": ["body", "wick", "drip", "base"],
    "cellular\_telephone": ["screen", "panel", "camera", "buttons", "speaker"],
    "hamper": ["basket-body", "lid", "handles", "base"],
    "espresso\_maker": ["body", "group-head", "portafilter", "steam-wand", "water-tank"],
    "combination\_lock": ["dial", "body", "shackle", "number ring"],
    "mouse": ["buttons", "scroll-wheel", "body", "wire"],
    "table\_lamp": ["base", "stem", "lampshade", "bulb socket", "wire"],
    "dishrag": ["dish", "body", "hem", "hanger-loop"],
    "doormat": ["door", "surface", "edge", "backing"],
    "Loafer": ["upper", "sole", "heel", "vamp"],
    "power\_drill": ["chuck", "body", "trigger", "handle", "battery pack", "wire"],
    "cup": ["rim", "body", "handle", "base"],
    "plate\_rack": ["slots", "frame", "base", "plate"] ,
    "envelope": ["body", "flap", "seal area", "window"],
    "electric\_fan": ["blades", "grill", "motor", "base"],
    "frying\_pan": ["body", "handle", "base", "rim"],
    "gown": ["bodice", "skirt", "sleeves", "neckline"],
    "hand\_blower": ["nozzle", "barrel", "handle", "air-intake", "wire"],
    "hammer": ["head", "claw", "face", "handle"],
    "iron": ["soleplate", "tank", "handle", "control panel"],
    "jean": ["waistband", "legs", "pockets", "fly"],
    "computer\_keyboard": ["keycaps", "frame", "spacebar", "function-row", "numpad"],
    "ladle": ["bowl", "handle", "hook"],
    "lampshade": ["shade", "rim", "fitting", "lining"],
    "laptop": ["screen", "keyboard", "touchpad", "hinge", "base"],
    "lemon": ["peel", "pulp", "stem-end"],
    "letter\_opener": ["blade", "handle", "point-tip"],
    "lighter": ["body", "nozzle", "wheel igniter", "fuel window"],
    "lipstick": ["cap", "tube", "bullet", "base"],
    "matchstick": ["head", "shaft", "box"],
    "measuring\_cup": ["body", "handle", "spout", "markings", "table"],
    "microwave": ["door", "control panel", "turntable", "cavity"],
    "mixing\_bowl": ["bowl", "rim", "base"],
    "monitor": ["screen", "bezel", "stand", "port", "wire"],
    "coffee\_mug": ["handle", "rim", "body", "base", "plate"],
    "nail": ["head", "shank-body", "tip-point"],
    "necklace": ["chain", "pendant", "clasp", "neck"],
    "orange": ["peel", "segments", "stem-end", "seed"],
    "padlock": ["body", "shackle", "keyway"],
    "paintbrush": ["bristles", "ferrule", "handle"],
    "paper\_towel": ["sheet", "perforation", "pack"],
    "pill\_bottle": ["cap", "body", "label", "rim"],
    "pillow": ["shell", "filling", "seam", "bed", "head"],
    "pitcher": ["spout", "handle", "body", "base"],
    "plastic\_bag": ["body", "handles", "gusset"],
    "plate": ["rim", "center", "underside"],
    "plunger": ["handle", "rubber cup", "shaft"],
    "pop\_bottle": ["neck", "cap", "body", "label", "base"],
    "space\_heater": ["grill", "control panel", "housing", "base"],
    "printer": ["paper-tray", "output-tray", "control-panel", "cartridge"],
    "remote\_control": ["buttons", "directional-pad", "battery compartment", "infrared emitter"],
    "rule": ["edge", "markings", "body"],
    "running\_shoe": ["upper", "sole", "tongue", "laces"],
    "safety\_pin": ["pin", "clasp", "spring"],
    "saltshaker": ["body", "cap", "base", "holes"],
    "sandal": ["sole", "straps", "buckle", "footbed"],
    "screw": ["head", "shank", "thread", 
            \end{tcolorbox}

\begin{tcolorbox}[
  colback=gray!15,
  colframe=gray!50,
  boxrule=0.5pt
]
"point"],
    "shovel": ["blade", "shaft", "grip", "collar"],
    "sleeping\_bag": ["shell", "zipper", "hood", "insulation"],
    "soap\_dispenser": ["pump", "bottle", "nozzle", "base"],
    "sock": ["cuff", "arch", "toe", "heel"],
    "soup\_bowl": ["rim", "bowl", "base"],
    "spatula": ["blade", "handle", "neck"],
    "loudspeaker": ["cone", "grille", "cabinet", "mount", "magnet"],
    "strainer": ["mesh", "rim", "handle", "hook"],
    "teddy": ["head", "limbs", "body", "face"],
    "suit": ["jacket", "trousers", "lapel", "buttons"],
    "sunglasses": ["lenses", "frame", "temples", "bridge"],
    "sweatshirt": ["hood", "pocket", "cuffs", "hem"],
    "swimming\_trunks": ["waistband", "leg openings", "pocket", "drawstring"],
    "jersey": ["pocket", "sleeves", "collar", "logo-number", "hem"],
    "television": ["screen", "bezel", "stand", "back panel"],
    "teapot": ["body", "spout", "handle", "lid"],
    "racket": ["head", "stringbed", "frame", "handle"],
    "toaster": ["slots", "lever", "body", "tray", "grille", "knobs"],
    "toilet\_tissue": ["tube", "paper", "perforation"],
    "ashcan": ["lid", "body", "rim", "foot pedal"],
    "tray": ["flat surface", "rim", "handles"],
    "umbrella": ["canopy", "ribs", "shaft", "handle"],
    "vacuum": ["hose", "housing", "brush-head", "handle"],
    "vase": ["neck", "body", "base", "rim"],
    "wallet": ["compartment", "card-slots", "coin-pocket", "fold"],
    "digital\_watch": ["face", "strap", "buttons", "buckle"],
    "water\_bottle": ["cap", "neck", "body", "base"],
    "dumbbell": ["handle", "weight-plates", "collar"],
    "scale": ["platform", "display", "buttons", "base"],
"whistle": ["body", "mouthpiece", "pea", "ring"],
    "wine\_bottle": ["neck", "body", "closure", "label"],
    "mitten": ["palm", "thumb", "cuff", "insulation"],
    "wok": ["bowl", "handle", "base"],
 "tench": ["head", "dorsal fin", "tail", "pectoral fins", "gills"],
 "goldfish": ["head", "dorsal fin", "tail", "pectoral fins", "eyes"],
 "great\_white\_shark": ["snout", "dorsal fin", "tail", "pectoral fins", "gill slits", "teeth"],
 "tiger\_shark": ["snout", "dorsal fin", "tail", "pectoral fins", "gill slits", "teeth"],
 "hammerhead": ["hammer head", "dorsal fin", "tail", "pectoral fins", "eyes", "teeth"],
 "electric\_ray": ["pectoral disc", "tail", "eyes", "mouth", "pectoral fin"],
 "stingray": ["pectoral disc", "tail barb", "eyes", "mouth", "pectoral fin"],
 "cock": ["comb", "wattle", "beak", "wings", "tail", "legs"],
 "hen": ["comb", "wattle", "beak", "wings", "tail", "legs"],
 "ostrich": ["head", "neck", "wings", "legs"],
 "brambling": ["beak", "head", "wings", "tail"],
 "goldfinch": ["beak", "head", "wings", "tail"],
 "house\_finch": ["beak", "head", "wings", "tail"],
 "junco": ["beak", "head", "wings", "tail"],
 "indigo\_bunting": ["beak", "head", "wings", "tail"],
 "robin": ["beak", "head", "wings", "tail"],
 "bulbul": ["beak", "head", "crest", "wings", "tail"],
 "jay\_bird": ["beak", "head", "crest", "wings", "tail"],
 "crane\_bird": ["beak", "head", "crest", "wings", "tail"],
 "magpie": ["beak", "head", "long tail", "wings"],
 "chickadee": ["beak", "head", "wings", "tail"],
 "water\_ouzel": ["beak", "head", "wings", "tail"],
 "kite": ["beak", "head", "wings", "talons", "tail"],
 "bald\_eagle": ["beak", "head", "wings", "talons", "tail"],
 "vulture": ["beak", "head", "neck", "wings", "legs"],
 "great\_grey\_owl": ["face disc", "beak", "wings", "talons", "tail"],
 "European\_fire\_salamander": ["head", "limbs", "tail", "eyes"],
 "common\_newt": ["head", "limbs", "tail", "eyes"],
 "eft": ["head", "limbs", "tail", "eyes"],
 "spotted\_salamander": ["head", "limbs", "tail", "spots"],
 "axolotl": ["head", "gills", "limbs", "tail"],
 "bullfrog": ["head", "eyes", "hind legs", "mouth"],
 "tree\_frog": ["head", "toe pads", "hind legs", "eyes"],
 "tailed\_frog": ["head", "tail organ", "hind legs", "eyes"],
 "loggerhead": ["head", "shell", "flippers", "tail", "belly"],
 "leatherback\_turtle": ["head", "leathery shell", "flippers", "tail", "belly"],
 "mud\_turtle": ["head", "shell", "legs", "tail", "belly"],
 "terrapin": ["head", "shell", "legs", "tail", "belly"],
 "box\_turtle": ["head", "domed shell", "legs", "tail", "belly"],
 "banded\_gecko": ["head", "legs", "tail", "eyes"],
 "common\_iguana": ["head crest", "dorsal crest", "legs", "tail", "dewlap"],
 "American\_chameleon": ["head helmet", "prehensile tail", "feet", "eyes"],
 "whiptail": ["head", "legs", "tail"],
 "agama": ["head", "dorsal crest", "legs", "tail"],
 "frilled\_lizard": ["head", "frill", "legs", "tail"],
 "alligator\_lizard": ["head", "legs", "tail"],
 "Gila\_monster": ["head", "legs", "tail"],
 "green\_lizard": ["head", "legs", "tail"],
 "African\_chameleon": ["head helmet", "prehensile tail", "feet", "eyes"],
 "Komodo\_dragon": ["head", "neck", "legs", "tail"],
 "African\_crocodile": ["snout", "eyes and nostrils", "legs", "tail"],
 "American\_alligator": ["snout", "eyes and nostrils", "legs", "tail"],
 "triceratops": ["head frill", "horns", "beak", "legs"],
 "thunder\_snake": ["head", "body", "belly", "tail"],
 "ringneck\_snake": ["head", "neck ring", "body", "tail"],
 "hognose\_snake": ["upturned snout", "head", "body pattern", "tail"],
 "green\_snake": ["head", "slender body", "tail"],
 "king\_snake": ["head", "banded pattern", "body", "tail"],
 "garter\_snake": ["head", "striped body", "tail"],
 "water\_snake": ["head", "body pattern", "tail"],
 "vine\_snake": ["head", "very slender body", "eyes", "tail"],
 "night\_snake": ["head", "body pattern", "tail"],
 "boa\_constrictor": ["head", "muscular body", "tail"],
 "rock\_python": ["head", "heavy body", "tail"],
 "Indian\_cobra": ["head", "hood", "fangs", "tail"],
 "green\_mamba": ["head", "slender body", "fangs", "tail"],
 "sea\_snake": ["head", "flattened tail", "vent", "tail tip"],
 "horned\_viper": ["head", "horn", "fangs", "tail"],
 "diamondback": ["head", "rattle", "fangs", "tail"],
 "sidewinder": ["head", "rattle", "belly scales", "tail"],
 "trilobite": ["head shield", "thoracic segments", "tail shield", "compound eyes"],
 "harvestman": ["body", "legs", "mouthparts"],
 "scorpion": ["pincers", "body", "tail", "legs", "stinger"],
 "black\_and\_gold\_garden\_spider": ["head", "abdomen", "legs", "spinnerets"],
 "barn\_spider": ["head", "abdomen", "legs", "spinnerets"],
 "garden\_spider": ["head", "abdomen", "legs", "spinnerets"],
 "black\_widow": ["head", "abdomen", "legs", "spinnerets"],
 "tarantula": ["head", "abdomen", "thick legs", "pedipalps"],
 "wolf\_spider": ["head", 
         \end{tcolorbox}

\begin{tcolorbox}[
  colback=gray!15,
  colframe=gray!50,
  boxrule=0.5pt
]
 "abdomen", "legs", "eye cluster"],
 "tick": ["body", "mouthparts", "legs"],
 "centipede": ["head", "body segments", "many legs", "claws"],
 "black\_grouse": ["beak", "head comb", "wings", "tail"],
 "ptarmigan": ["beak", "head", "wings", "tail"],
 "ruffed\_grouse": ["beak", "head", "ruff", "wings"],
 "prairie\_chicken": ["beak", "air sacs", "wings", "tail"],
 "peacock": ["head", "crest", "display tail", "wings"],
 "quail": ["beak", "head crest", "wings", "tail"],
 "partridge": ["beak", "head", "wings", "tail"],
 "African\_grey": ["beak", "head", "wings", "tail"],
 "macaw": ["beak", "head", "wings", "long tail", "feet"],
 "sulphur-crested\_cockatoo": ["crest", "beak", "head", "wings", "tail"],
 "lorikeet\_parrot": ["beak", "head", "wings", "tail"],
 "coucal\_bird": ["beak", "head", "long tail", "wings"],
 "bee\_eater": ["long beak", "head", "wings", "tail"],
"hornbill": ["beak", "head", "wings", "tail", "legs"],
"hummingbird": ["beak", "head", "wings", "tail"],
"jacamar": ["beak", "head", "wings", "tail"],
"toucan": ["beak", "head", "wings", "tail"],
"drake": ["beak", "head", "wings", "tail", "feet"],
"red-breasted\_merganser": ["beak", "head", "wings", "tail", "feet"],
"goose": ["beak", "head", "wings", "tail", "feet"],
"black\_swan": ["bill", "head", "neck", "wings", "feet"],
"tusker": ["head", "tusks", "trunk", "ears", "legs"],
"echidna": ["snout", "spines", "legs", "eyes"],
"platypus": ["bill", "webbed feet", "tail", "legs"],
"wallaby": ["head", "ears", "forepaws", "hind legs", "tail"],
"koala": ["head", "ears", "forepaws", "hind legs", "tail"],
"wombat": ["head", "ears", "forepaws", "hind legs", "tail"],
"jellyfish": ["bell", "tentacles", "mouth"],
"sea\_anemone": ["base", "tentacles", "mouth"],
"brain\_coral": ["lobes", "ridges", "valleys"],
"flatworm": ["head", "tail", "body"],
"nematode": ["head", "tail", "body"],
"conch": ["shell", "aperture", "foot", "tentacles"],
"snail": ["shell", "aperture", "foot", "tentacles"],
"land\_slug": ["head", "mantle", "foot", "tentacles"],
"sea\_slug": ["head", "mantle", "gills", "foot"],
"chiton": ["shell plates", "girdle", "foot"],
"chambered\_nautilus": ["shell", "tentacles", "hood"],
"Dungeness\_crab": ["shell", "claws", "walking legs", "antennae"],
"rock\_crab": ["shell", "claws", "walking legs", "antennae"],
"fiddler\_crab": ["shell", "major claw", "walking legs", "eye stalks"],
"king\_crab": ["shell", "large legs", "claws", "antennae"],
"American\_lobster": ["shell", "claws", "tail fan", "walking legs", "antennae"],
"spiny\_lobster": ["shell", "antennae", "tail fan", "walking legs"],
"crayfish": ["shell", "claws", "tail fan", "walking legs", "antennae"],
"hermit\_crab": ["shell", "claws", "walking legs", "antennae"],
"isopod": ["body segments", "legs", "antennae"],
"white\_stork": ["beak", "head", "neck", "wings", "legs"],
"black\_stork": ["beak", "head", "neck", "wings", "legs"],
"spoonbill": ["bill", "head", "neck", "wings", "legs"],
"flamingo": ["bill", "head", "neck", "wings", "legs"],
"little\_blue\_heron": ["beak", "head", "neck", "wings", "legs"],
"American\_egret": ["beak", "head", "neck", "wings", "legs"],
"bittern": ["beak", "head", "neck", "wings", "legs"],
"limpkin": ["beak", "head", "neck", "wings", "legs"],
"European\_gallinule": ["beak", "head", "wings", "legs", "tail"],
"American\_coot": ["bill", "head", "wings", "legs", "feet"],
"bustard": ["beak", "head", "wings", "legs", "tail"],
"ruddy\_turnstone": ["beak", "head", "wings", "legs", "tail"],
"red-backed\_sandpiper": ["beak", "head", "wings", "legs", "tail"],
"redshank": ["beak", "head", "wings", "legs", "tail"],
"dowitcher": ["beak", "head", "wings", "legs", "tail"],
"oystercatcher": ["beak", "head", "wings", "legs", "tail"],
"pelican": ["beak", "head", "throat pouch", "wings", "feet"],
"king\_penguin": ["beak", "head", "flippers", "feet", "tail"],
"albatross": ["beak", "head", "wings", "tail", "legs"],
"grey\_whale": ["head", "blowhole", "flippers", "tail flukes", "dorsal ridge"],
"killer\_whale": ["head", "dorsal fin", "flippers", "tail flukes"],
"dugong": ["head", "flippers", "tail flukes"],
"sea\_lion": ["head", "flippers", "whiskers", "tail"],
"Chihuahua": ["head", "ears", "muzzle", "body", "legs", "tail"],
"Japanese\_spaniel": ["head", "ears", "muzzle", "body", "legs", "tail"],
"Maltese\_dog": ["head", "ears", "muzzle", "body", "legs", "tail"],
"Pekinese": ["head", "ears", "muzzle", "body", "legs", "tail"],
"Shih-Tzu": ["head", "ears", "muzzle", "body", "legs", "tail"],
"Blenheim\_spaniel": ["head", "ears", "muzzle", "body", "legs", "tail"],
"papillon": ["head", "ears", "muzzle", "body", "legs", "tail"],
"toy\_terrier": ["head", "ears", "muzzle", "body", "legs", "tail"],
"Rhodesian\_ridgeback": ["head", "ears", "muzzle", "body", "legs", "tail"],
"Afghan\_hound": ["head", "ears", "muzzle", "body", "legs", "tail"],
"basset": ["head", "ears", "muzzle", "body", "legs", "tail"],
"beagle": ["head", "ears", "muzzle", "body", "legs", "tail"],
"bloodhound": ["head", "ears", "muzzle", "body", "legs", "tail"],
"bluetick": ["head", "ears", "muzzle", "body", "legs", "tail"],
"black-and-tan\_coonhound": ["head", "ears", "muzzle", "body", "legs", "tail"],
"Walker\_hound": ["head", "ears", "muzzle", "body", "legs", "tail"],
"English\_foxhound": ["head", "ears", "muzzle", "body", "legs", "tail"],
"redbone": ["head", "ears", "muzzle", "body", "legs", "tail"],
"borzoi": ["head", "ears", "muzzle", "body", "legs", "tail"],
"Irish\_wolfhound": ["head", "ears", "muzzle", "body", "legs", "tail"],
"Italian\_greyhound": ["head", "ears", "muzzle", "body", "legs", "tail"],
"whippet": ["head", "ears", "muzzle", "body", "legs", "tail"],
"Ibizan\_hound": ["head", "ears", "muzzle", "body", "legs", "tail"],
"Norwegian\_elkhound": ["head", "ears", "muzzle", "body", "legs", "tail"],
"otterhound": ["head", "ears", "muzzle", "body", "legs", "tail"],
"Saluki": ["head", "ears", "muzzle", "body", "legs", "tail"],
"Scottish\_deerhound": ["head", "ears", "muzzle", "body", "legs", "tail"],
"Weimaraner": ["head", "ears", "muzzle", "body", "legs", "tail"],
"Staffordshire\_bullterrier": ["head", "ears", "muzzle", "body", "legs", "tail"],
"American\_Staffordshire\_terrier": ["head", "ears", "muzzle", "body", "legs", "tail"],
"Bedlington\_terrier": ["head", "ears", "muzzle", "body", "legs", "tail"],
"Border\_terrier": 
        \end{tcolorbox}

\begin{tcolorbox}[
  colback=gray!15,
  colframe=gray!50,
  boxrule=0.5pt
]
["head", "ears", "muzzle", "body", "legs", "tail"],
"Kerry\_blue\_terrier": ["head", "ears", "muzzle", "body", "legs", "tail"],
"Irish\_terrier": ["head", "ears", "muzzle", "body", "legs", "tail"],
"Norfolk\_terrier": ["head", "ears", "muzzle", "body", "legs", "tail"],
"Norwich\_terrier": ["head", "ears", "muzzle", "body", "legs", "tail"],
"Yorkshire\_terrier": ["head", "ears", "muzzle", "body", "legs", "tail"],
"wire-haired\_fox\_terrier": ["head", "ears", "muzzle", "body", "legs", "tail"],
"Lakeland\_terrier": ["head", "ears", "muzzle", "body", "legs", "tail"],
"Sealyham\_terrier": ["head", "ears", "muzzle", "body", "legs", "tail"],
"Airedale": ["head", "ears", "muzzle", "body", "legs", "tail"],
"cairn": ["head", "ears", "muzzle", "body", "legs", "tail"],
"Australian\_terrier": ["head", "ears", "muzzle", "body", "legs", "tail"],
"Dandie\_Dinmont": ["head", "ears", "muzzle", "body", "legs", "tail"],
"Boston\_bull": ["head", "ears", "muzzle", "body", "legs", "tail"],
"miniature\_schnauzer": ["head", "ears", "muzzle", "body", "legs", "tail"],
"giant\_schnauzer": ["head", "ears", "muzzle", "body", "legs", "tail"],
"standard\_schnauzer": ["head", "ears", "muzzle", "body", "legs", "tail"],
"Scotch\_terrier": ["head", "ears", "muzzle", "body", "legs", "tail"],
"Tibetan\_terrier": ["head", "ears", "muzzle", "body", "legs", "tail"],
"silky\_terrier": ["head", "ears", "muzzle", "body", "legs", "tail"],
"soft-coated\_wheaten\_terrier": ["head", "ears", "muzzle", "body", "legs", "tail"],
"West\_Highland\_white\_terrier": ["head", "ears", "muzzle", "body", "legs", "tail"],
"Lhasa": ["head", "ears", "muzzle", "body", "legs", "tail"],
"flat-coated\_retriever": ["head", "ears", "muzzle", "body", "legs", "tail"],
"curly-coated\_retriever": ["head", "ears", "muzzle", "body", "legs", "tail"],
"golden\_retriever": ["head", "ears", "muzzle", "body", "legs", "tail"],
"Labrador\_retriever": ["head", "ears", "muzzle", "body", "legs", "tail"],
"Chesapeake\_Bay\_retriever": ["head", "ears", "muzzle", "body", "legs", "tail"],
"German\_short-haired\_pointer": ["head", "ears", "muzzle", "body", "legs", "tail"],
"vizsla": ["head", "ears", "muzzle", "body", "legs", "tail"],
"English\_setter": ["head", "ears", "muzzle", "body", "legs", "tail"],
"Irish\_setter": ["head", "ears", "muzzle", "body", "legs", "tail"],
"Gordon\_setter": ["head", "ears", "muzzle", "body", "legs", "tail"],
"Brittany\_spaniel": ["head", "ears", "muzzle", "body", "legs", "tail"],
"clumber": ["head", "ears", "muzzle", "body", "legs", "tail"],
"English\_springer": ["head", "ears", "muzzle", "body", "legs", "tail"],
"Welsh\_springer\_spaniel": ["head", "ears", "muzzle", "body", "legs", "tail"],
"cocker\_spaniel": ["head", "ears", "muzzle", "body", "legs", "tail"],
"Sussex\_spaniel": ["head", "ears", "muzzle", "body", "legs", "tail"],
"Irish\_water\_spaniel": ["head", "ears", "muzzle", "body", "legs", "tail"],
"kuvasz": ["head", "ears", "muzzle", "body", "legs", "tail"],
"schipperke": ["head", "ears", "muzzle", "body", "legs", "tail"],
"groenendael": ["head", "ears", "muzzle", "body", "legs", "tail"],
"malinois": ["head", "ears", "muzzle", "body", "legs", "tail"],
"briard": ["head", "ears", "muzzle", "body", "legs", "tail"],
"kelpie": ["head", "ears", "muzzle", "body", "legs", "tail"],
"komondor": ["head", "ears", "muzzle", "body", "legs", "tail"],
"Old\_English\_sheepdog": ["head", "ears", "muzzle", "body", "legs", "tail"],
"Shetland\_sheepdog": ["head", "ears", "muzzle", "body", "legs", "tail"],
"collie": ["head", "ears", "muzzle", "body", "legs", "tail"],
"Border\_collie": ["head", "ears", "muzzle", "body", "legs", "tail"],
"Bouvier\_des\_Flandres": ["head", "ears", "muzzle", "body", "legs", "tail"],
"Rottweiler": ["head", "ears", "muzzle", "body", "legs", "tail"],
"German\_shepherd": ["head", "ears", "muzzle", "body", "legs", "tail"],
"Doberman": ["head", "ears", "muzzle", "body", "legs", "tail"],
"miniature\_pinscher": ["head", "ears", "muzzle", "body", "legs", "tail"],
"Greater\_Swiss\_Mountain\_dog": ["head", "ears", "muzzle", "body", "legs", "tail"],
"Bernese\_mountain\_dog": ["head", "ears", "muzzle", "body", "legs", "tail"],
"Appenzeller": ["head", "ears", "muzzle", "body", "legs", "tail"],
"EntleBucher": ["head", "ears", "muzzle", "body", "legs", "tail"],
"boxer": ["head", "ears", "muzzle", "body", "legs", "tail"],
"bull\_mastiff": ["head", "ears", "muzzle", "body", "legs", "tail"],
"Tibetan\_mastiff": ["head", "ears", "muzzle", "body", "legs", "tail"],
"French\_bulldog": ["head", "ears", "muzzle", "body", "legs", "tail"],
"Great\_Dane": ["head", "ears", "muzzle", "body", "legs", "tail"],
"Saint\_Bernard": ["head", "ears", "muzzle", "body", "legs", "tail"],
"Eskimo\_dog": ["head", "ears", "muzzle", "body", "legs", "tail"],
"malamute": ["head", "ears", "muzzle", "body", "legs", "tail"],
"Siberian\_husky": ["head", "ears", "muzzle", "body", "legs", "tail"],
"dalmatian": ["head", "ears", "muzzle", "body", "legs", "tail"],
"affenpinscher": ["head", "ears", "muzzle", "body", "legs", "tail"],
"basenji": ["head", "ears", "muzzle", "body", "legs", "tail"],
"pug": ["head", "ears", "muzzle", "body", "legs", "tail"],
"Leonberg": ["head", "ears", "muzzle", "body", "legs", "tail"],
"Newfoundland": ["head", "ears", "muzzle", "body", "legs", "tail"],
"Great\_Pyrenees": ["head", "ears", "muzzle", "body", "legs", "tail"],
"Samoyed": ["head", "ears", "muzzle", "body", "legs", "tail"],
 "Pomeranian": ["head", "ears", "muzzle", "body", "legs", "tail"],
 "chow": ["head", "ears", "muzzle", "body", "legs", "tail"],
 "keeshond": ["head", "ears", "muzzle", "body", "legs", "tail"],
 "Brabancon\_griffon": ["head", "ears", "muzzle", "body", "legs", "tail"],
 "Pembroke": ["head", "ears", "muzzle", "body", "legs", "tail"],
 "Cardigan": ["head", "ears", "muzzle", "body", "legs", "tail"],
 "toy\_poodle": ["head", "ears", "muzzle", "body", "legs", "tail"],
 "miniature\_poodle": ["head", "ears", "muzzle", "body", "legs", "tail"],
 "standard\_poodle": ["head", "ears", "muzzle", "body", "legs", "tail"],
 "Mexican\_hairless": ["head", "ears", "muzzle", "body", "legs", "tail"],
 "timber\_wolf": ["head", "ears", "muzzle", "body", "legs", "tail"],
 "white\_wolf": ["head", "ears", "muzzle", "body", "legs", "tail"],
 "red\_wolf": ["head", "ears", "muzzle", "body", "legs", "tail"],
 "coyote": ["head", "ears", "muzzle", "body", "legs", "tail"],
 "dingo": ["head", "ears", "muzzle", 
         \end{tcolorbox}

\begin{tcolorbox}[
  colback=gray!15,
  colframe=gray!50,
  boxrule=0.5pt
]
"body", "legs", "tail"],
 "dhole\_dog": ["head", "ears", "muzzle", "body", "legs", "tail"],
 "African\_hunting\_dog": ["head", "ears", "muzzle", "body", "legs", "tail"],
 "hyena": ["head", "ears", "muzzle", "body", "legs", "tail"],
 "red\_fox": ["head", "ears", "muzzle", "body", "legs", "tail"],
 "kit\_fox": ["head", "ears", "muzzle", "body", "legs", "tail"],
 "Arctic\_fox": ["head", "ears", "muzzle", "body", "legs", "tail"],
 "grey\_fox": ["head", "ears", "muzzle", "body", "legs", "tail"],
 "tabby\_cat": ["head", "ears", "muzzle", "body", "legs", "tail"],
 "tiger\_cat": ["head", "ears", "muzzle", "body", "legs", "tail"],
 "Persian\_cat": ["head", "ears", "eyes", "body", "tail"],
 "Siamese\_cat": ["head", "ears", "eyes", "body", "tail"],
 "Egyptian\_cat": ["head", "ears", "eyes", "body", "tail"],
 "cougar": ["head", "ears", "eyes", "body", "tail"],
 "lynx": ["head", "ears", "eyes", "body", "tail"],
 "leopard": ["head", "ears", "eyes", "body", "tail"],
 "snow\_leopard": ["head", "ears", "eyes", "body", "tail"],
 "jaguar": ["head", "ears", "eyes", "body", "tail"],
 "lion": ["head", "mane", "eyes", "body", "tail"],
 "tiger": ["head", "ears", "eyes", "body", "tail"],
 "cheetah": ["head", "ears", "eyes", "body", "tail"],
 "brown\_bear": ["head", "ears", "muzzle", "body", "legs"],
 "American\_black\_bear": ["head", "ears", "muzzle", "body", "legs"],
 "ice\_bear": ["head", "ears", "muzzle", "body", "legs"],
 "sloth\_bear": ["head", "ears", "muzzle", "body", "legs"],
 "mongoose": ["head", "ears", "muzzle", "body", "tail"],
 "meerkat": ["head", "ears", "eyes", "body", "tail"],
 "tiger\_beetle": ["head", "thorax", "abdomen", "legs", "antennae"],
 "ladybug": ["head", "thorax", "wings", "legs", "antennae"],
 "ground\_beetle": ["head", "thorax", "abdomen", "legs", "antennae"],
 "long-horned\_beetle": ["head", "thorax", "abdomen", "long antennae", "legs"],
 "leaf\_beetle": ["head", "thorax", "wings", "legs", "antennae"],
 "dung\_beetle": ["head", "thorax", "abdomen", "legs"],
 "rhinoceros\_beetle": ["head horn", "thorax", "abdomen", "legs"],
 "weevil": ["head snout", "thorax", "abdomen", "legs", "antennae"],
 "fly": ["head", "thorax", "abdomen", "wings", "legs"],
 "bee": ["head", "thorax", "abdomen", "wings", "legs"],
 "ant": ["head", "thorax", "abdomen", "antennae", "legs"],
 "grasshopper": ["head", "thorax", "abdomen", "hind legs", "wings"],
 "cricket": ["head", "thorax", "abdomen", "hind legs", "wings"],
 "walking\_stick": ["head", "thorax", "abdomen", "legs"],
 "cockroach": ["head", "thorax", "abdomen", "legs", "antennae"],
 "mantis": ["head", "thorax", "abdomen", "raptorial forelegs", "wings"],
 "cicada": ["head", "thorax", "abdomen", "wings", "eyes"],
 "leafhopper": ["head", "thorax", "abdomen", "wings", "legs"],
 "lacewing": ["head", "thorax", "abdomen", "wings", "antennae"],
 "dragonfly": ["head", "thorax", "abdomen", "wings", "legs"],
 "damselfly": ["head", "thorax", "abdomen", "wings", "legs"],
 "admiral": ["head", "thorax", "abdomen", "wings", "antennae"],
 "ringlet": ["head", "thorax", "abdomen", "wings", "antennae"],
 "monarch": ["head", "thorax", "abdomen", "wings", "antennae"],
 "cabbage\_butterfly": ["head", "thorax", "abdomen", "wings", "antennae"],
 "sulphur\_butterfly": ["head", "thorax", "abdomen", "wings", "antennae"],
 "lycaenid": ["head", "thorax", "abdomen", "wings", "antennae"],
 "starfish": ["central disc", "arms", "tube feet"],
 "sea\_urchin": ["test", "spines", "tube feet"],
 "sea\_cucumber": ["body", "tentacles", "tube feet"],
 "wood\_rabbit": ["head", "ears", "eyes", "body", "tail"],
 "hare": ["head", "ears", "eyes", "body", "tail"],
 "Angora\_rabbit": ["head", "ears", "body", "legs", "tail"],
 "hamster": ["head", "ears", "eyes", "body", "tail"],
 "porcupine": ["head", "quills", "body", "legs"],
 "fox\_squirrel": ["head", "ears", "body", "legs", "tail"],
 "marmot": ["head", "ears", "body", "legs", "tail"],
 "beaver": ["head", "teeth", "body", "tail", "legs"],
 "guinea\_pig": ["head", "ears", "body", "legs", "tail"],
 "sorrel\_horse": ["head", "mane", "body", "legs", "tail"],
 "zebra": ["head", "mane", "body", "legs", "tail"],
 "hog": ["head", "snout", "body", "legs", "tail"],
 "wild\_boar": ["head", "tusks", "body", "legs", "tail"],
 "warthog": ["head", "tusks", "body", "legs", "tail"],
 "hippopotamus": ["head", "eyes and nostrils", "body", "legs", "tail"],
 "ox": ["head", "horns", "body", "legs", "tail"],
 "water\_buffalo": ["head", "horns", "body", "legs", "tail"],
 "bison": ["head", "horns", "body", "legs", "tail"],
 "ram": ["head", "horns", "body", "legs", "tail"],
 "bighorn": ["head", "curved horns", "body", "legs", "tail"],
 "ibex": ["head", "horns", "body", "legs", "tail"],
 "hartebeest": ["head", "horns", "body", "legs", "tail"],
 "impala": ["head", "horns", "body", "legs", "tail"],
 "gazelle": ["head", "horns", "body", "legs", "tail"],
 "Arabian\_camel": ["head", "hump", "neck", "legs", "tail"],
 "llama": ["head", "neck", "body", "legs", "tail"],
 "weasel": ["head", "body", "legs", "tail"],
 "mink": ["head", "body", "legs", "tail"],
 "polecat": ["head", "body", "legs", "tail"],
 "black-footed\_ferret": ["head", "body", "legs", "tail"],
 "otter": ["head", "body", "legs", "tail"],
 "skunk": ["head", "body", "legs", "tail"],
 "badger": ["head", "body", "legs", "tail"],
 "armadillo": ["head", "armor plates", "legs", "tail"],
 "three-toed\_sloth": ["head", "forelimbs", "hindlimbs", "claws"],
 "orangutan": ["head", "torso", "arms", "legs", "face"],
 "gorilla": ["head", "torso", "arms", "legs", "face"],
 "chimpanzee": ["head", "torso", "arms", "legs", "face"],
 "gibbon": ["head", "torso", "arms", "legs", "face"],
 "siamang": ["head", "torso", "arms", "legs", "face"],
 "guenon": ["head", "torso", "arms", "legs", "tail"],
 "patas\_monkey": ["head", "torso", "arms", "legs", "tail"],
 "baboon": ["head", "torso", "arms", "legs", "tail"],
 "macaque": ["head", "torso", "arms", "legs", "tail"],
 "langur": ["head", "torso", "arms", "legs", "tail"],
 "colobus": ["head", "torso", "arms", "legs", "tail"],
 "proboscis\_monkey": ["head", "nose", "torso", "arms", "legs"],
 "marmoset": ["head", "torso", "arms", "legs", "tail"],
 "capuchin": ["head", "torso", "arms", "legs", "tail"],
 "howler\_monkey": ["head", "torso", 
"arms", "legs", "tail"],
 "titi": ["head", "torso", "arms", "legs", "tail"],
 "spider\_monkey": ["head", "torso", "arms", 
         \end{tcolorbox}

\begin{tcolorbox}[
  colback=gray!15,
  colframe=gray!50,
  boxrule=0.5pt
]
"legs", "tail"],
 "squirrel\_monkey": ["head", "torso", "arms", "legs", "tail"],
 "ring\_tailed\_monkey": ["head", "torso", "arms", "legs", "tail"],
 "indri\_monkey": ["head", "torso", "arms", "legs", "tail"],
 "Indian\_elephant": ["head", "trunk", "ears", "tusks", "legs"],
 "African\_elephant": ["head", "trunk", "ears", "tusks", "legs"],
 "lesser\_panda": ["head", "ears", "body", "legs", "tail"],
 "giant\_panda": ["head", "ears", "body", "legs", "tail"],
 "barracouta": ["head", "body", "tail", "fins"],
 "eel": ["head", "body", "tail"],
 "coho": ["head", "body", "tail", "fins"],
 "rock\_beauty": ["head", "body", "tail", "fins"],
 "anemone\_fish": ["head", "body", "tail", "fins"],
 "sturgeon": ["head", "body", "tail", "fins"],
 "gar": ["head", "body", "tail", "fins"],
 "lionfish": ["head", "body", "tail", "fins", "spines"],
 "puffer": ["head", "body", "tail", "fins"],
 "abacus": ["frame", "rods", "beads"],
 "abaya": ["body", "sleeves", "hem", "front opening"],
 "academic\_gown": ["yoke", "sleeves", "body", "hem"],
 "accordion": ["bellows", "keyboard", "bass buttons", "grille"],
 "acoustic\_guitar": ["body", "neck", "fretboard", "headstock", "strings"],
 "aircraft\_carrier": ["flight deck", "island", "hull", "elevator"],
 "airliner": ["nose", "fuselage", "wings", "tail", "engines"],
 "airship": ["envelope", "gondola", "fins", "propellers"],
 "altar": ["table", "top surface", "steps", "ornament"],
 "ambulance": ["cab", "patient compartment", "lights", "wheels"],
 "amphibian": ["head", "body", "limbs"],
 "analog\_clock": ["face", "hands", "numbers", "bezel"],
 "apiary": ["hive boxes", "frames", "entrance"],
 "apron": ["body", "neck strap", "ties", "pocket"],
 "assault\_rifle": ["stock", "barrel", "magazine", "sights"],
 "bakery": ["display case", "oven", "counter", "shelves"],
 "balance\_beam": ["beam", "supports", "base"],
 "balloon": ["envelope", "knot", "string", "valve"],
 "banjo": ["body", "neck", "head", "strings"],
 "bannister": ["handrail", "balusters", "newel post"],
 "barbell": ["bar", "weight plates", "collars"],
 "barber\_chair": ["seat", "backrest", "armrests", "footrest", "lever"],
 "barbershop": ["chair", "mirror", "sink", "counter"],
 "barn": ["roof", "walls", "doors", "loft"],
 "barometer": ["dial", "glass cover", "case"],
 "barrel": ["staves", "hoops", "heads"], "barrow": ["wheel", "tray", "handles", "legs"],
 "baseball": ["leather cover", "stitches", "core"],
 "basketball": ["surface", "seams"],
 "bassinet": ["frame", "mattress", "hood"],
 "bassoon": ["body", "crook", "bell", "keys"],
 "bathing\_cap": ["cap body", "rim"],
 "bathtub": ["bowl", "rim", "drain", "faucet"],
 "beach\_wagon": ["bed", "wheels", "handle"],
 "beacon": ["light", "housing", "base"],
 "beaker": ["body", "lip", "base", "graduations"],
 "bearskin": ["fur cap", "chin strap", "rim"],
 "bell\_cote": ["opening", "roof", "support"],
 "bib": ["body", "neck strap", "ties"],
 "bicycle-built-for-two": ["frame", "seats", "handlebars", "wheels", "chain"],
 "bikini": ["top", "bottom", "ties"],
 "birdhouse": ["entrance", "roof", "body", "perch"],
 "boathouse": ["slip", "roof", "walls", "door"],
 "bobsled": ["nose", "cockpit", "runners", "shell"],
 "bolo\_tie": ["cord", "slide", "tips"],
 "bonnet": ["hood", "rim", "ties"],
 "bookcase": ["shelves", "back", "sides", "top"],
 "bookshop": ["shelves", "counter", "display", "entrance"],
 "bow": ["limb", "string", "grip", "arrow rest"],
 "bow\_tie": ["knot", "wings", "band"],
 "brass": ["bell", "mouthpiece", "valves", "tubing"],
 "brassiere": ["cups", "band", "straps", "hook"],
 "breakwater": ["core", "armor blocks", "seaward face"],
 "breastplate": ["chest panel", "shoulder straps", "lower edge"],
 "buckle": ["frame", "prong", "bar"],
 "bulletproof\_vest": ["front panel", "back panel", "shoulder straps"],
 "bullet\_train": ["nose", "windows", "doors", "pantograph"],
 "butcher\_shop": ["counter", "display case", "hooks", "scale"],
 "cab": ["roof", "doors", "windows", "wheel"],
 "canoe": ["bow", "stern", "hull", "thwarts"],
 "cardigan": ["body", "sleeves", "buttons", "collar"],
 "car\_mirror": ["glass", "housing", "arm", "mount"],
 "carousel": ["platform", "poles", "animals", "canopy"],
 "carpenter\_kit": ["hammer", "saw", "nails", "screwdriver"],
 "carton": ["walls", "top flap", "bottom flap"],
 "car\_wheel": ["rim", "tire", "hub", "spokes"],
 "cash\_machine": ["screen", "keypad", "cash slot", "card slot"],
 "cassette": ["case", "spool holes", "tape"],
 "cassette\_player": ["cassette slot", "buttons", "display", "speaker"],
 "castle": ["tower", "wall", "gate", "keep"],
 "catamaran": ["hulls", "deck", "mast", "crossbeam"],
 "CD\_player": ["tray", "buttons", "display"],
 "cello": ["body", "neck", "fingerboard", "strings", "bridge"],
 "chain": ["links", "end links"],
 "chainlink\_fence": ["posts", "mesh", "top rail"],
 "chain\_mail": ["rings", "coif", "hauberk"],
 "chain\_saw": ["bar", "chain", "engine housing", "handle"],
 "chest": ["lid", "body", "handles", "lock"],
 "chiffonier": ["drawers", "top", "sides", "legs"],
 "chime": ["tubes", "frame", "striker"],
 "china\_cabinet": ["glass doors", "shelves", "base", "top"],
 "Christmas\_stocking": ["body", "cuff", "loop"],
 "church": ["tower", "nave", "roof", "entrance"],
 "cinema": ["screen", "seats", "aisle", "lobby"],
 "cliff\_dwelling": ["rooms", "terraces", "walls"],
 "clog": ["upper", "sole", "heel"],
 "coffeepot": ["body", "spout", "handle", "lid"],
 "coil": ["turns", "core", "lead"],
 "confectionery": ["display", "counter", "shelves"],
 "container\_ship": ["deck", "hull", "bridge", "containers"],
 "convertible": ["hood", "windshield", "seats", "wheels"],
 "corkscrew": ["spiral", "handle", "lever"],
 "cornet": ["bell", "mouthpiece", "valves", "tubing"],
 "cowboy\_boot": ["upper", "heel", "toe", "pull tabs"],
 "cowboy\_hat": ["brim", "crown", "band"],
 "cradle": ["frame", "rocker", "slats", "mattress"],
 "crib": ["slats", "mattress", "rail"],
 "Crock\_Pot": ["pot", "lid", "handle", "base", "dial"],
 "croquet\_ball": ["surface"],
 "crutch": ["handle", "shaft", "pad", "tip"],
 "cuirass": ["chest panel", "shoulder straps", "lower edge"],
 "dam": ["crest", "face", "spillway"],
 "desk": ["top", "drawers", "legs", "keyboard tray"],
 "dial\_telephone": ["handset", "dial", "base", "cord"],
 "diaper": ["absorbent pad", "waistband", "fasteners"],
 "digital\_clock": ["display", "buttons", "case"],
 "dining\_table": ["top", "legs", "apron"],
 "dishwasher": 
         \end{tcolorbox}

\begin{tcolorbox}[
  colback=gray!15,
  colframe=gray!50,
  boxrule=0.5pt
]
["door", "rack", "control panel", "spray arm"],
 "disk\_brake": ["rotor", "caliper", "pad", "hub"],
 "dock": ["deck", "pilings", "cleats"],
 "dogsled": ["sled", "runners", "harness attachments"],
 "dome": ["shell", "drum", "opening"],
 "drum": ["shell", "head", "rim", "lugs"],
 "drumstick": ["shaft", "tip"],
 "Dutch\_oven": ["body", "lid", "handles"],
 "electric\_guitar": ["body", "neck", "pickups", "bridge", "headstock"],
 "electric\_locomotive": ["nose", "cab", "pantograph", "bogies"],
 "entertainment\_center": ["shelves", "cabinet", "openings"],
 "face\_powder": ["compact", "pad", "powder pan"],
 "feather\_boa": ["feathers", "core"],
 "file": ["blade", "handle"],
 "fireboat": ["hull", "deck", "water monitors", "superstructure"],
 "fire\_engine": ["cab", "pump panel", "ladders", "hose"],
 "fire\_screen": ["frame", "mesh panels", "stand"],
 "flagpole": ["pole", "finial", "halyard", "cleat"],
 "flute": ["body", "headjoint", "keys", "footjoint"],
 "folding\_chair": ["seat", "back", "frame", "hinge"],
 "football\_helmet": ["shell", "face mask", "chin strap", "padding"],
 "forklift": ["mast", "forks", "cab", "counterweight"],
 "fountain": ["bowl", "nozzle", "basin"],
 "fountain\_pen": ["nib", "barrel", "cap", "clip"],
 "four-poster": ["posts", "frame", "canopy", "headboard"],
 "freight\_car": ["body", "doors", "bogies"],
 "French\_horn": ["bell", "leadpipe", "rotary valves"],
 "fur\_coat": ["body", "collar", "sleeves", "lining"],
 "garbage\_truck": ["cab", "hopper", "lift mechanism", "wheels"],
 "gasmask": ["facepiece", "filters", "straps"],
 "goblet": ["cup", "stem", "base"],
 "go-kart": ["chassis", "steering wheel", "wheels", "seat"],
 "golf\_ball": ["surface", "dimples"],
 "golfcart": ["body", "seats", "steering", "wheels"],
 "gondola": ["hull", "seat", "oarlock"],
 "gong": ["disk", "suspension", "mallet"],
 "grand\_piano": ["lid", "keyboard", "strings", "legs"],
 "greenhouse": ["glazing", "frame", "benches", "vent"],
 "grocery\_store": ["shelves", "checkout", "produce display"],
 "guillotine": ["frame", "blade", "lunette"], "hair\_slide": ["body", "clip"],
 "hair\_spray": ["can", "nozzle", "cap"],
 "half\_track": ["cab", "tracks", "wheels", "body"],
 "hand-held\_computer": ["screen", "buttons", "housing"],
 "handkerchief": ["cloth", "hem"],
 "hard\_disc": ["platter", "spindle", "housing"],
 "harmonica": ["cover plates", "comb", "reeds"],
 "harp": ["soundboard", "neck", "strings", "pedals"],
 "harvester": ["header", "body", "wheels", "grain tank"],
 "hatchet": ["head", "handle"],
 "holster": ["pouch", "belt loop", "retention strap"],
 "honeycomb": ["cells", "frame"],
 "hook": ["shank", "point", "eye"],
 "hoopskirt": ["hoops", "fabric", "waistband"],
 "horizontal\_bar": ["bar", "supports", "base"],
 "horse\_cart": ["body", "wheels", "shaft", "seat"],
 "hourglass": ["upper bulb", "lower bulb", "narrow waist"],
 "iPod": ["screen", "wheel", "body", "case"],
 "jack-o-lantern": ["hollowed shell", "carved face", "handle"],
 "jeep": ["grille", "windshield", "doors", "wheels"],
 "jigsaw\_puzzle": ["pieces", "image surface"],
 "jinrikisha": ["shaft", "seat", "wheels", "handles"],
 "joystick": ["stick", "base", "buttons"],
 "knot": ["loop", "bight", "tail"],
 "lab\_coat": ["body", "sleeves", "pockets", "collar"],
 "lawn\_mower": ["deck", "blade", "handle", "wheels"],
 "lens\_cap": ["cap body", "retention strap"],
 "library": ["shelves", "reading tables", "stacks"],
 "lifeboat": ["hull", "oars", "seats", "lifelines"],
 "limousine": ["body", "windows", "doors", "wheels"],
 "liner": ["hull", "deck", "superstructure"],
 "lotion": ["bottle", "pump", "cap"],
 "loupe": ["lens", "frame", "handle"],
 "lumbermill": ["saw", "conveyor", "log deck"],
 "magnetic\_compass": ["card", "housing", "needle", "base"],
 "mailbag": ["body", "strap", "closure"],
 "mailbox": ["box", "door", "flag"],
 "maillot\_jersey": ["body", "straps", "leg openings"],
 "maillot\_swimsuit": ["body", "straps", "leg openings"],
 "manhole\_cover": ["cover", "rim"],
 "maraca": ["body", "handle"],
 "marimba": ["keys", "resonators", "frame", "mallets"],
 "mask": ["face piece", "straps", "eye openings"],
 "maypole": ["pole", "ribbons", "base"],
 "maze": ["paths", "walls", "entrance"],
 "medicine\_chest": ["door", "shelves", "mirror"],
 "megalith": ["stone block", "capstone"],
 "microphone": ["head", "body", "stand mount"],
 "military\_uniform": ["jacket", "trousers", "insignia", "buttons"],
 "milk\_can": ["body", "lid", "handle"],
 "minibus": ["body", "doors", "windows", "wheels"],
 "miniskirt": ["waistband", "skirt body", "hem"],
 "minivan": ["body", "sliding door", "windows", "wheels"],
 "missile": ["warhead", "body", "fins"],
 "mobile\_home": ["body", "windows", "door", "skirting"],
 "Model\_T": ["body", "wheels", "roof", "windshield"],
 "modem": ["ports", "indicator lights", "case"],
 "monastery": ["cloister", "church", "cells"],
 "moped": ["frame", "engine", "seat", "wheels"],
 "mortar": ["barrel", "base plate", "breech"],
 "mortarboard": ["board", "tassel", "band"],
 "mosque": ["dome", "minaret", "prayer hall", "entrance"],
 "mosquito\_net": ["mesh", "frame", "suspension"],
 "motor\_scooter": ["deck", "handlebar", "wheels", "seat"],
 "mountain\_bike": ["frame", "fork", "wheels", "handlebar"],
 "mousetrap": ["base", "spring", "bar", "trigger"],
 "moving\_van": ["box", "cab", "door", "lift"],
 "neck\_brace": ["support band", "chin rest", "straps"],
 "nipple": ["teat", "base"],
 "notebook": ["cover", "pages", "binding"],
 "obelisk": ["shaft", "base", "capstone"],
 "oboe": ["body", "keys", "reed"],
 "ocarina": ["body", "mouthpiece", "finger holes"],
 "odometer": ["display", "gear housing"],
 "oil\_filter": ["canister", "filter media", "gasket"],
 "organ": ["pipes", "keyboard", "pedals", "case"],
 "oscilloscope": ["screen", "controls", "case"],
 "overskirt": ["outer skirt", "waistband", "hem"],
 "oxcart": ["bed", "wheels", "yoke"],
 "oxygen\_mask": ["facepiece", "straps", "hose"],
 "packet": ["wrapper", "seal"],
 "paddle": ["shaft", "blade", "grip"],
 "paddlewheel": ["paddles", "axle", "housing"],
 "pajama": ["top", "bottom", "waistband"],
 "palace": ["facade", "towers", "gate", "courtyard"],
 "panpipe": ["pipes", "binding"],
 "parachute": ["canopy", "suspension lines", "harness"],
 "parallel\_bars": ["bars", "supports", "base"],
 "parking\_meter": ["head", 
         \end{tcolorbox}

\begin{tcolorbox}[
  colback=gray!15,
  colframe=gray!50,
  boxrule=0.5pt
]
 "dial", "post"],
 "passenger\_car": ["body", "doors", "windows", "wheels"],
 "pay-phone": ["handset", "coin slot", "dial", "housing"],
 "pedestal": ["platform", "shaft", "base"],
 "Petri\_dish": ["dish", "lid"],
 "photocopier": ["feeder", "glass platen", "control panel", "output tray"],
 "pick": ["tip", "shaft", "handle"],
 "pickelhaube": ["helmet shell", "spike", "chin strap"],
 "picket\_fence": ["posts", "slats", "rails"],
 "pickup\_truck": ["cab", "bed", "wheels", "grille"],
 "pier": ["deck", "piles", "railings"],
 "piggy\_bank": ["body", "slot", "snout", "feet"],
 "ping-pong\_ball": ["surface"],
 "pinwheel": ["blades", "hub", "stick"],
 "pirate": ["hat", "coat", "sash", "boot"],
 "plane": ["nose", "wings", "fuselage", "tail"],
 "planetarium": ["dome", "projector", "seating"],
 "plow": ["blade", "frame", "hitch", "wheels"],
 "Polaroid\_camera": ["body", "lens", "viewfinder", "film slot"],
 "pole": ["shaft", "base", "top"],
 "police\_van": ["cab", "rear compartment", "lights", "wheels"],
 "poncho": ["cape", "neck opening", "hem"],
 "pool\_table": ["playing surface", "rails", "pockets", "legs"],
 "pot": ["rim", "body", "handle", "base"],
 "potter\_wheel": ["head", "foot pedal", "wheel"],
 "prayer\_rug": ["field", "border", "fringe"],
 "prison": ["walls", "towers", "gates", "cells"],
 "missile\_projectile": ["nose", "body", "tail"],
 "projector": ["lens", "body", "mount", "controls"],
 "puck": ["disc"],
 "purse": ["body", "strap", "closure", "pocket"],
 "quill": ["shaft", "barbs", "tip"],
 "quilt": ["patches", "binding", "stitch lines"],
 "racer": ["cockpit", "body", "wheels", "wing"],
 "radiator": ["fins", "inlet", "outlet", "core"],
 "radio": ["speaker", "tuning dial", "antenna", "housing"],
 "radio\_telescope": ["dish", "feed", "support structure"],
 "rain\_barrel": ["body", "inlet", "spigot", "lid"],
 "recreational\_vehicle": ["cab", "living area", "windows", "wheels"],
 "reel": ["spool", "handle", "frame"],
 "reflex\_camera": ["body", "lens", "viewfinder", "shutter"],
 "refrigerator": ["door", "shelves", "handle", "compressor"],
 "restaurant": ["entrance", "dining area", "counter", "kitchen view"],
 "revolver": ["barrel", "cylinder", "grip", "trigger"],
 "rifle": ["barrel", "stock", "sight", "trigger"],
 "rocking\_chair": ["seat", "back", "rockers", "armrests"],
 "rotisserie": ["spit", "motor", "frame", "drip tray"],
 "rubber\_eraser": ["body", "edge"],
 "rugby\_ball": ["panel", "seams"],
 "safe": ["door", "dial", "body", "hinge"],
 "sarong": ["wrap", "waist tie", "hem"],
 "sax": ["body", "neck", "mouthpiece", "keys"],
 "school\_bus": ["front", "windows", "doors", "wheels"],
 "schooner": ["hull", "masts", "sails", "deck"],
 "screen": ["frame", "mesh", "mount"],
 "screwdriver": ["shaft", "tip", "handle"],
 "sewing\_machine": ["needle area", "bed", "hand wheel", "foot pedal"],
 "shield": ["face", "rim", "handle"],
 "shoe\_shop": ["shelves", "display", "counter"],
 "shoji": ["frame", "paper panels", "sliding track"],
 "shopping\_cart": ["basket", "handle", "wheels"],
 "shower\_curtain": ["curtain panel", "grommets", "rod"],
 "ski": ["tip", "camber", "binding", "tail"],
 "ski\_mask": ["face opening", "eye openings", "edge"],
 "slide\_rule": ["body", "slider", "scales"],
 "sliding\_door": ["panel", "track", "handle"],
 "slot": ["opening", "face", "controls"],
 "snowmobile": ["front ski", "track", "seat", "handlebar"],
 "snowplow": ["blade", "frame", "hydraulics", "mount"],
 "soccer\_ball": ["panels", "seams"],
 "solar\_dish": ["dish", "receiver", "mount"],
 "sombrero": ["crown", "brim", "band"],
 "space\_bar": ["keycap"],
 "space\_shuttle": ["nose", "fuselage", "wings", "tail"],
 "speedboat": ["hull", "deck", "windshield", "engine"],
 "spider\_web": ["radial threads", "spiral threads", "anchor points"],
 "spindle": ["shaft", "whorl", "hook"],
 "sports\_car": ["body", "windshield", "wheels", "spoiler"],
 "spotlight": ["lamp", "reflector", "housing", "mount"],
 "stage": ["platform", "backdrop", "human", "lighting"],
 "steam\_locomotive": ["boiler", "cab", "tender", "wheels"],
 "steel\_arch\_bridge": ["arch", "deck", "support piers"],
 "steel\_drum": ["body", "lip", "notes area"],
 "stethoscope": ["chest piece", "tubing", "earpieces"],
 "stole": ["strip", "ends"],
 "stone\_wall": ["stones", "mortar", "cap"],
 "stopwatch": ["face", "buttons", "case"],
 "stove": ["burners", "control knobs", "oven door"],
 "streetcar": ["body", "windows", "doors", "wheels"],
 "stretcher": ["frame", "canvas", "handles", "wheels"],
 "studio\_couch": ["seat", "back", "arms", "legs"],
 "stupa": ["dome", "harmika", "base"],
 "submarine": ["conning tower", "hull", "propeller", "rudder"],
 "sundial": ["dial plate", "gnomon", "base"],
 "sunglass": ["lens", "frame", "temples"],
 "sunscreen": ["tube", "cap", "dispensing nozzle"],
 "suspension\_bridge": ["towers", "cables", "deck"],
 "swing": ["seat", "chains", "support"],
 "switch": ["toggle", "plate", "mount"],
 "tank": ["turret", "body", "tracks", "gun barrel"],
 "tape\_player": ["reels", "controls", "play head"],
 "tennis\_ball": ["surface"],
 "thatch": ["bundles", "ridge", "eaves"],
 "theater\_curtain": ["curtain panel", "hem", "pelmet"],
 "thimble": ["cap", "rim"],
 "thresher": ["drum", "feeder", "frame"],
 "throne": ["seat", "back", "armrests", "base"],
 "tile\_roof": ["tiles", "ridge", "eaves"],
 "tobacco\_shop": ["shelves", "display", "counter"],
 "toilet\_seat": ["seat", "lid", "hinges"],
 "torch": ["head", "body", "switch", "lens"],
 "tow\_truck": ["cab", "boom", "wheels", "winch"],
 "toyshop": ["shelves", "display", "counter"],
 "tractor": ["body", "wheels", "cab", "three point hitch"],
 "trailer\_truck": ["cab", "trailer", "wheels", "doors"],
 "trench\_coat": ["collar", "buttons", "belt", "sleeves"],
 "tricycle": ["frame", "front wheel", "rear wheels", "seat"],
 "trimaran": ["center hull", "amas", "deck", "mast"],
 "tripod": ["legs", "head", "mount"],
 "triumphal\_arch": ["archway", "column", "entablature"],
 "trombone": ["slide", "bell", "mouthpiece"],
 "tub": ["rim", "bowl", "drain"],
 "turnstile": ["arms", "shaft", "housing"],
 "typewriter\_keyboard": ["keycaps", "carriage", "space bar"],
 "unicycle": ["wheel", "seat", "pedal"],
 "upright\_piano": ["body", "keyboard", "strings", "pedal"],
 "vault": ["door", "chamber", "locking mechanism"],
 "velvet": ["pile", "edge"],
 "vending\_machine": ["front panel", "selection buttons", "dispense slot"],
 "viaduct": 
         \end{tcolorbox}

\begin{tcolorbox}[
  colback=gray!15,
  colframe=gray!50,
  boxrule=0.5pt
]["arches", "deck", "piers"],
 "violin": ["body", "neck", "fingerboard", "strings", "bridge"],
 "volleyball": ["panels", "seams"],
 "waffle\_iron": ["plates", "hinge", "handle"],
 "wall\_clock": ["face", "hands", "case"],
 "wardrobe": ["doors", "rails", "drawers", "top"],
 "warplane": ["nose", "wings", "fuselage", "engines"],
 "washbasin": ["bowl", "tap", "overflow"],
 "washer": ["drum", "door", "control panel"],
 "water\_jug": ["neck", "body", "handle", "base"],
 "water\_tower": ["tank", "support", "ladder"],
 "wig": ["cap", "hair"],
 "window\_screen": ["frame", "mesh", "spline"],
 "window\_shade": ["shade panel", "roller", "cord"],
 "Windsor\_tie": ["knot", "blade", "tail"],
 "wing": ["airfoil", "flap", "slat"],
 "wool": ["fibre", "bundle"],
 "worm\_fence": ["posts", "rails", "weave"],
 "wreck": ["bow", "stern", "hull"],
 "yawl": ["hull", "mast", "sails", "rudder"],
 "yurt": ["wall", "roof ring", "cover", "door"],
 "web\_site": ["header", "navigation", "content area", "footer"],
 "crossword\_puzzle": ["grid", "clues", "filled squares"],
 "street\_sign": ["post", "sign panel", "mount"],
 "traffic\_light": ["lanes", "lights", "housing"],
 "menu": ["cover", "pages", "items"],
 "guacamole": ["bowl", "dip"],
 "consomme": ["bowl", "liquid"],
 "hot\_pot": ["pot", "broth", "ingredients"],
 "trifle": ["layers", "glass", "topping"],
 "ice\_cream": ["scoop", "cone", "topping"],
 "bagel": ["crust", "hole", "surface"],
 "pretzel": ["twist", "surface"],
 "cheeseburger": ["bun", "patty", "cheese", "toppings"],
 "hotdog": ["bun", "sausage", "toppings"],
 "mashed\_potato": ["pile", "scoop"],
 "head\_cabbage": ["outer leaves", "core", "head"],
 "broccoli": ["head", "stalk", "florets"],
 "cauliflower": ["curd", "stem", "florets"],
 "zucchini": ["stem end", "body", "blossom end"],
 "spaghetti\_squash": ["skin", "flesh", "seeds"],
 "acorn\_squash": ["skin", "flesh", "stem"],
 "butternut\_squash": ["skin", "flesh", "stem"],
 "cucumber": ["stem end", "body", "blossom end"],
 "artichoke": ["head", "bracts", "stem"],
 "bell\_pepper": ["stem", "body", "core"],
 "cardoon": ["stem", "leaves"],
 "mushroom": ["cap", "gills", "stem"],
 "Granny\_Smith": ["skin", "flesh", "stem"],
 "strawberry": ["body", "calyx", "seeds"],
 "fig": ["skin", "flesh", "seeds"],
 "pineapple": ["crown", "skin", "core"],
 "jackfruit": ["skin", "flesh pods", "seeds"],
 "custard\_apple": ["skin", "segments", "seeds"],
 "pomegranate": ["crown", "arils", "rind"],
 "hay": ["stalks", "bales"],
 "chocolate\_sauce": ["bottle", "pour stream"],
 "dough": ["surface", "edges"],
 "meat\_loaf": ["slice", "crust"],
 "pizza": ["crust", "cheese", "toppings", "slice"],
 "potpie": ["crust", "filling"],
 "burrito": ["tortilla", "filling", "ends"],
 "red\_wine": ["glass", "liquid"],
 "espresso": ["cup", "crema"],
 "eggnog": ["cup", "foam"],
 "alp": ["peak", "slope", "snow cap"],
 "bubble": ["film", "reflection"],
 "cliff": ["face", "edge", "base"],
 "coral\_reef": ["reef crest", "patches", "channels"],
 "geyser": ["vent", "plume", "pool"],
 "lakeside": ["shore", "water", "vegetation"],
 "promontory": ["headland", "cliff", "shoreline"],
 "sandbar": ["crest", "sand", "shallow water"],
 "seashore": ["beach", "waves", "shoreline"],
 "valley": ["floor", "sides", "stream"],
 "volcano": ["cone", "crater", "lava flow"],
 "ballplayer": ["helmet", "jersey", "glove"],
 "groom": ["suit", "tie", "bouquet"],
 "scuba\_diver": ["mask", "tank", "fins", "wetsuit"],
 "rapeseed": ["flowers", "stalks", "pods"],
 "daisy": ["disk", "petals", "stem"],
 "yellow\_lady\_slipper": ["pouch", "sepals", "stem"],
 "corn": ["ear", "kernels", "husks"],
 "acorn": ["cap", "nut"],
 "hip": ["fruit", "sepals"],
 "buckeye": ["shell", "nut"],
 "coral\_fungus": ["branches", "tips"],
 "agaric": ["cap", "gills", "stem"],
 "gyromitra": ["cap", "stem"],
 "stinkhorn": ["stalk", "cap", "gleba"],
 "earthstar": ["central disc", "rays"],
 "hen-of-the-woods": ["clusters", "caps"],
 "bolete": ["cap", "pore surface", "stem"],
 "corn\_ear": ["kernels", "cob", "husk", "lobe", "corn"]
\}

\textbf{ImageNet/ObjectNet-54 Class names}: ['snorkel', 'loudspeaker', 'syringe', 'binoculars', 'cannon', 'computer\_keyboard', 'crane\_machine', 'desktop\_computer', 'gas\_pump', 'remote\_control', 'screw', 'muzzle', 'seat\_belt', 'patio', 'home\_theater', 'totem\_pole', 'grille', 'mountain\_tent', 'scoreboard', 'plate\_rack', 'ashcan', 'cocktail\_shaker', 'kimono', 'whiskey\_jug', 'knee\_pad', 'book\_jacket', 'crash\_helmet', 'vestment', 'cloak', 'scabbard', 'beer\_glass', 'swab', 'drilling\_platform', 'sweatshirt', 'pencil\_box', 'punching\_bag', 'pencil\_sharpener', 'broom', 'cup', 'shower\_cap', 'envelope', 'trolleybus', 'perfume', 'crate', 'ballpoint', 'comic\_book', 'shopping\_basket', 'wooden\_spoon', 'ice\_lolly', 'carbonara', 'paintbrush', 'iron', 'toaster', 'caldron']

\textbf{ImageNet/ObjectNet-104 Class names}: ['backpack', 'banana', 'Band\_Aid', 'shopping\_basket', 'bath\_towel', 'beer\_bottle', 'park\_bench', 'binder', 'bottlecap', 'French\_loaf', 'broom', 'bucket', 'cleaver', 'can\_opener', 'candle', 'cellular\_telephone', 'hamper', 'espresso\_maker', 'combination\_lock', 'mouse', 'table\_lamp', 'dishrag', 'doormat', 'Loafer', 'power\_drill', 'cup', 'plate\_rack', 'envelope', 'electric\_fan', 'frying\_pan', 'gown', 'hand\_blower', 'hammer', 'iron', 'jean', 'computer\_keyboard', 'ladle', 'lampshade', 'laptop', 'lemon', 'letter\_opener', 'lighter', 'lipstick', 'matchstick', 'measuring\_cup', 'microwave', 'mixing\_bowl', 'monitor', 'coffee\_mug', 'nail', 'necklace', 'orange', 'padlock', 'paintbrush', 'paper\_towel', 'pill\_bottle', 'pillow', 'pitcher', 'plastic\_bag', 'plate', 'plunger', 'pop\_bottle', 'space\_heater', 'printer', 'remote\_control', 'rule', 'running\_shoe', 'safety\_pin', 'saltshaker', 'sandal', 'screw', 'shovel', 'sleeping\_bag', 'soap\_dispenser', 'sock', 'soup\_bowl', 'spatula', 'loudspeaker', 'strainer', 'teddy', 'suit', 'sunglasses', 'sweatshirt', 'swimming\_trunks', 'jersey', 'television', 'teapot', 'racket', 'toaster', 'toilet\_tissue', 'ashcan', 'tray', 'umbrella', 'vacuum', 'vase', 'wallet', 'digital\_watch', 'water\_bottle', 'dumbbell', 'scale', 'whistle', 'wine\_bottle', 'mitten', 'wok']

\end{tcolorbox}

\begin{tcolorbox}[
  colback=gray!15,
  colframe=gray!50,
  boxrule=0.5pt
]
\textbf{CUB}

\textbf{ID Class names}: ['Sooty\_Albatross', 'Crested\_Auklet', 'Least\_Auklet', 'Brewer\_Blackbird', 'Rusty\_Blackbird', 'Indigo\_Bunting', 'Spotted\_Catbird', 'Brandt\_Cormorant', 'Bronzed\_Cowbird', 'American\_Crow', 'Mangrove\_Cuckoo', 'Purple\_Finch', 'Acadian\_Flycatcher', 'Least\_Flycatcher', 'Vermilion\_Flycatcher', 'American\_Goldfinch', 'Eared\_Grebe', 'Horned\_Grebe', 'Evening\_Grosbeak', 'Pine\_Grosbeak', 'California\_Gull', 'Heermann\_Gull', 'Herring\_Gull', 'Ivory\_Gull', 'Western\_Gull', 'Anna\_Hummingbird', 'Pomarine\_Jaeger', 'Florida\_Jay', 'Tropical\_Kingbird', 'Belted\_Kingfisher', 'Pied\_Kingfisher', 'Hooded\_Merganser', 'Baltimore\_Oriole', 'Hooded\_Oriole', 'Brown\_Pelican', 'Common\_Raven', 'Loggerhead\_Shrike', 'Baird\_Sparrow', 'Brewer\_Sparrow', 'Fox\_Sparrow', 'Grasshopper\_Sparrow', 'Harris\_Sparrow', 'Henslow\_Sparrow', 'Lincoln\_Sparrow', 'Savannah\_Sparrow', 'Song\_Sparrow', 'Vesper\_Sparrow', 'Barn\_Swallow', 'Cliff\_Swallow', 'Scarlet\_Tanager', 'Artic\_Tern', 'Caspian\_Tern', 'Common\_Tern', 'Sage\_Thrasher', 'Philadelphia\_Vireo', 'Warbling\_Vireo', 'Canada\_Warbler', 'Cerulean\_Warbler', 'Kentucky\_Warbler', 'Magnolia\_Warbler', 'Mourning\_Warbler', 'Myrtle\_Warbler', 'Nashville\_Warbler', 'Prairie\_Warbler', 'Prothonotary\_Warbler', 'Swainson\_Warbler', 'Tennessee\_Warbler', 'Wilson\_Warbler', 'Louisiana\_Waterthrush', 'Bohemian\_Waxwing', 'Pileated\_Woodpecker', 'Downy\_Woodpecker', 'Bewick\_Wren', 'Carolina\_Wren']

\textbf{Coarse-grained OOD Class names}: ['White\_breasted\_Nuthatch', 'Horned\_Puffin', 'Sayornis', 'Northern\_Fulmar', 'Brown\_Creeper', 'Green\_tailed\_Towhee', 'Mallard', 'Northern\_Flicker', 'Ovenbird', 'Chuck\_will\_Widow', 'Nighthawk', 'Cardinal', 'Bobolink', 'Geococcyx', 'Gadwall', 'American\_Pipit', 'Whip\_poor\_Will', 'Green\_Violetear', 'Cape\_Glossy\_Starling', 'Yellow\_breasted\_Chat', 'Western\_Wood\_Pewee', 'Eastern\_Towhee', 'Frigatebird', 'Boat\_tailed\_Grackle', 'Western\_Meadowlark', 'Pigeon\_Guillemot', 'Clark\_Nutcracker', 'Dark\_eyed\_Junco', 'American\_Redstart', 'Common\_Yellowthroat', 'Pacific\_Loon', 'Mockingbird', 'Horned\_Lark', 'Groove\_billed\_Ani', 'Red\_legged\_Kittiwake']

\textbf{Fine-grained OOD Class names}: ['Northern\_Waterthrush', 'Marsh\_Wren', 'White\_eyed\_Vireo', 'Blue\_Jay', 'European\_Goldfinch', 'Cactus\_Wren', 'Cedar\_Waxwing', 'Nelson\_Sharp\_tailed\_Sparrow', 'Tree\_Sparrow', 'Rufous\_Hummingbird', 'Shiny\_Cowbird', 'White\_Pelican', 'Black\_billed\_Cuckoo', 'Scissor\_tailed\_Flycatcher', 'Yellow\_headed\_Blackbird', 'Blue\_winged\_Warbler', 'Red\_winged\_Blackbird', 'Gray\_Kingbird', 'Western\_Grebe', 'Le\_Conte\_Sparrow', 'White\_breasted\_Kingfisher', 'Fish\_Crow', 'Chipping\_Sparrow', 'House\_Wren', 'Tree\_Swallow', 'Long\_tailed\_Jaeger', 'Red\_eyed\_Vireo', 'Red\_faced\_Cormorant', 'Ringed\_Kingfisher', 'Winter\_Wren', 'Black\_and\_white\_Warbler', 'Rhinoceros\_Auklet', 'Field\_Sparrow', 'Black\_throated\_Blue\_Warbler', 'Yellow\_throated\_Vireo', 'Slaty\_backed\_Gull', 'Red\_bellied\_Woodpecker', 'Bank\_Swallow', 'Bay\_breasted\_Warbler', 'Rose\_breasted\_Grosbeak', 'Green\_Jay', 'Pelagic\_Cormorant', 'Glaucous\_winged\_Gull', 'White\_necked\_Raven', 'Summer\_Tanager', 'White\_throated\_Sparrow', 'Green\_Kingfisher', 'Black\_Tern', 'Hooded\_Warbler', 
'Yellow\_Warbler', 'Ring\_billed\_Gull', 'Lazuli\_Bunting', 'Laysan\_Albatross', 'Olive\_sided\_Flycatcher', 'Seaside\_Sparrow', 'Parakeet\_Auklet', 'Gray\_crowned\_Rosy\_Finch', 'Gray\_Catbird', 'Orange\_crowned\_Warbler', 'Pied\_billed\_Grebe', 'Red\_breasted\_Merganser', 'Yellow\_bellied\_Flycatcher', 'Painted\_Bunting', 'Black\_throated\_Sparrow', 'Yellow\_billed\_Cuckoo', 'Orchard\_Oriole', 'Forsters\_Tern', 'Pine\_Warbler', 'White\_crowned\_Sparrow', 'Black\_footed\_Albatross', 'Black\_capped\_Vireo', 'Great\_Grey\_Shrike', 'Blue\_Grosbeak', 'Rock\_Wren', 'Palm\_Warbler', 'House\_Sparrow', 'Least\_Tern', 'Worm\_eating\_Warbler', 'Clay\_colored\_Sparrow', 'Chestnut\_sided\_Warbler', 'Golden\_winged\_Warbler', 'Scott\_Oriole', 'Red\_headed\_Woodpecker', 'Cape\_May\_Warbler', 'Blue\_headed\_Vireo', 'American\_Three\_toed\_Woodpecker', 'Red\_cockaded\_Woodpecker', 'Ruby\_throated\_Hummingbird', 'Great\_Crested\_Flycatcher', 'Elegant\_Tern', 'Brown\_Thrasher']

\textbf{1-Component name}: \ \ ['body']

\textbf{2-Component names}: ['head', 'body']

\textbf{3-Component names}: ['head', 'body', 'legs']

\textbf{4-Component names}: ['head', 'body', 'wing', 'legs']

\textbf{5-Component names}: ['head', 'body', 'wing', 'legs', 'tail']

\textbf{6-Component names}: ['head', 'body', 'back', 'wing', 'legs', 'tail']
\end{tcolorbox}

\section{FPR Reduction at Small Component Numbers}
\label{pbfpr}
Constrained by the finite capacity of visual encoders to represent extremely fine-grained or low-resolution components, we consider the applicability of the normal approximation when the number of components is small. 

We retain the notation from the main paper. For each component index $p$ define the existence score as $s_{yp}\!=\!\langle\bm z_{yp},\bm t_{yp}\rangle$, and model the indicator $\mathds{1}\{\!s_{yp}\!>\!\tau\}$ as a Bernoulli variable with probability of being an ID component $\psi\!=\!P(\!s_{yp}\!>\!\tau|\bm x\!)$. The aggregated score from $|\mathcal{P}_y|$ components $\mathcal{S}_{|\mathcal{P}_y|}=\sum_{p=1}^{|\mathcal{P}_y|}\mathds{1}\{s_{yp}>\tau\}$ follows a Binomial distribution $\mathcal{B}(|\mathcal{P}_y|,\psi)$. 
Fix a target ID true positive rate $\lambda$, for each component number $|\mathcal{P}_y|$, let
$T_{|\mathcal{P}_y|}\in\mathbb{Z}_{\ge 0}$ denote the integer threshold chosen by the minimal integer satisfying the TPR constraint, \textit{i.e.} $T_{|\mathcal{P}_y|}=\min\big\{t\in\mathbb{Z}:\ \mathbb{P}\big(\mathcal S_{|\mathcal{P}_y|}>t\mid\mathrm{ID}\big)\le\lambda\big\}$. The corresponding false positive rate FPR$_{\lambda}$ is given by:
\begin{equation}
\text{FPR}_{\lambda}(|\mathcal{P}_y|) = P(\mathcal{S}_{|\mathcal{P}_y|}>T_{|\mathcal{P}_y|}|\mathrm{OOD})=\sum_{p>T_{|\mathcal{P}_y|}}^{{|\mathcal{P}_y|}}\binom{|\mathcal{P}_y|}{p}(\psi^\mathrm{out})^p(1-\psi^\mathrm{out})^{|\mathcal{P}_y|-p}.
\end{equation}
For an arbitrary integer $T$, the binomial recursion gives:
\begin{equation}
\label{eq:recursion}
P(\mathcal S_{|\mathcal{P}_y|+1}>T\mid\mathrm{OOD})
=(1-\psi^{\mathrm{out}})\,P(\mathcal S_{|\mathcal{P}_y|}>T\mid\mathrm{OOD})
+\psi^{\mathrm{out}}\,P(\mathcal S_{|\mathcal{P}_y|}>T-1\mid\mathrm{OOD}).
\end{equation}

We consider the change in FPR when adding one component:
\begin{equation}
    \Delta=\text{FPR}_{\lambda}(|\mathcal{P}_y|+1)-\text{FPR}_{\lambda}(|\mathcal{P}_y|)=P(\mathcal{S}_{|\mathcal{P}_y|+1}>T_{|\mathcal{P}_y|+1}|\mathrm{OOD})-P(\mathcal{S}_{|\mathcal{P}_y|}>T_{|\mathcal{P}_y|}|\mathrm{OOD})
\end{equation}

Applying \cref{eq:recursion} with $T=T_{|\mathcal{P}_y|+1}$ and expanding the binomial sums yields:
\begin{align}
\Delta
&=(1-\psi^\mathrm{out})P(\mathcal{S}_{|\mathcal{P}_y|}>T_{|\mathcal{P}_y|+1}|\mathrm{OOD})+\psi^\mathrm{out} P(\mathcal{S}_{|\mathcal{P}_y|}>T_{|\mathcal{P}_y|+1}-1|\mathrm{OOD})-P(\mathcal{S}_{|\mathcal{P}_y|}>T_{|\mathcal{P}_y|}|\mathrm{OOD})\notag\\
&=(1-\psi^\mathrm{out})\sum_{p=T_{|\mathcal{P}_y|+1}+1}^{|\mathcal{P}_y|}\binom{|\mathcal{P}_y|}{p}
(\psi^\mathrm{out})^p(1-\psi^\mathrm{out})^{|\mathcal{P}_y|-p}+
\psi^\mathrm{out}\sum_{p=T_{|\mathcal{P}_y|+1}}^{|\mathcal{P}_y|}\binom{|\mathcal{P}_y|}{p}
(\psi^\mathrm{out})^p(1-\psi^\mathrm{out})^{|\mathcal{P}_y|-p}\notag\\
&\ \ \ \ \ -\sum_{p=T_{|\mathcal{P}_y|}+1}^{|\mathcal{P}_y|}\binom{|\mathcal{P}_y|}{p}
(\psi^\mathrm{out})^p(1-\psi^\mathrm{out})^{|\mathcal{P}_y|-p}\notag
\end{align}
\begin{align}
\label{taylor2}
&=\sum_{p=T_{|\mathcal{P}_y|+1}+1}^{|\mathcal{P}_y|}\binom{|\mathcal{P}_y|}{p}
(\psi^\mathrm{out})^p(1-\psi^\mathrm{out})^{|\mathcal{P}_y|-p}+
\psi^\mathrm{out}\sum_{p=T_{|\mathcal{P}_y|+1}}^{T_{|\mathcal{P}_y|+1}}\binom{|\mathcal{P}_y|}{p}
(\psi^\mathrm{out})^p(1-\psi^\mathrm{out})^{|\mathcal{P}_y|-p}\ \ \ \ \ \ \ \ \ \ \notag\\
&\ \ \ \ \ -\sum_{p=T_{|\mathcal{P}_y|+1}+1}^{|\mathcal{P}_y|}\binom{|\mathcal{P}_y|}{p}
(\psi^\mathrm{out})^p(1-\psi^\mathrm{out})^{|\mathcal{P}_y|-p}-\sum_{p=T_{|\mathcal{P}_y|}+1}^{T_{|\mathcal{P}_y|+1}}\binom{|\mathcal{P}_y|}{p}
(\psi^\mathrm{out})^p(1-\psi^\mathrm{out})^{|\mathcal{P}_y|-p},
\end{align}
where the first and third terms cancel out, the second term has $p=T_{|\mathcal{P}_y|+1}$, and the fourth term exists only if $T_{|\mathcal{P}_y|+1}\ge T_{|\mathcal{P}_y|}+1$. Evaluate $\Delta$ in the two possible cases, \textit{i.e.},
\begin{equation}
\Delta\!=\!\left\{\!
\begin{array}{ll}
    \textstyle \ \ \ \psi^\mathrm{out}\ \ \binom{|\mathcal{P}_y|}{T_{|\mathcal{P}_y|\!+\!1}}
(\psi^\mathrm{out})^{T_{|\mathcal{P}_y|\!+\!1}}(1-\psi^\mathrm{out})^{|\mathcal{P}_y|-T_{|\mathcal{P}_y|\!+\!1}}\ge0,\!&\! T_{|\mathcal{P}_y|\!+\!1}\!=\!T_{|\mathcal{P}_y|},\\
    \!\textstyle \!(\!\psi^\mathrm{\!out}\!-\!1\!)\!\binom{|\mathcal{P}_y|}{T_{|\mathcal{P}_y|+1}}\!
(\!\psi^\mathrm{\!out}\!)^{T_{|\mathcal{P}_y|+1}}\!(\!1\!-\!\psi^\mathrm{\!out}\!)^{\!|\mathcal{P}_y|-T_{|\mathcal{P}_y|+1}}\!-\!\sum\limits_{\!p\!=\!T_{|\mathcal{P}_y|}\!+\!1}^{T_{|\mathcal{P}_y|\!+\!1}\!-\!1}\!\binom{|\mathcal{P}_y|}{p}\!
(\!\psi^\mathrm{\!out}\!)^p(\!1\!-\!\psi^\mathrm{out}\!)^{\!|\mathcal{P}_y|\!-\!p}\!\le\!0,\!&\!T_{|\mathcal{P}_y|+1}\!\ge\! T_{|\mathcal{P}_y|}\!+\!1.
\end{array}
\right.
\end{equation}
When $\psi^\mathrm{out}<\psi^\mathrm{in}\le1$, $\Delta<0$ in the second case. 
That is, adding a component tends to increase the FPR$_\lambda$ if the threshold $T_{|\mathcal{P}_y|+1}$ does not move, and the FPR$_\lambda$ decreases when the threshold increases.

Then we link this formulation to $\psi^{\mathrm{in}}$. 
The threshold update occurs when the ID-side probability with $|\mathcal{P}_y|+1$ components exceeds the previous threshold $\lambda$, \textit{i.e.}
\begin{align}
\label{eq:threshold-update}
P(\mathcal{S}_{|\mathcal{P}_y|+1}>&T_{|\mathcal{P}_y|}|\mathrm{ID})=(1-\psi^\mathrm{in})P(\mathcal{S}_{|\mathcal{P}_y|}>T_{|\mathcal{P}_y|}|\mathrm{ID})+\psi^\mathrm{in}P(\mathcal{S}_{|\mathcal{P}_y|}>T_{|\mathcal{P}_y|}-1|\mathrm{ID})>\lambda.
\end{align}

Note that \cref{eq:threshold-update} increases monotonically \textit{w.r.t.} $\psi^{\mathrm{in}}$, because $P(\mathcal{S}_{|\mathcal{P}_y|}>T_{|\mathcal{P}_y|}|\mathrm{ID})<P(\mathcal{S}_{|\mathcal{P}_y|}>T_{|\mathcal{P}_y|}-1|\mathrm{ID})$ in practice. Therefore, a larger $\psi^{\mathrm{in}}$ increases the probability of the threshold to increase, which reduces FPR when the update occurs. 
Taking expectation over multistep threshold update gives that, in aggregate, an increase in $\psi^{\mathrm{in}}$ together with a decrease in $\psi^{\mathrm{out}}$ is more likely to make $\mathrm{FPR}_\lambda$ decreases after introducing a new component.

\end{document}